\documentclass[10pt,twocolumn,letterpaper]{article}
\pdfoutput=1 

\usepackage{cvpr}
\usepackage{times}
\usepackage{epsfig}
\usepackage{graphicx}
\graphicspath{{figures/}}

\usepackage{amsmath}
\usepackage{amssymb}

\usepackage{xspace}
\usepackage{caption}
\usepackage{subcaption}
\usepackage{tikz}
\usepackage{multirow}

\usepackage{algorithm,algpseudocode}
\newcommand{\rev}[1]{#1}
\newcommand{\revb}[1]{#1}

\usepackage[pagebackref=true,breaklinks=true,letterpaper=true,colorlinks,bookmarks=false]{hyperref}

\cvprfinalcopy 

\newcommand{\image}{\mathbf{I}}
\newcommand{\clip}{\mathbf{V}}

\newcommand{\pose}{\mathbf{p}}
\newcommand{\confidence}{\mathbf{c}}
\newcommand{\action}{\mathbf{a}}

\newcommand{\pfeat}{\mathcal{X}}
\newcommand{\afeat}{\mathcal{Y}}
\newcommand{\zfeat}{\mathcal{Z}}
\newcommand{\vfeat}{\mathcal{V}}

\newcommand{\probmap}{\mathbf{h}}
\newcommand{\depthmap}{\mathbf{d}}

\newcommand{\h}{\mathit{h}}
\newcommand{\xd}{\mathit{d}}

\newcommand{\njoints}{N_j}
\newcommand{\nactions}{N_a}

\newcommand{\ok}{\checkmark}

\begin{document}

\title{
  Multi-task Deep Learning for Real-Time 3D Human Pose Estimation and Action
  Recognition
}

\author{Diogo C. Luvizon$^{1,2}$ \hspace{1cm}  Hedi Tabia$^{1,3}$  \hspace{1cm} David Picard$^{1,4}$
\vspace{0.254cm}\\
$^1$ETIS UMR 8051, Paris Seine University, ENSEA, CNRS, F-95000, Cergy, France\\
$^2$Advanced Technologies, Samsung Research Institute, Campinas, SP, Brazil\\
$^3$IBISC, Univ. d\'{}Evry Val d\'{}Essonne, Universit\'{e} Paris Saclay\\
$^4$LIGM, UMR 8049, \'{E}cole des Ponts, UPE, Champs-sur-Marne, France\\
{\tt\small diogo.luvizon@ensea.fr}
}

\maketitle

\begin{abstract}
  Human pose estimation and action recognition are related tasks since both
  problems are strongly dependent on the human body representation and
  analysis. Nonetheless, most recent methods in the literature handle the two
  problems separately.
  In this work, we propose a multi-task framework for jointly estimating 2D or
  3D human poses from monocular color images and classifying human actions from
  video sequences.
  We show that a single architecture can be used to solve both problems in an
  efficient way and still achieves state-of-the-art or comparable results at
  each task while running with a throughput of more than 100 frames per second.
  The proposed method benefits from high parameters sharing between the two
  tasks by unifying still images and video clips processing in a single
  pipeline, allowing the model to be trained with data from different
  categories simultaneously and in a seamlessly way.
  Additionally, we provide important insights for end-to-end training the
  proposed multi-task model by decoupling key prediction parts, which
  consistently leads to better accuracy on both tasks.
  The reported results on four datasets (MPII, Human3.6M, Penn Action and NTU
  RGB+D) demonstrate the effectiveness of our method on the targeted tasks.
  Our source code and trained weights are publicly available at
  \url{https://github.com/dluvizon/deephar}.
\end{abstract}




\section{Introduction}
\label{sec:introduction}

\begin{figure}[htbp]
  \centering
  \includegraphics[width=8.45cm]{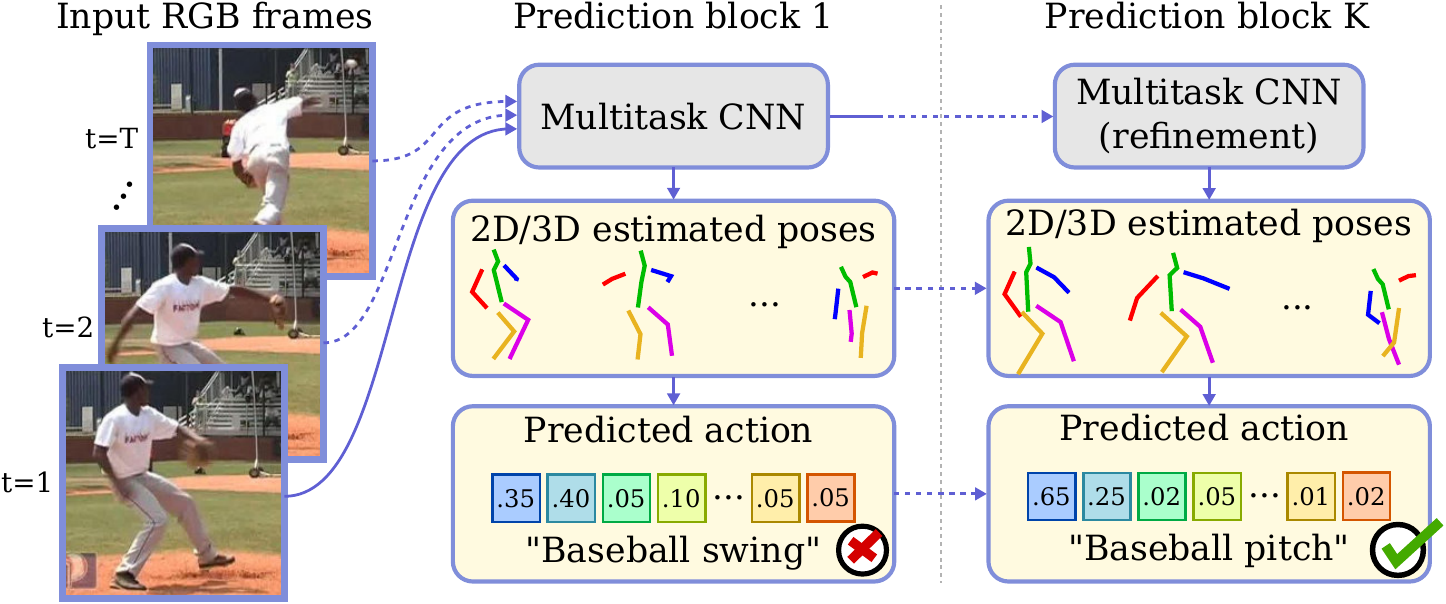}
  \caption{
    The proposed multi-task approach for human pose estimation and action
    recognition.  Our method provides 2D/3D pose estimation from single images
    or frame sequences. Pose and visual information are used to predict actions
    in a unified framework and both predictions are refined by K prediction
    blocks.
  }
  \label{fig:intro}
\end{figure}

Human action recognition has been intensively studied in the
last years, specially because it is a very challenging problem, but also due to
the several applications that can benefit from it.
Similarly, human pose estimation has also rapidly progressed with the advent
of powerful methods based on convolutional neural networks (CNN) and deep
learning.
Despite the fact that action recognition benefits from precise body
poses, the two problems are usually handled as distinct tasks in
the literature~\cite{cheronICCV15}, or action recognition is used
as a prior for pose estimation~\cite{Yao2012, Iqbal_2017}.
To the best of our knowledge, there is no recent method in the literature that
tackles both problems in a joint way to the benefit of action recognition.
In this paper, we propose a unique end-to-end trainable
multi-task framework to handle human pose estimation and action recognition
jointly, as illustrated in Fig.~\ref{fig:intro}.

One of the major advantages of deep learning methods is their capability to
perform end-to-end optimization. This is all the more true for multi-task
problems, where related tasks can benefit from one another, as suggested by
Kokkinos~\cite{Kokkinos17cvpr}.
\rev{
  Action recognition and pose estimation are usually hard to be stitched
  together to perform a beneficial joint optimization, usually requiring 3D
  convolutions~\cite{Zolfaghari_2017_ICCV} or heatmaps
  transformations~\cite{Choutas_2018_CVPR}.}
Detection based approaches require the non-differentiable argmax function
to recover the joint coordinates as a post processing stage, which breaks the
backpropagation chain needed for end-to-end learning.
We propose to solve this problem by extending the differentiable
soft-argmax~\cite{Luvizon_2017_CoRR, Yi_2016} for joint 2D and 3D pose
estimation.
This allows us to stack action recognition on top of pose estimation, resulting
in a multi-task framework trainable from end-to-end.

In comparison with our previous work~\cite{Luvizon_2018_CVPR}, we propose
a new network architecture carefully designed for pose and action prediction
simultaneously at different feature map resolutions.
Each prediction is supervised and re-injected into the network for further
refinement.
Differently from~\cite{Luvizon_2018_CVPR}, where we first predict poses then
actions, here poses and actions are predicted in parallel and successively
refined, strengthening the multi-task aspect of our method.
Another improvement is the proposed depth estimation approach for 3D poses,
which allows us to depart from learning the costly volumetric heat maps
while improving the overall accuracy of the method.

The main contributions of our work are presented as follows:
\textit{First}, we propose a new multi-task method for jointly estimating
2D/3D human poses and recognizing associated actions. Our method is simultaneously trained from end-to-end
for both tasks with multimodal data, including still images and video clips.
\textit{Second}, we propose a new regression approach for 3D pose estimation
from single frames, benefiting at the same time from images ``in-the-wild''
with 2D annotated poses and 3D data.  This has been proven a very efficient
way to learn good visual features, which is also very important for action
recognition.
\textit{Third}, our action recognition approach is based only on RGB images,
from which we extract 3D poses and visual information. Despite that, our
multi-task method achieves state-of-the-art on both 2D and 3D scenarios, even
when compared with methods using ground-truth poses.
\textit{Fourth}, the proposed network architecture is scalable without any
additional training procedure, which allows us to choose the right trade-off
between speed and accuracy \textit{a posteriori}.
Finally, we show that the hard problem of multi-tasking pose estimation and
action recognition can be tackled efficiently by a single and carefully
designed architecture, handling both problems together and in a better way than
separately.
As a result, our method provides acceptable pose and action predictions at more
than 180 frames per second (FPS), while achieving its best scores at 90 FPS on a customer GPU.

The remaining of this paper is organized as follows.
In Section~\ref{sec:relatedwork} we present a review of the most
relevant works related to our method.
The proposed multi-task framework is presented in Section~\ref{sec:multitask}.
Extensive experiments on both pose estimation and action recognition
are presented in Section~\ref{sec:experiments}, followed by our conclusions in
Section~\ref{sec:conclusions}.

\section{Related Work}
\label{sec:relatedwork}


In this section, we present some of the most relevant methods related to our
work, which are divided into \textit{human pose estimation} and \textit{action
recognition}.
Since an extensive literature review is out of the scope of the paper, we
encourage the readers to refer to the surveys in~\cite{SARAFIANOS20161,
Herath_2017} for respectively pose estimation and action recognition.

\subsection{Human Pose Estimation}

\subsubsection{2D Pose Estimation}

The problem of human pose estimation has been intensively studied in the
last years, from Pictorial Structures~\cite{Andriluka_CVPR_2009,
Dantone_CVPR_2013, Pishchulin_CVPR_2013} to more recent CNN based
approaches~\cite{Ning2017, Lifshitz_ECCV_2016, Pishchulin_CVPR_2016,%
Insafutdinov_ECCV_2016, Rafi_BMVC_2016, Wei_CVPR_2016,%
Belagiannis_ICCV_2015, Tompson_CVPR_2015, Toshev_CVPR_2014, pfister2014deep}.
We can identify from the literature two distinct families of methods
for pose estimation: detection and regression based methods.
Recent detection methods handle pose estimation as a heat map prediction
problem, where each pixel in a heat map represents the detection score of a
given body joint being localized at this pixel~\cite{Bulat_ECCV_2016,
Gkioxari_ECCV_2016}.
Exploring the concepts of stacked architectures, residual connections, and
multiscale processing, Newell
\etal~\cite{Newell_ECCV_2016} proposed
the Stacked Hourglass networks (SHG), which improved scores on 2D pose
estimation challenges significantly.
Since then, methods in the state of the art are frequently proposing complex
variations of the SHG architecture.
For example, Chu \etal~\cite{Chu_CVPR_17} proposed an attention model based
on conditional random field (CRF) and Yang \etal~\cite{Yang_2017_ICCV}
replaced the residual unit from SHG by the Pyramid Residual Module (PRM).
\rev{Very recently, \cite{Sun_2019_CVPR} proposed a high-resolution network
that keeps a high-resolution flow, resulting in more precise predictions.}
With the emergence of Generative Adversarial Networks
(GANs)~\cite{Goodfellow_GANS}, Chou \etal~\cite{ChouCC17} proposed to
use a discriminative network to distinguish between estimated and target heat
maps. This process could increase the quality of predictions, since the
generator is stimulated to produce more plausible predictions. Another
application of GANs in that sense is to enforce the structural representation
of the human body~\cite{Chen_2017_ICCV}.

However, all the previous mentioned detection based approaches do not provide
body joint coordinates directly.  To recover the body joints in $(x, y)$
coordinates, predicted heat maps have to be converted to joint positions,
generally using the argument of the maximum a posteriori probability (MAP),
called $\mathit{argmax}$.
On the other hand, regression based approaches use a nonlinear function to
project the input image directly to the desired output, which can be the joint
coordinates.
Following this paradigm, Toshev and Szegedy~\cite{Toshev_CVPR_2014} proposed
a holistic solution based on cascade regression for body part regression and
Carreira \etal~\cite{Carreira_CVPR_2016} proposed the Iterative Error
Feedback.
The limitation of current regression methods is that the regression function is
frequently sub-optimal. In order to tackle this weakness, the soft-argmax
function~\cite{Luvizon_2017_CoRR} has been proposed to compute body joint
coordinates from heat maps in a differentiable way.

\subsubsection{3D Pose Estimation}

Recently, deep architectures have been used to learn 3D representations
from RGB images~\cite{Zhou_2017, Tome_2017_CVPR, Martinez_2017, Tekin_2016,
MehtaRCSXT16, Popa_2017_CVPR} thanks to the availability of high precise
3D data~\cite{h36m_pami}, and are now able to surpass
depth-sensors~\cite{VNect_SIGGRAPH2017}.
Chen and Ramanan~\cite{Chen_2017_CVPR} divided the problem of 3D pose estimation
into two parts. First, they target 2D pose estimation considering the
camera coordinates and second, the 2D estimated poses are matched to 3D
representations by means of a nonparametric shape model.
However, this is an ill-defined problem, since two different 3D poses could
have the same 2D projection.
Other methods propose to regress the 3D relative position of joints, which
usually presents a lower variance than the absolute position. For example,
Sun \etal~\cite{Sun_2017_ICCV} proposed a bone representation of the human
body. However, since the errors are accumulative, such a structural
transformation might effect tasks that depend on the extremities of the human
body, like action recognition.

Pavlakos \etal~\cite{Pavlakos_2017_CVPR} proposed the volumetric stacked
hourglass architecture, but the method suffers from significant increase in
the number of parameters and from the required memory to store all the gradients.
A similar technique is used in~\cite{Sun_2018_ECCV}, but instead of using
argmax for coordinate estimation, the authors use a numerical integral
regression, which is similar to the soft-argmax
operation~\cite{Luvizon_2018_CVPR}.
More recently, Yang \etal~\cite{Yang_2018_CVPR} proposed to use adversarial
networks to distinguish between generated and ground truth poses, improving
predictions on uncontrolled environments.
\rev{
  Differently form our previous work in~\cite{Luvizon_2018_CVPR}, we show that
  a volumetric representation is not required for 3D prediction. Similarly to
  methods on hand pose estimation~\cite{Iqbal_2018_ECCV} and on
3D human pose estimation~\cite{VNect_SIGGRAPH2017}}, we predict 2D
depth maps which encode the relative depth of each body joint.

\subsection{Action Recognition}

\subsubsection{2D Action Recognition}

\rev{
  In this section we revisited some methods that exploit pose information
  for action recognition.} For example, classical methods for feature
extraction have been used in~\cite{Nie_2015_CVPR, Jhuang_2013_ICCV}, where the key idea is to use
body joint locations to select visual features in space and time.
3D convolutions have been stated as the best option to handle the temporal
dimension of images sequences~\cite{Cao_2017, Carreira_2017_CVPR, varol17a},
but they involve a high number of parameters and cannot efficiently benefit
from the abundant still images during training.
Another option to integrate the temporal aspect is by analysing motion from
image sequences~\cite{cheronICCV15, Du_2017_ICCV}, but these methods require
the difficult estimation of optical flow.
Unconstrained temporal and spatial analysis are also promising approaches to
tackle action recognition, since it is very likely that, in a sequence of
frames, some very specific regions in a few frames are more
relevant than the remaining parts. Inspired on this observation,
Baradel \etal~\cite{Baradel_CVPR_2018} proposed an attention model called
Glimpse Clouds, which learns to focus on specific image patches in space and
time, aggregating the patterns and soft-assigning each feature to workers
that contribute to the final action decision.
The influence of occlusions could be alleviated by multi-view
videos~\cite{Wang_2018_ECCV} and inaccurate pose sequences could be replaced by
heat maps for better accuracy~\cite{Liu_2018_CVPR}.  However, this
improvement is not observed when pose predictions are sufficiently precise.

2D action recognition methods \rev{usually} use the body joint information only
to extract localized visual features~\cite{Nie_2015_CVPR, cheronICCV15}, as an
attention mechanism.  Methods that directly explore the body joints
\rev{usually do not generate it~\cite{Jhuang_2013_ICCV} or present lower
precision with estimated poses~\cite{Cao_2017}}.
Our approach removes these limitations by performing pose estimation
together with action recognition. As such, our model only needs the
input RGB frames while still performing discriminative visual
recognition guided by the estimated body joints.

\begin{figure*}[htbp]
  \centering
  \includegraphics[width=0.95\textwidth]{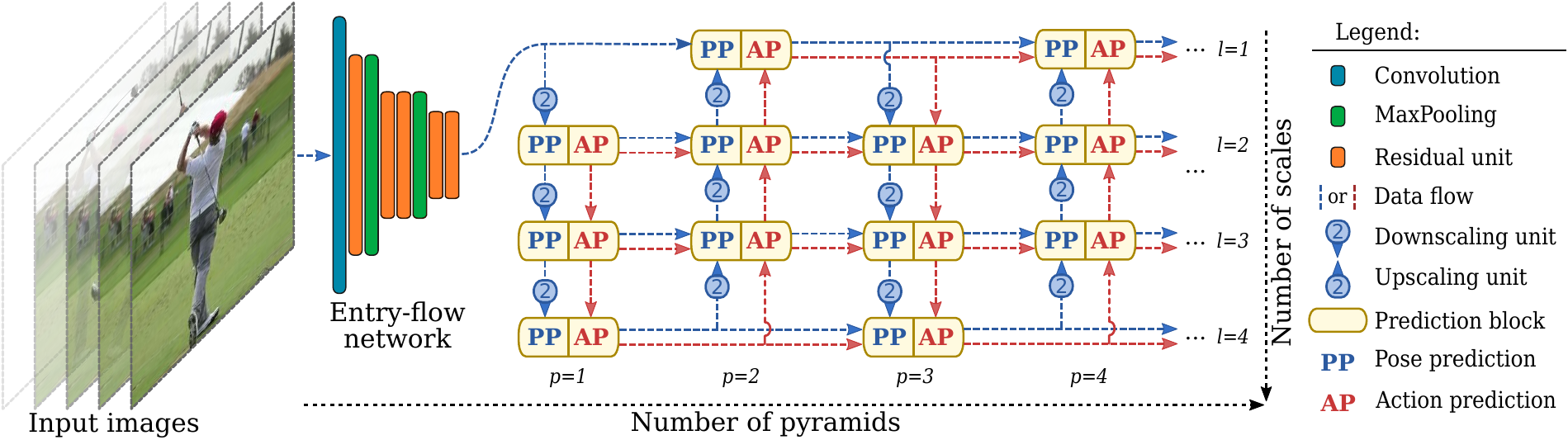}
  \caption{
    Overview of the proposed multi-task network architecture.  The entry-flow
    extracts feature maps from the input images, which are fed through a
    sequence of CNNs composed of prediction blocks (PB), downscaling and
    upscaling units (DU and UU), and simple (skip) connections.  Each PB
    outputs supervised pose and action predictions that are refined by further
    blocks and units.  The information flow related to pose estimation and
    action recognition are independently propagated from one prediction block
    to another, respectively depicted by blue and red arrows.  See
    Fig.~\ref{fig:units} and Fig.~\ref{fig:prediction-block} for details about
    DU, UU, and PB.
  }
  \label{fig:netarch}
\end{figure*}

\subsubsection{3D Action Recognition}

Differently from video based action recognition, 3D action recognition is
mostly based on skeleton data as the primary
information~\cite{Luvizon_PRL_2017, Presti20163DSH}.
With depth sensors such as the Microsoft Kinect, it is
possible to capture 3D skeletal data without a complex installation
procedure frequently required for motion capture systems (MoCap).
However, due to the required infrared projector, depth sensors are limited
to indoor environments, have a low range of operation, and are not robust to
occlusions, frequently resulting in noisy skeletons.
To cope with the noisy skeletons, Spatio-Temporal LSTM networks~\cite{Liu2016}
have been widely used to learn the reliability of skeleton sequences or as an
attention mechanism~\cite{Liu_2017_CVPR, Song_2017_AAAI}.  In addition to the
skeleton data, multimodal approaches can also benefit from visual
cues~\cite{Shahroudy2017DeepMF}.  In that direction, pose-conditioned attention
mechanisms have been proposed~\cite{baradel2017a} to focus on image patches
centered around the hands.

Since our architecture predicts precise 3D poses from RGB frames,
we do not have to cope with the noisy skeletons from Kinect.
Moreover, we show in the experiments that, despite being based on
temporal convolution instead of the more common LSTM, our system is
able to reach state of the art performance on 3D action recognition,
indicating that action recognition does not \rev{necessarily} require long term
memory.

\section{Proposed Multi-task Approach}
\label{sec:multitask}

\rev{The goal of the proposed method is to jointly handle human pose estimation
and action recognition, prioritizing the use of predicted poses on action
recognition and benefiting from shared computations between the two tasks.}
For convenience, we define
the input of our method as either a still RGB image
$\image{} \in \mathbb{R}^{H\times{W}\times3}$
or a video clip (sequence of images)
$\clip{} \in \mathbb{R}^{T\times{H}\times{W}\times3}$,
where $T$ is the number of frames in a video clip and $H\times{W}$ is the frame
size. 
\rev{This distinction is important because we handle pose estimation as a
single frame problem.}
The outputs of our method for each frame are: predicted human pose
$\hat{\pose{}} \in \mathbb{R}^{\njoints{}\times{3}}$
and per body joint confidence score
$\hat{\confidence{}} \in \mathbb{R}^{\njoints{}\times{1}}$,
where $\njoints{}$ is the number of body joints.
When taking a video clip as input, the method also outputs a vector of action
probabilities $\hat{\action{}} \in \mathbb{R}^{\nactions{}\times{1}}$, where
$\nactions{}$ is the number of action classes.
To simplify notation, in this section we omit batch normalization layers
and ReLU activations, which are used in between convolutional layers
as a common practice in deep neural networks.

\subsection{Network Architecture}

\begin{figure}[htbp]
  \centering
  \begin{subfigure}[b]{0.15\textwidth}
    \centering
    \includegraphics[scale=0.85]{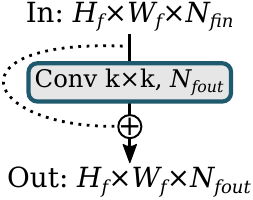}
    \caption{}
    \label{fig:net-sepconv}
  \end{subfigure}
  \begin{subfigure}[b]{0.15\textwidth}
    \centering
    \includegraphics[scale=0.85]{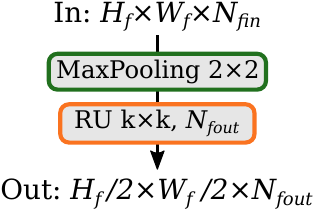}
    \caption{}
    \label{fig:net-downscaling}
  \end{subfigure}
  \begin{subfigure}[b]{0.15\textwidth}
    \centering
    \includegraphics[scale=0.85]{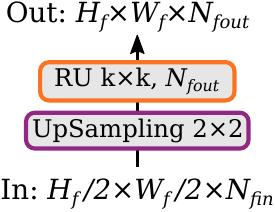}
    \caption{}
    \label{fig:net-upscaling}
  \end{subfigure}
  \caption{
    Network elementary units: in (a) residual unit (RU), in (b) downscaling
    unit (DU), and in (c) upscaling unit (UU).  $N_{fin}$ and $N_{fout}$
    represent the input and output number of features, $H_f\times{W_f}$ is the
    feature map size, and $k$ is the filter size.
  }
  \label{fig:units}
\end{figure}

\rev{Differently from our previous work~\cite{Luvizon_2018_CVPR} where poses and
actions are predicted sequentially, here we want to strengthen the multi-task
aspect of our method by predicting and refining poses and actions in parallel.
This is implemented by the proposed architecture, illustrated in Fig.~\ref{fig:netarch}. }
Input images are fed through the entry-flow, which
extracts low level visual features. The extracted features are then processed
by a sequence of downscaling and upscaling pyramids indexed by
$p \in \{1, 2, \dots, P\}$, which are respectively
composed of downscaling and upscaling units (DU and UU), and prediction blocks
(PB), indexed by $l \in \{1, 2, \dots, L\}$. Each PB is supervised on pose and action predictions, which are then
re-injected into the network, producing a new feature map that is refined by
further downscaling and upscaling pyramids.  Downscaling or upscaling units
are respectively composed by maxpooling or upsampling layers followed by a
residual unit that is a standard or a depthwise separable
convolution~\cite{Chollet_2017_CVPR} with skip connection. These units are
detailed in Fig.~\ref{fig:units}.

\rev{In order to be able to handle human poses and actions in
a unified framework,} the network can operate into two distinct modes: (i)
\textit{single frame} processing or (ii) \textit{video clip} processing.  In
the first operational mode (single frame), only layers related to pose
estimation are active, from which connections correspond to the blue arrows in
Fig.~\ref{fig:netarch}.  In the second operational mode (video clip), both pose
estimation and action recognition layers are active.  In this case, layers in
the single frame processing part handle each video frame as a single sample in
the batch.  Independently on the operational mode, pose estimation is always
performed from single frames, which prevents the method from depending on the
temporal information for this task.  For video clip processing, the information
flow from single frame processing (pose estimation) and from video clip
processing (action recognition) are independently propagated from one
prediction block to another, as demonstrated in Fig.~\ref{fig:netarch}
respectively by blue and red arrows.

\subsubsection{Multi-task Prediction Block}

\begin{figure}[htbp]
  \centering
  \includegraphics[width=0.485\textwidth]{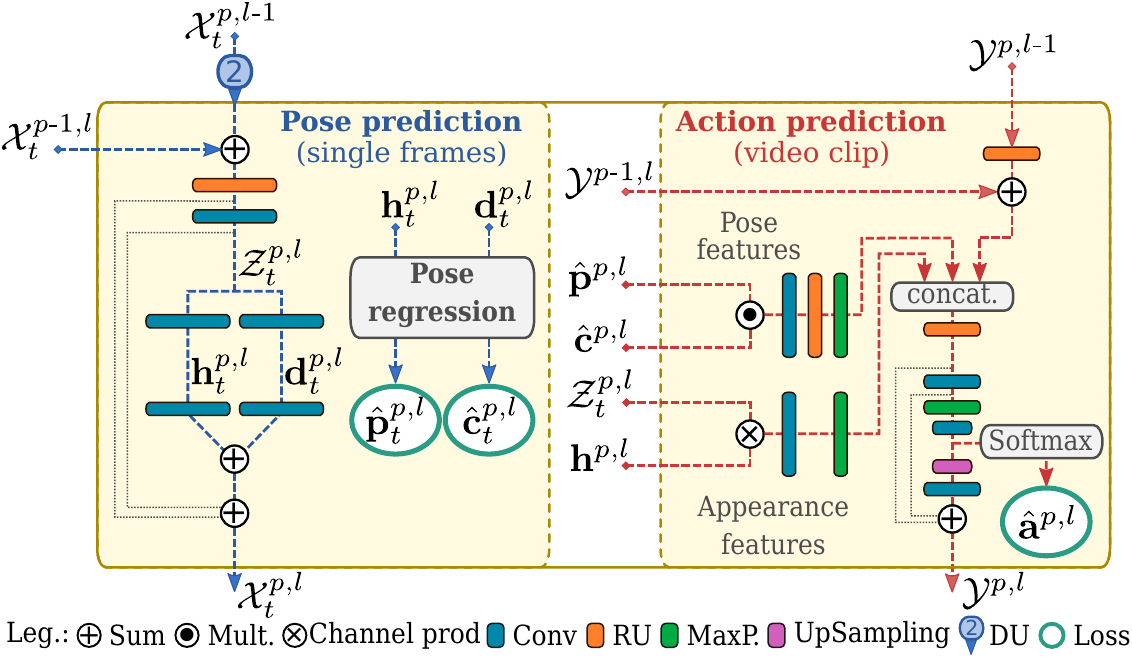}
  \caption{
    Network architecture of prediction blocks (PB) for a downscaling pyramid.
    With the exception of the PB in the first pyramid, all PB get as input
    features from the previous pyramid in the same level ($\pfeat{}_t^{p-1,l}$,
    $\afeat{}^{p-1,l}$), and features from lower or higher levels
    ($\pfeat{}_t^{p,l\mp{1}}$, $\afeat{}^{p,l\mp{1}}$), depending if it
    composes a downscaling or an upscaling pyramid, respectively.
  }
  \label{fig:prediction-block}
\end{figure}

\rev{
The main challenges related to the design of the network architecture is how
to handle multimodal data (single frames and video clips) in a unified way and
how to allow predictions refinement for both poses and actions. To this end,
we propose a multi-task prediction block (PB),
}
detailed in
Fig.~\ref{fig:prediction-block}. In the PB, pose and action are simultaneously
predicted and re-injected into the network for further refinement. In the
global architecture, each PB is indexed by pyramid $p$ and level $l$, and
produces the following three feature maps:
\begin{align}
  \pfeat{}_t^{p,l} &\in \mathbb{R}^{H_f\times{W_f}\times{N_f}}\\
  \zfeat{}_t^{p,l} &\in \mathbb{R}^{H_f\times{W_f}\times{N_f}}\\
  \afeat{}^{p,l} &\in \mathbb{R}^{T\times{\njoints{}}\times{N_v}}.
\end{align}
Namely, $\pfeat{}_t^{p,l}$ is a tensor of single frame features, which is
propagated from one PB to another, $\zfeat{}_t^{p,l}$ is a tensor of multi-task
(single frame) features used for both pose and action, and $\afeat{}^{p,l}$ is
a tensor of video clip features, exclusively used for action predictions and
also propagated from one PB to another. $t = \{1,\dots,{T}\}$ is the index of
single frames in a video clip, and $N_f$ and $N_v$ are respectively the size
of single frame features and video clip features.

For pose estimation, prediction blocks take as input the single frame features
$\pfeat{}_t^{p-1,l}$ from the previous pyramid and the features
$\pfeat{}_t^{p,l\mp{1}}$ from lower or higher levels, respectively for
downscaling and upscaling pyramids.  A similar propagation of previous features
$\afeat{}^{p-1,l}$ and $\afeat{}^{p,l\mp{1}}$
happens for action. Note that both $\pfeat{}_t^{p,l}$ and $\afeat{}^{p,l}$
feature maps are three-dimensional tensors (2D maps plus channels) that can be
easily handled by 2D convolutions.

The tensor of multi-task features is defined by:
\begin{align}
  \label{eq:zfeatprime}
  \zfeat{}_t^{'p,l} &= RU(\pfeat{}_t^{p-1,l} + DU(\pfeat{}_t^{p,l-1}))\\
  \label{eq:zfeat}
  \zfeat{}_t^{p,l} &= \textbf{W}_z^{p,l}\ast{\zfeat{}_t^{'p,l}},
\end{align}
where DU is the downscaling unit (replaced by UU for upscaling pyramids), RU is
the residual unit, $\ast{}$ is a convolution, and $\textbf{W}_z^{p,l}$ is a
weight matrix.
Then, $\zfeat{}_t^{p,l}$ is used to produce body joint probability maps:
\begin{equation}
  \probmap{}_t^{p,l} = \Phi(\textbf{W}_h^{p,l}\ast{\zfeat{}_t^{p,l}}),
  \label{eq:probmap}
\end{equation}
and body joint depth maps:
\begin{equation}
  \depthmap{}_t^{p,l} = Sigmoid(\textbf{W}_d^{p,l}\ast{\zfeat{}_t^{p,l}}),
  \label{eq:depthmap}
\end{equation}
where $\Phi$ is the spatial softmax~\cite{Luvizon_2017_CoRR}, and
$\textbf{W}_h^{p,l}$ and $\textbf{W}_d^{p,l}$ are weight matrices.
Probability maps and body joint depth maps encode, respectively, the probability of
a body joint being at a given location and the depth with respect to the root
joint, normalized in the interval $[0, 1]$. Both $\probmap{}_t^{p,l}$ and
$\depthmap{}_t^{p,l}$ have shape
$\mathbb{R}^{H_f\times{W_f}\times{\njoints{}}}$.

\subsection{Pose Regression}
\label{sec:pose-regression}

Once a set of body joint probability maps and depth maps are computed from
multi-task features, we aim to estimate the corresponding
3D points by a differentiable and non-parametrized function. For that, we
decouple the problem in \textit{2D pose estimation} and \textit{depth
estimation}, and the final 3D pose is the concatenation of the intermediate
parts.

\subsubsection{The Soft-argmax Layer for 2D Estimation}
\label{sec:sam}

Given a 2D input signal, the main idea is to consider that the argument of the
maximum (\textit{argmax}) can be approximated by the expectation of the input
signal after being normalized to have the properties of a distribution.
Indeed, for a sufficiently pointy (Leptokurtic) distribution, the expectation
should be close to the maximum a posteriori (MAP) estimation.
For a 2D heat map as input, the normalized exponential function (softmax)
can be used, since it alleviates the undesirable influences of values below
the maximum and increases the ``pointiness'' of the resulting distribution,
producing a probability map, as defined in Equation~\ref{eq:probmap}.

Let's define a single probability map for the $j$th joint as $\h{}^j$,
in such a way that $\probmap{} \equiv [\h{}^1, \dots, \h{}^{\njoints{}}]$.
Then, the expected coordinates $(x^j, y^j)$ are given by the function $\Psi$:
\begin{equation}
  \Psi(\h{}^j) =\\
  \left(\sum_{c=0}^{W_{\h{}}}\sum_{r=0}^{H_{\h{}}}\frac{c}{W_{\h{}}}\h{}_{r,c},\\
        \sum_{c=0}^{W_{\h{}}}\sum_{r=0}^{H_{\h{}}}\frac{r}{H_{\h{}}}\h{}_{r,c}\right),
  \label{eq:sam}
\end{equation}
where $H_{\h{}}\times{W_{\h{}}}$ is the size of the input probability map, and
$l$ and $c$ are line and column indexes of $\h{}$.
According to Equation~\ref{eq:sam}, the coordinates $(x^j, y^j)$ are constrained
between the interval $[0,1]$, which corresponds to the normalized limits of the
input image.

\subsubsection{Depth Estimation}
\label{sec:depthestimation}

Differently from our previous work~\cite{Luvizon_2018_CVPR}, where volumetric
heat maps were required to estimate the third dimension of body joints, here
\rev{we use a similar apprach to~\cite{Iqbal_2018_ECCV}, where} specialized
depth maps $\depthmap{}$ \rev{are used} to encode the depth information. Similarly to the
probability maps decomposition from section~\ref{sec:sam}, here we define
$\xd{}^j$ as a depth map for the $j$th body joint. Thus, the regressed depth
coordinate $z^j$ is defined by:
\begin{equation}
  z^j = \sum_{c=0}^{W_{\h{}}}\sum_{r=0}^{H_{\h{}}}\h{}_{r,c}^j\xd{}_{r,c}^j.
  \label{eq:depth}
\end{equation}
Since $\h{}^j$ is a normalized unitary and positive probability map,
Equation \ref{eq:depth} represents a spatially weighted pooling
of depth map $\xd{}^j$ based on the 2D body joint location.

\subsubsection{Body Joint Confidence Scores}

The probability of a certain body joint being present (even if occluded)
in the image is computed by the maximum value in the corresponding
probability map. Considering a pose layout with $\njoints{}$ body joints,
the estimated joint confidence vector is represented by
$\hat{\confidence{}} \in \mathbb{R}^{\njoints{}\times{1}}$.
If the probability map is very pointy, this score is close to 1.
On the other hand, if the probability map is uniform or has more than one
region with high response, the confidence score drops.

\subsubsection{Pose Re-injection}
\label{sec:pose-re-inject}

As systematically noted in recent works~\cite{Bulat_ECCV_2016,
Gkioxari_ECCV_2016, Newell_ECCV_2016, Pavlakos_2017_CVPR}, predictions
re-injection is a very efficient way to improve precision on estimated poses.
Differently from all previous methods based on direct heat map regression, our
approach can benefit from prediction re-injection at different resolutions,
since our pose regression method is invariant to the feature map resolution.
Specifically, in each PB at different pyramid and different level, we compute a
new set of features $\pfeat{}_t^{p,l}$ based on features from previous blocks
and on the current prediction, as follows:
\begin{equation}
  \pfeat{}_t^{p,l} = \textbf{W}_r^{p,l}\ast{\probmap{}_t^{p,l}}
  + \textbf{W}_s^{p,l}\ast{\depthmap{}_t^{p,l}}
  + \zfeat{}_t^{'p,l} + \zfeat{}_t^{p,l},
  \label{eq:reinject}
\end{equation}
where $\textbf{W}_r^{p,l}$ and $\textbf{W}_s^{p,l}$ are weight matrices
related to the re-injection of 2D pose and depth information,
respectively. With this approach, further PB at different pyramids and levels
are able to refine predictions, considering different sets of features at
different resolutions.

\subsection{Human Action Recognition}

Another important advantage in our method is its ability to integrate high
level pose information with low level visual features in a multi-task framework.
This characteristic allows sharing the single frame processing pipeline
for both pose estimation and visual features extraction.
Additionally, visual features are trained using both action sequences and still
images captured ``in-the-wild'', which have been proven as a very efficient way
to learn robust visual representations.
As shown in Fig.~\ref{fig:prediction-block}, the action prediction part takes
as input two different sources of information: \textit{pose features} and
\textit{appearance features}.
Additionally, similarly to the pose prediction part, action features from
previous pyramids ($\afeat{}^{p-1,l}$) and levels ($\afeat{}^{p,l\mp{1}}$) are
also aggregated in each prediction.

\subsubsection{Pose Features}
\label{sec:pose-features}

\begin{figure}[htbp]
  \centering
  \begin{subfigure}[b]{0.20\textwidth}
    \centering
    \includegraphics[scale=0.5]{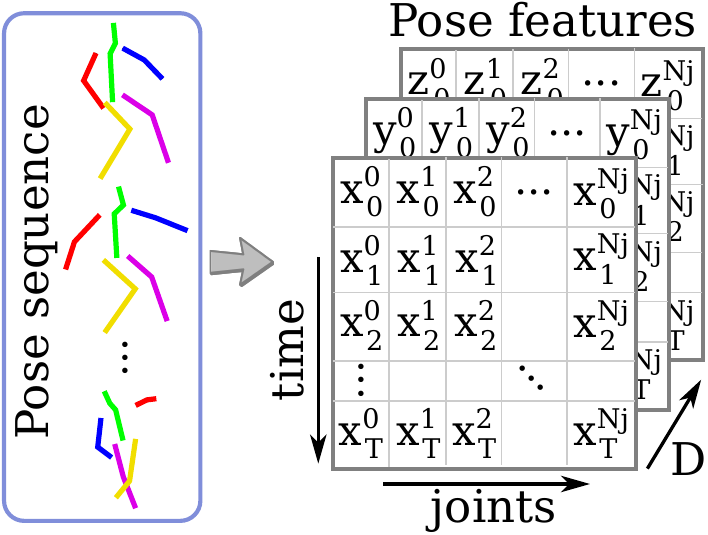}
    \caption{}
    \label{fig:pose-features}
  \end{subfigure}\hspace{2mm}
  \begin{subfigure}[b]{0.24\textwidth}
    \centering
    \includegraphics[scale=0.85]{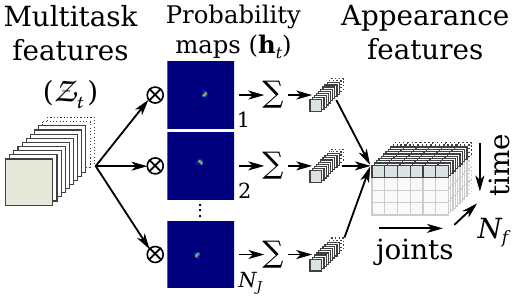}
    \caption{}
    \label{fig:appearance-features}
  \end{subfigure}
  \caption{
    Extraction of (a) pose and (b) appearance features.
  }
  \label{fig:action-features}
\end{figure}

In order to explore the rich information encoded with body joint positions, we
convert a sequence of $T$ poses with $\njoints{}$ joints each into an
image-like representation.
\rev{Similar representations were previously used in~\cite{baradel2017a,
Ke_2017_CVPR}.
}
We choose to encode the temporal dimension as the
vertical axis, the joints as the horizontal axis, and the coordinates of each
point ($(x,y)$ for 2D, $(x,y,z)$ for 3D) as the channels.
With this approach, we can use classical 2D convolutions to extract patterns
directly from the temporal sequence of body joints.
The predicted coordinates of each body joints are pondered by their confidence
scores, thus points that are not present in the image (and consequently cannot
be correctly predicted) have less influence on action recognition.  A graphical
representation of pose features is presented in Fig.~\ref{fig:pose-features}.

\subsubsection{Appearance Features}
\label{sec:appearance-features}

\rev{In addition to the pose information, visual cues are very important
to action recognition, since they bring contextual information. In our method,
localized visual information is encoded as \textit{appearance features}, which
are extracted in} a similar process to the one of pose
features, with the difference that the first relies on local visual information
instead of joint coordinates.
In order to extract localized appearance features, we multiply each channel
from the tensor of multi-task features
$\zfeat{}_t^{p,l} \in \mathbb{R}^{H_f\times{W_f}\times{N_f}}$ by each channel
from the probability maps
$\probmap{}_t \in \mathbb{R}^{H_f\times{W_f}\times{\njoints{}}}$ (outer product of $N_f$ and $\njoints{}$), which is
learned as a byproduct of the pose estimation process.
Then, the spatial dimensions are collapsed by a sum, resulting in the
appearance features for time $t$ of size $\mathbb{R}^{\njoints{}\times{N_f}}$.
For a sequence of frames, we concatenate each appearance feature map for
$t=\{1,2,\dots,T\}$ resulting in the video clip appearance features
$\vfeat{} \in \mathbb{R}^{T\times\njoints{}\times{N_f}}$.
To clarify this process, a graphical
representation is shown in Fig.~\ref{fig:appearance-features}.

We argue that our multi-task framework has two benefits for the appearance
based part: First, it is computationally very efficient since most part of
the computations are shared. Second, the extracted visual features are more
robust since they are trained simultaneously for different but related tasks
and on different datasets.

\subsubsection{Action Features Aggregation and Re-injection}

Some actions are hard to be distinguished from others only by the high level
pose representation. For example, the actions \textit{drink water} and
\textit{make a phone call} are very similar if we take into account only the
body joints, but are easily separated if we have the visual information
corresponding to the objects cup and phone.
On the other hand, other actions are not directly related to visual information
but with body movements, like \textit{salute} and \textit{touch chest}, and in
this case the pose information can provide complementary information.
\rev{In our method, we combine visual cues and body movements by aggregating
pose and appearance features. This aggregation is a straightforward process,
since both feature types have the same spacial dimensions.}

Similarly to the single frame features re-injection mechanism discussed in
section~\ref{sec:pose-re-inject}, our approach also allows action features
re-injection, as detailed in the action prediction part in
Fig.~\ref{fig:prediction-block}.
We demonstrate in the experiments that this
technique also improves action recognition results with no additional
parameters.

\subsubsection{Decoupled Action Poses}
\label{sec:decoupled}

\begin{figure}[htbp]
  \centering
  \includegraphics[scale=1.0]{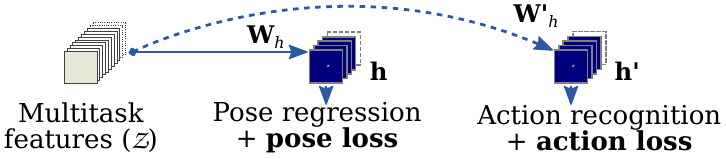}
  \caption[Decoupled poses for action prediction]{
    Decoupled poses for action prediction. The weight matrix
    $\mathbf{W}^\prime_h$ is initialized with a copy of $\mathbf{W}_h$ after
    \rev{the main training process}.  The same is done to
    depth maps ($\mathbf{W}_d$ and $\mathbf{d}$).
  }
  \label{fig:decoupled-poses}
\end{figure}

Since the multi-task architecture is trained simultaneously on pose estimation
and on action recognition, we may have an effect of competing gradients from
poses and actions, specially in the predicted poses, which are used as the output
for the first task and as the input for the second task.
To mitigate that influence, \rev{late in the training process,} we propose to decouple estimated poses (used to
compute pose scores) from action poses (used by the action recognition part) as
illustrated in Fig.~\ref{fig:decoupled-poses}.

Specifically, we first train the \rev{network on pose estimation for about one
half of the full training iterations}, then we replicate only the last layers
that project the multi-task feature map $\zfeat{}$ to heat maps and depth maps
(parameters $\textbf{W}_h$ and $\textbf{W}_d$),
resulting in a ``copy'' of probability maps $\probmap{}^\prime$
and depth maps $\depthmap{}^\prime$.
Note that this replica corresponds to a simple $1\times1$ convolution from the
feature space to the number of joints, which is almost insignificant in terms of
parameters and computations. \rev{The ``copy'' of this layer is a new convolutional layer with its weights $\textbf{W}^\prime$ initialized with $\textbf{W}$}.
Finally, \rev{for the remaining training, the action recognition part propagates its loss through the replica poses.}
This process allows the original pose predictions to stay specialized on
the first task, while the replicated poses absorb partially the action
gradients and are optimized accordingly to the action recognition task.
Despite the replicated poses not being directly supervised in the final
training stage (which corresponds to a few more epochs), we show in our
experiments that they still remain coherent with supervised estimated poses.

\section{Experiments}
\label{sec:experiments}

In this section, we present quantitative and qualitative results by evaluating
the proposed method on two different tasks and on two different modalities:
human pose estimation and human action recognition on 2D and 3D scenarios.
\revb{Since our method relies on body coordinates, we consider four publicly
available datasets mostly composed of full poses, which are} detailed as follows.

\subsection{Datasets}

\textbf{MPII} Human Pose Dataset~\cite{Andriluka_CVPR_2014} is a well known 2D
human pose dataset composed of about 25K images collected from YouTube videos.
2D poses were manually annotated with up to 16 body joints.
\textbf{Human3.6M}~\cite{h36m_pami} is a 3D human pose dataset composed by
videos with 11 subjects performing 17 different activities, all recorded
simultaneously by 4 cameras. High precision 3D poses were captured by a MoCap
system, from which 17 body joints are used for evaluation.
\textbf{Penn Action}~\cite{Zhang_ICCV_2013} is a 2D dataset for action
recognition composed by 2,326 videos with sports people performing 15 different
actions.  Human poses were manually annotated with up to 13 body joints.
\textbf{NTU RGB+D}~\cite{Shahroudy_2016_CVPR} is a large scale 3D action
recognition dataset composed by 56K videos in Full HD with 60 actions performed
by 40 different actors and recorded by 3 cameras in 17 different
configurations.  Each color video has an associated depth map video and 3D
Kinect poses.

\subsubsection{Evaluation Metrics}

On 2D pose estimation, we evaluate our method on the MPII validation set
composed of 3K images, using the probability of correct keypoints measure with
respect to the head size (PCKh)~\cite{Andriluka_CVPR_2014}.
On 3D pose estimation, we evaluate our method on Human3.6M by measuring the
mean per joint position error (MPJPE) after alignment of the root joint.
We follow the most common evaluation protocol~\cite{Yang_2018_CVPR,
Sun_2017_ICCV, Martinez_2017, MehtaRCSXT16, Pavlakos_2017_CVPR} by taking five
subjects for training (S1, S5, S6, S7, S8) and evaluating on two subjects (S9,
S11) on one every 64 frames. We use ground truth person bounding boxes for a
fair comparison with previous methods on single person pose estimation.
We report results using a single cropped bounding box per sample.

On action recognition, we report results using the percentage of correct action
classification score.
We use the proposed evaluation protocol for Penn Action~\cite{Nie_2015_CVPR},
splitting the data as 50/50 for training/testing, \rev{and the
more realistic cross-subject scenario for NTU}, on which
20 subjects are used for training, and the remaining are used for testing.
Our method is evaluated on \textit{single-clip} and/or \textit{multi-clip}.
In the first case, we crop a single clip with $T$ frames in the middle of the
video. In the second case, we crop multiple video clips temporally spaced of
$T/2$ frames one from another, and the final predicted action is the average
decision among all clips from one video.

In our experiments, we consider two scenarios: A) 2D pose estimation and action
recognition, on which we use respectively MPII and Penn Action datasets, and
B) 3D pose estimation and action recognition, using MPII, Human3.6M, and NTU
datasets.

\subsection{Implementation and Training Details}

\subsubsection{Function Loss}
For the pose estimation task, we train the network using the elastic net
loss~\cite{Zou05regularizationand} function on predicted poses:
\begin{equation}
  \mathcal{L}_\pose{} = \frac{1}{\njoints{}}%
    \sum_{j=1}^{\njoints{}}\big(%
    \|\hat{\pose{}}^j-\pose{}^j\|_1 +%
    \|\hat{\pose{}}^j-\pose{}^j \|_2^2\big),
    \label{eq:l1l2-loss}
\end{equation}
where $\hat{\pose{}}^j$ and $\pose{}^j$ are respectively the estimated and the
ground truth positions of the j$th$ body joint. The same loss is used for
both 2D and 3D cases, but only available values
($(x,y)$ for 2D and $(x,y,z)$ for 3D)
are taken into account for backpropagation, depending on the dataset.
We use poses in the camera coordinate system, with $(x,y)$ laying on the image
plane and $z$ corresponding to the depth distance, normalized in the interval
$[0,1]$, where the top-left image corner corresponds to $(0,0)$, and the
bottom-right image corner corresponds to $(1,1)$.
For depth normalization, the root joint is assumed to have $z=0.5$, and a range
of 2 meters is used to represent the remaining joints.
If a given body joint falls outside the cropped bounding box on training, we
set the ground truth confidence flag $\confidence{}^j$ to zero, otherwise we
set it to one.
The ground truth confidence information is used to supervise predicted joint
confidence scores $\hat{\confidence{}}$ with the binary cross entropy loss.
Despite giving an additional information, the supervision on confidence scores
has negligible influence on the precision of estimated poses.
For the action recognition part, we use categorical cross entropy loss on
predicted actions.

\subsubsection{Network Architecture}

Since the pose estimation part is the most computationally expensive, we chose
to use separable convolutions with kernel size equals to $5\times5$ for single
frame layers and standard convolutions with kernel size equals to $3\times3$
for video clip processing layers (action recognition layers).
We performed experiments with the network architecture using 4 levels and
up to 8 pyramids ($L=4$ and $P=8$).
No further significant improvement was noticed on pose estimation by using more
than 8 pyramids.
On action recognition, this limit was observed at 4 pyramids.
For that reason, when using the full model with 8 pyramids, the action
recognition part starts only at the 5$th$ pyramid, reducing the computational
load.
In our experiments, we used normalized RGB images of size $256\times256\times3$
as input, which are reduced to a feature map of size $32\times32\times288$ by
the entry flow network, corresponding to level $l=1$.
At each level, the spatial resolution is reduced by a factor of 2 and the
size of features is arithmetically increased by $96$. For action recognition,
we used $N_v=160$ and $N_v=192$ features for Penn Action and NTU, respectively.

\subsubsection{Multi-task Training}

For all the experiments, we first initialize the network by training pose
estimation only, for about 32k iterations with mini batches of 32 images
(equivalent to 40 epochs on MPII).
Then, all the weights related to pose estimation are fixed and only the action
recognition part is trained for 2 and 50 epochs, respectively for Penn Action
and NTU datasets.
Finally, the full network is trained in a multi-task scenario, simultaneously
for pose estimation and action recognition, until the validation scores
plateau.
Training the network on pose estimation for a few epochs provides a good
general initialization and a better convergence of the action recognition part.
The intermediate training stage of action recognition has two objectives:
first, it is useful to allow a good initialization of the action part,
since it is built on top of the pre-initialized pose estimator; and second, it
is about 3 times faster than performing multi-task training directly while
resulting in similar scores. This process is specially useful for NTU, due to
the large amount of training data.
The training procedure takes about one day for the pose estimation
initialization, then two/three days for the remaining process for Penn
Action/NTU, using a desktop GeForce GTX 1080Ti GPU.

For initialization on pose estimation, the network was optimized with RMSprop
and initial learning rate of 0.001. For action and multi-task training, we use
RMSprop for Penn Action with learning rate reduced by a factor of 0.1 after
15 and 25 epochs, and, for NTU, a vanilla SGD with Nesterov momentum of 0.9 and
initial learning rate of 0.01, reduced by a factor of 0.1 after 50 and 55
epochs.
We weight the loss on body joint confidence scores and action estimations by a
factor of 0.01, since the gradients from the cross entropy loss are much stronger
than the gradients from the elastic net loss on pose estimation.
\revb{This parameter was empirically chosen and we did not observe a significant variation in
the results with slightly different values (e.g., with 0.02)}.
Each iteration
is performed on 4 batches of 8 frames, composed of random images for pose
estimation and video clips for action.
We train the model by alternating one batch containing pose estimation samples
only and another batch containing action samples only. This strategy resulted in
slightly better results compared to batches composed of mixed pose and action
samples.
We augment training data by performing
random rotations from $-40^{\circ}$ to $+40^{\circ}$, scaling from $0.7$ to
$1.3$, video \rev{temporal} subsampling by a factor from 3 to 10, random horizontal flipping,
and random color shifting.  On evaluation, we also subsampled Penn Action/NTU
videos by a factor of 6/8, respectively.

\subsection{Evaluation on 3D Pose Estimation}
\label{sec:eval3dpose}

\begin{figure*}[t!]
  \centering
    \includegraphics[width=0.15\textwidth]{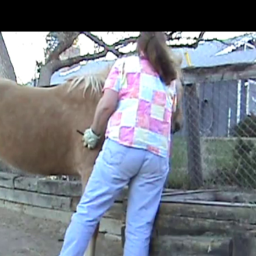}
    \includegraphics[width=0.17\textwidth]{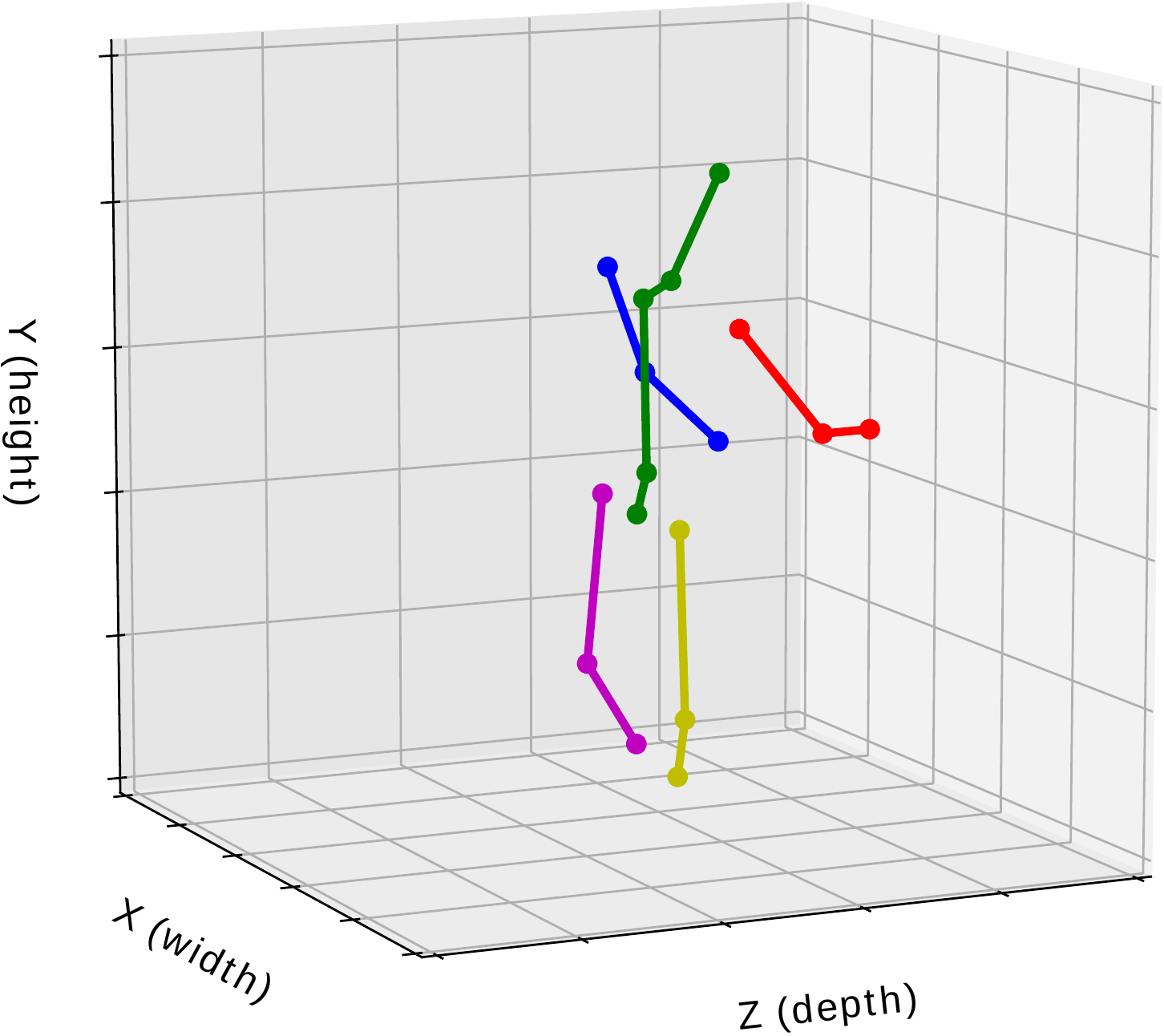}
    \includegraphics[width=0.15\textwidth]{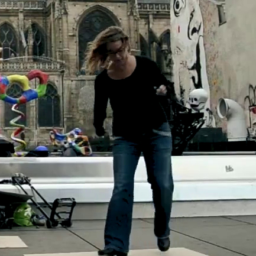}
    \includegraphics[width=0.17\textwidth]{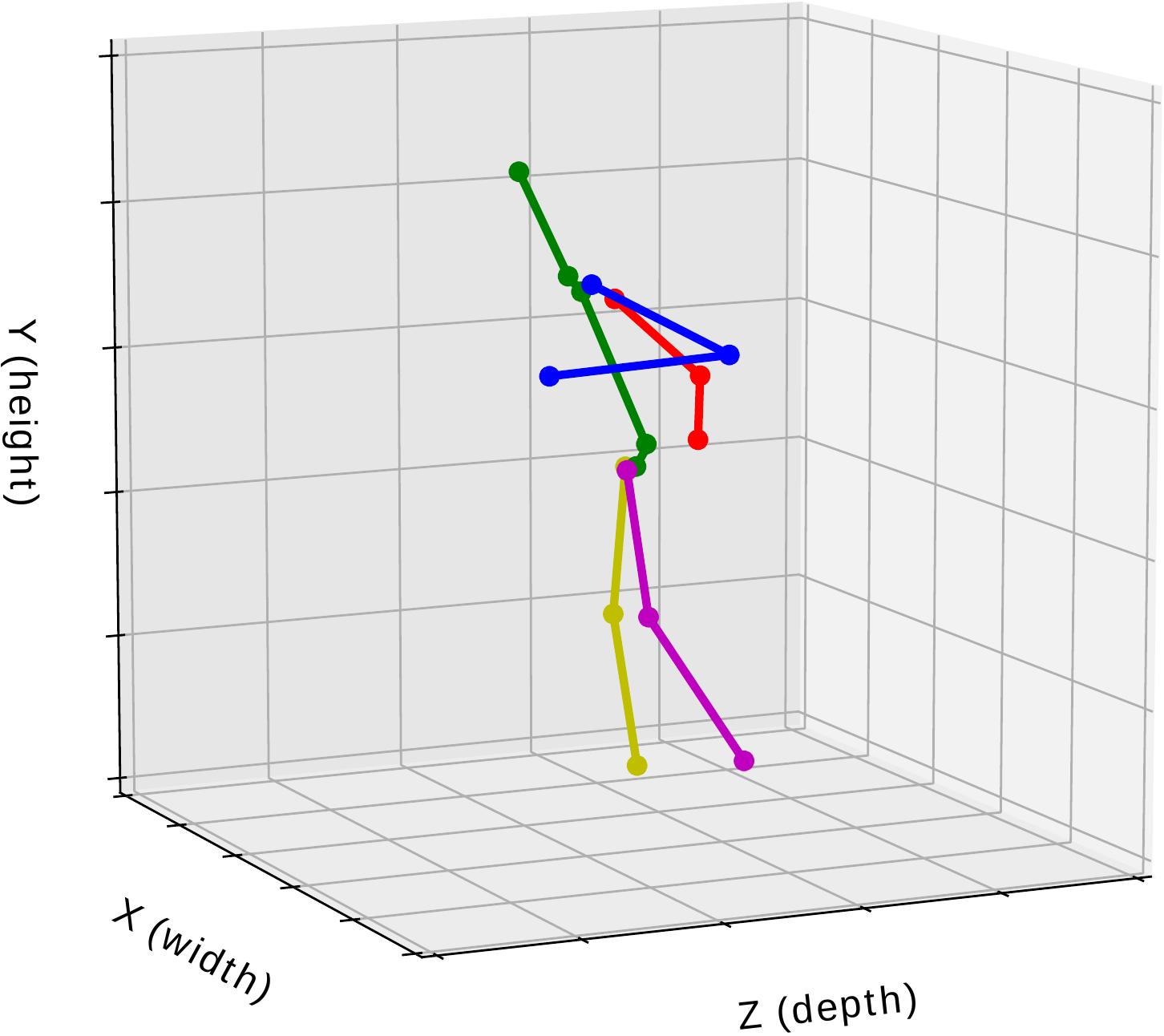}
    \includegraphics[width=0.15\textwidth]{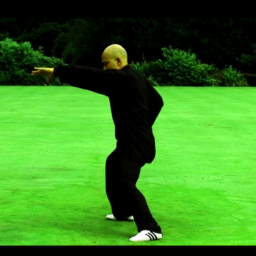}
    \includegraphics[width=0.17\textwidth]{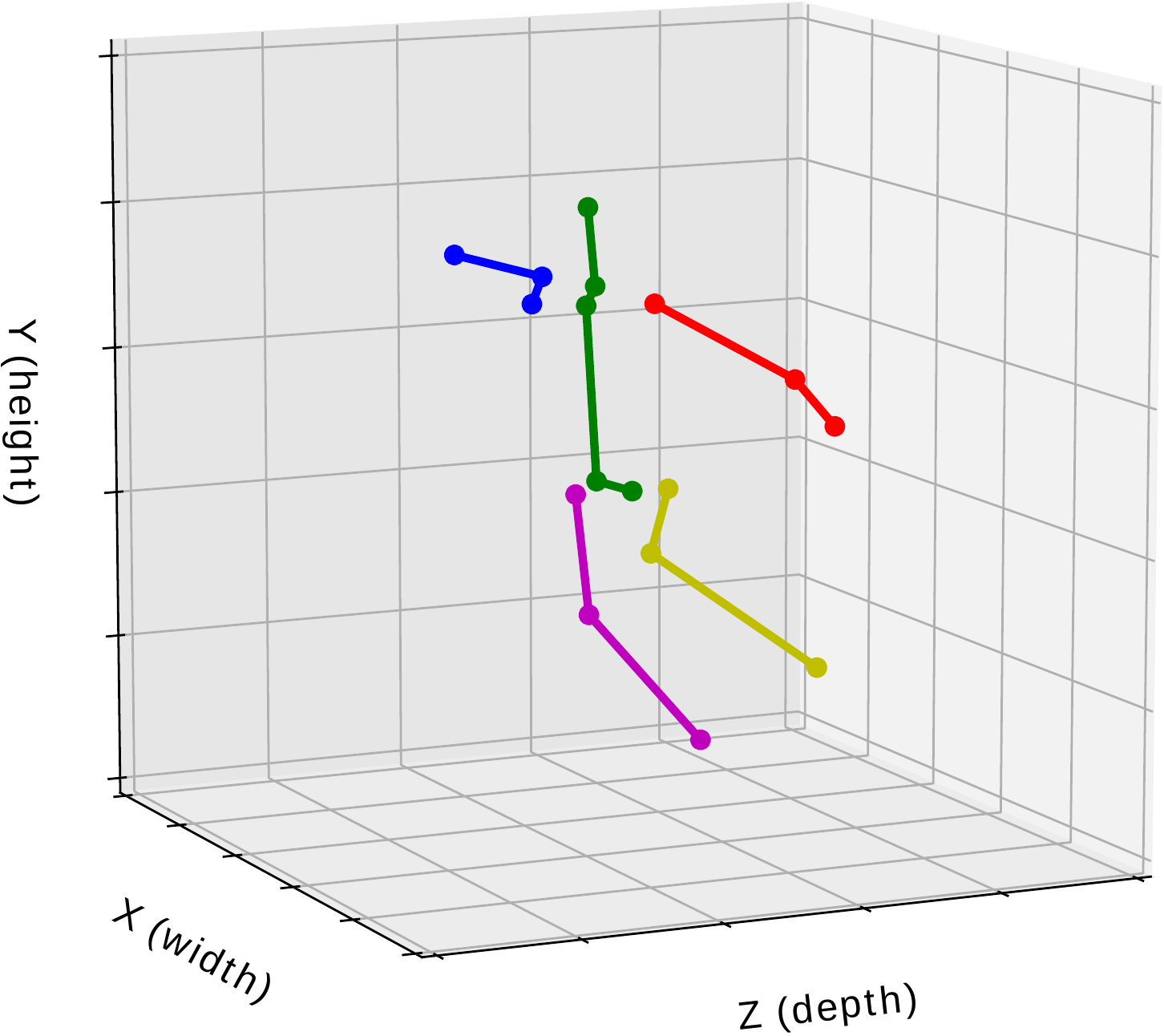}\\
    \includegraphics[width=0.15\textwidth]{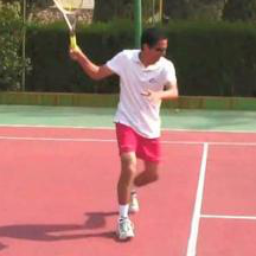}
    \includegraphics[width=0.17\textwidth]{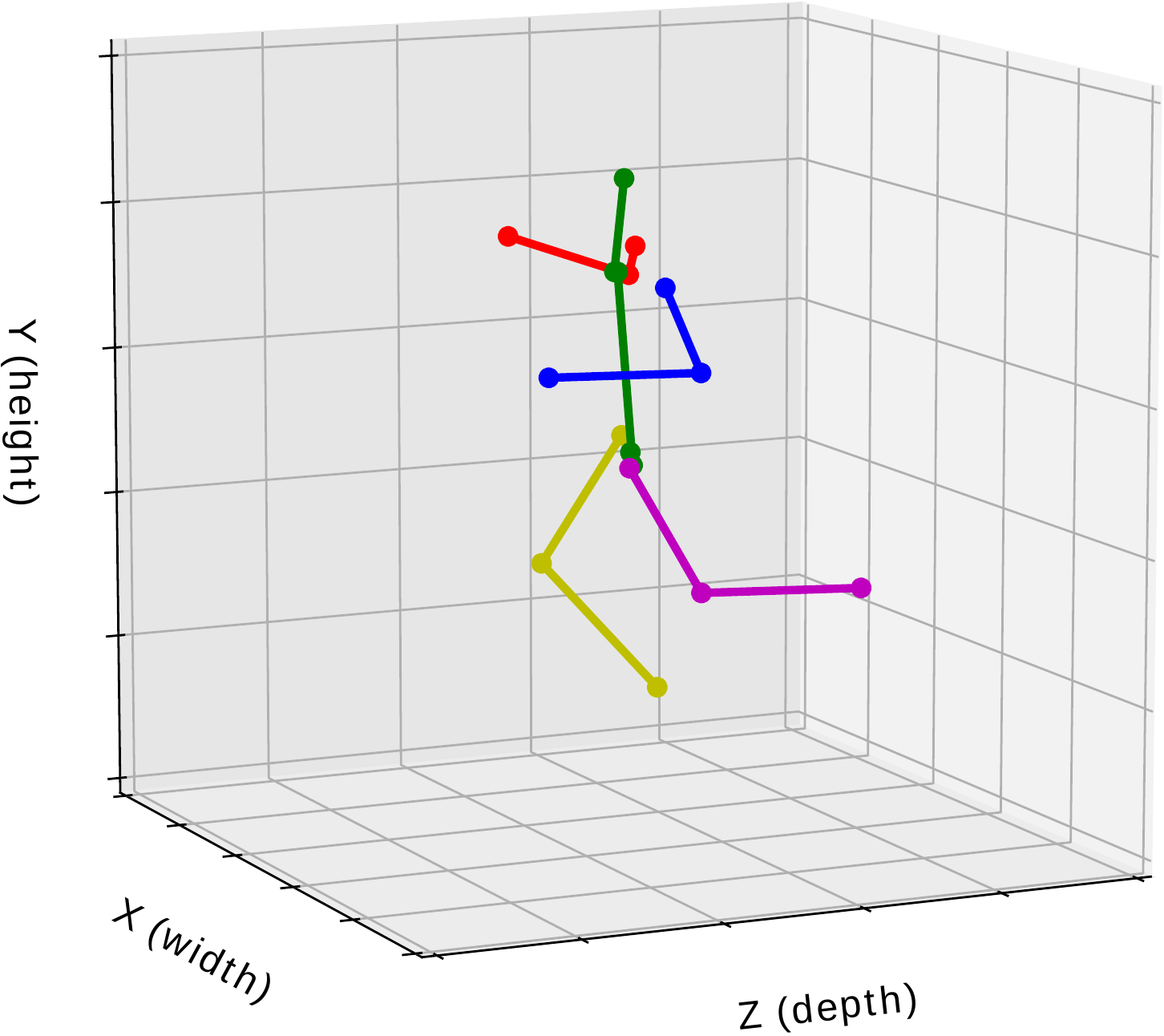}
    \includegraphics[width=0.15\textwidth]{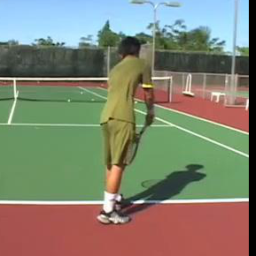}
    \includegraphics[width=0.17\textwidth]{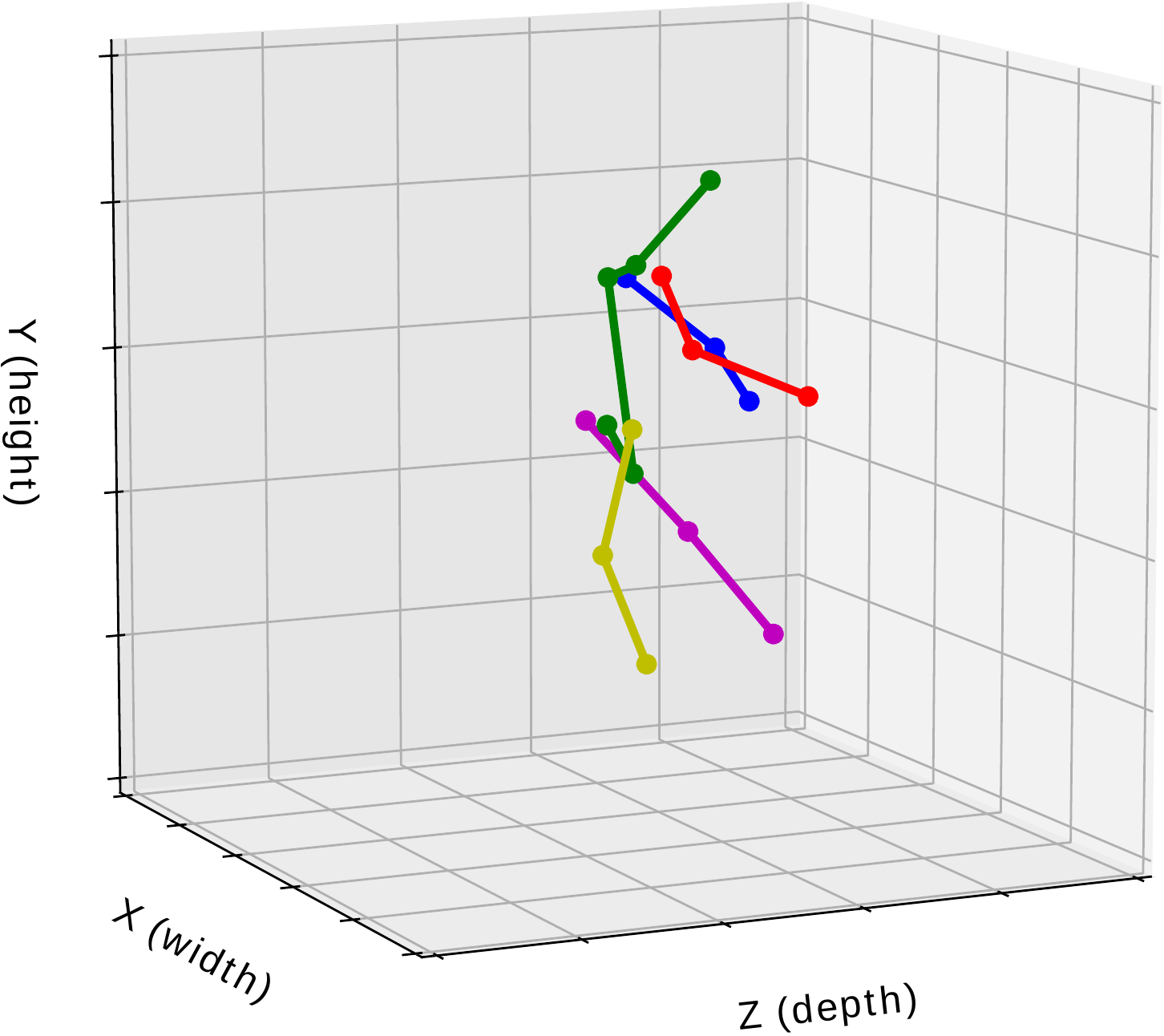}
    \includegraphics[width=0.15\textwidth]{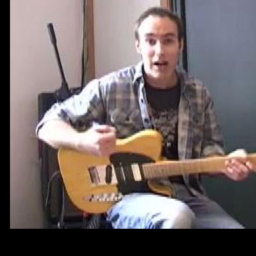}
    \includegraphics[width=0.17\textwidth]{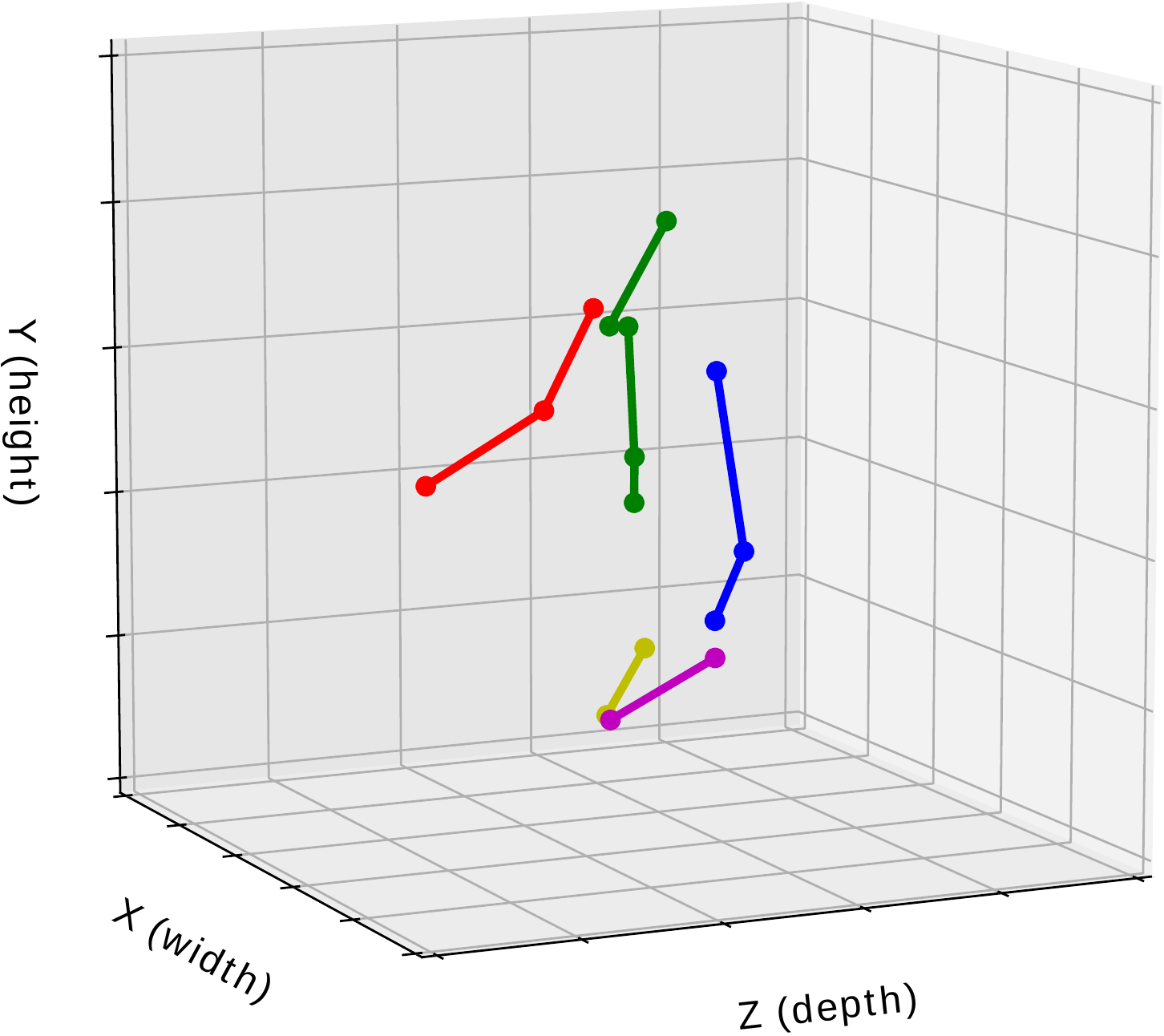}\\
    \includegraphics[width=0.15\textwidth]{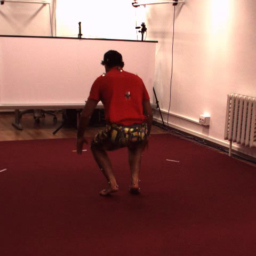}
    \includegraphics[width=0.17\textwidth]{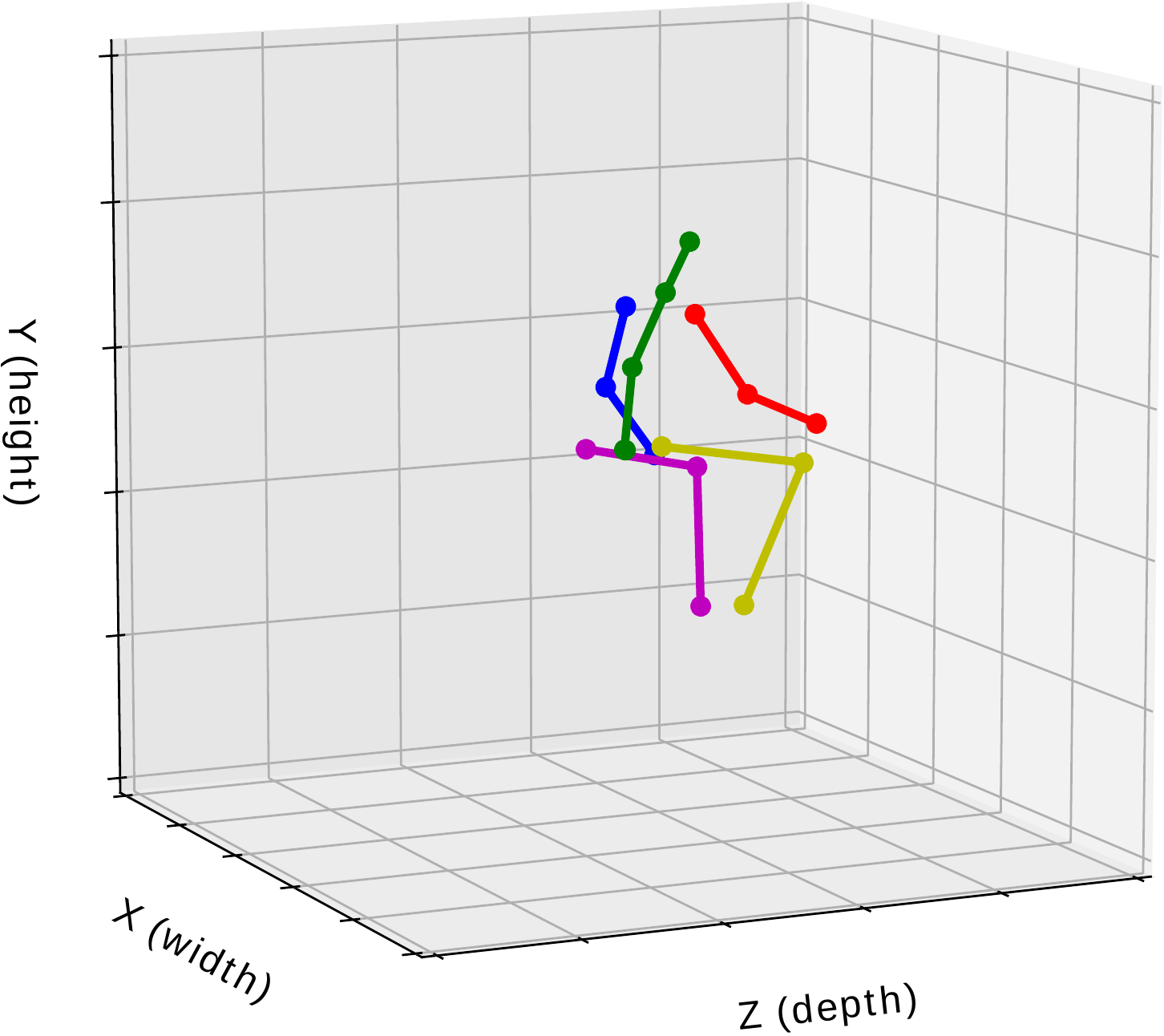}
    \includegraphics[width=0.15\textwidth]{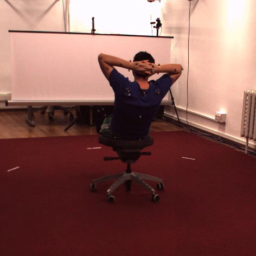}
    \includegraphics[width=0.17\textwidth]{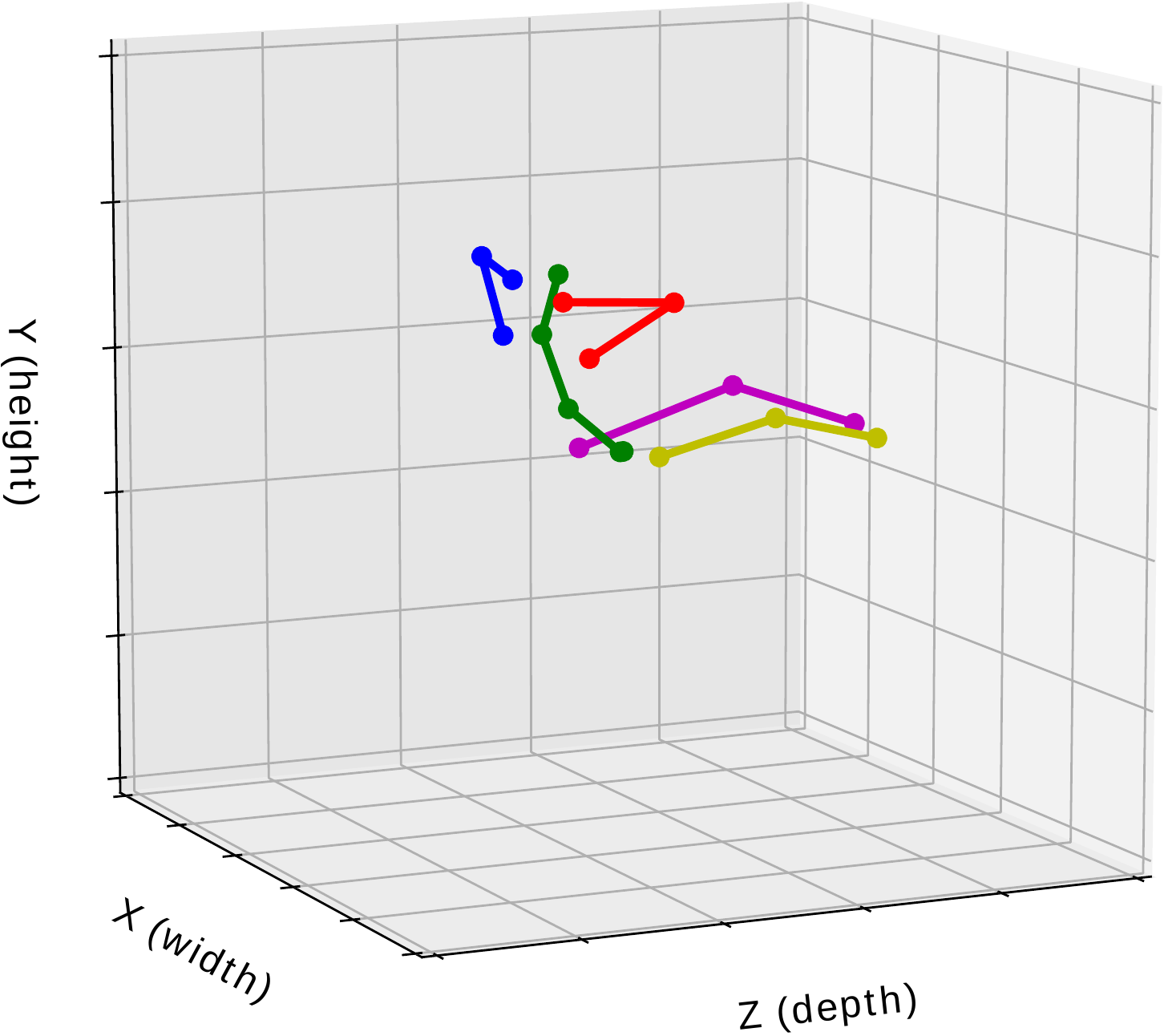}
    \includegraphics[width=0.15\textwidth]{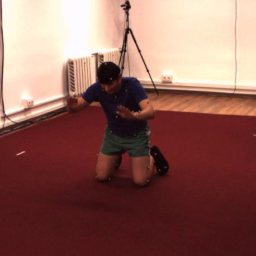}
    \includegraphics[width=0.17\textwidth]{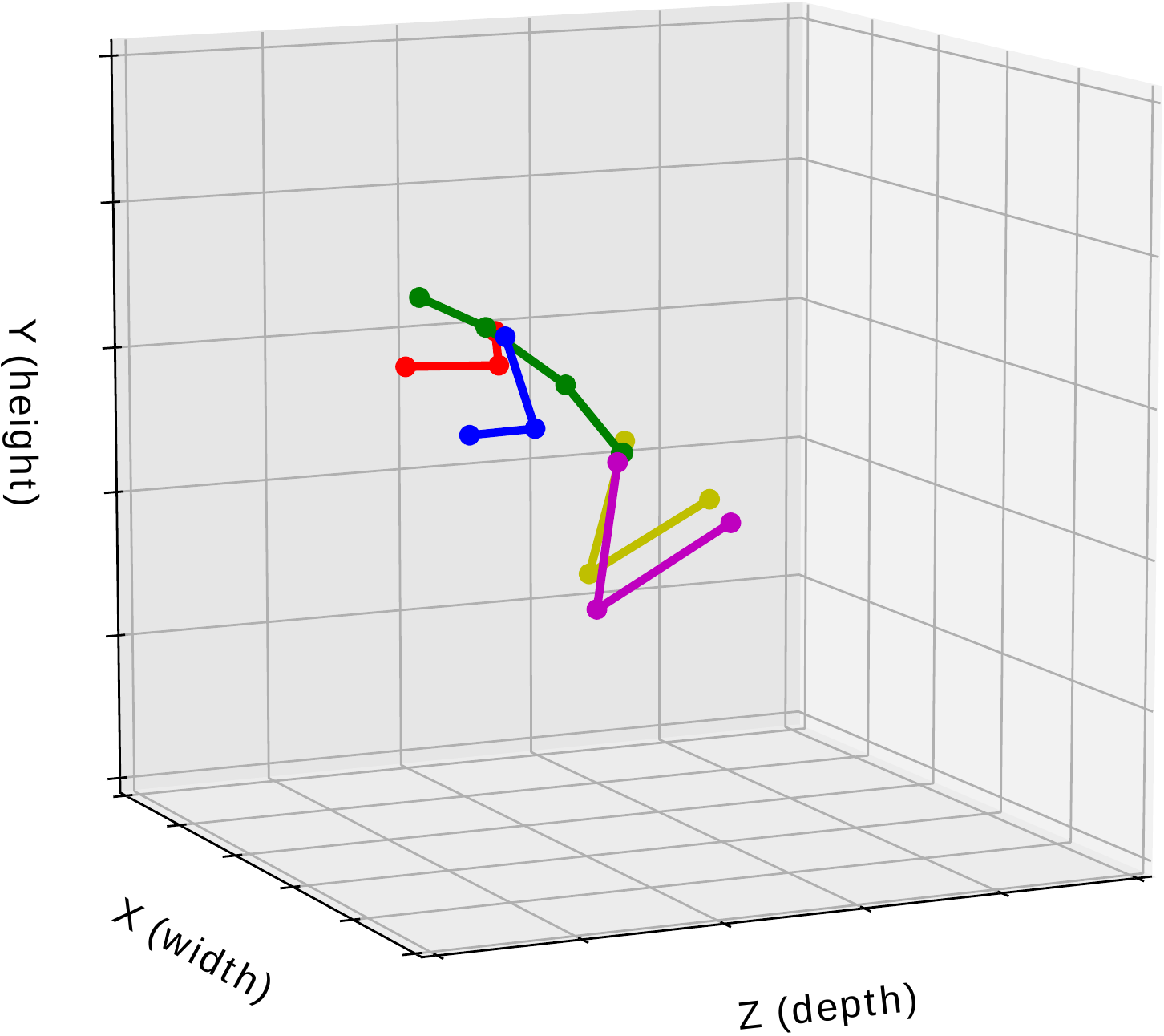}\\
    \includegraphics[width=0.15\textwidth]{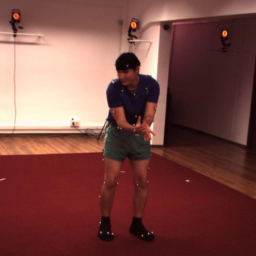}
    \includegraphics[width=0.17\textwidth]{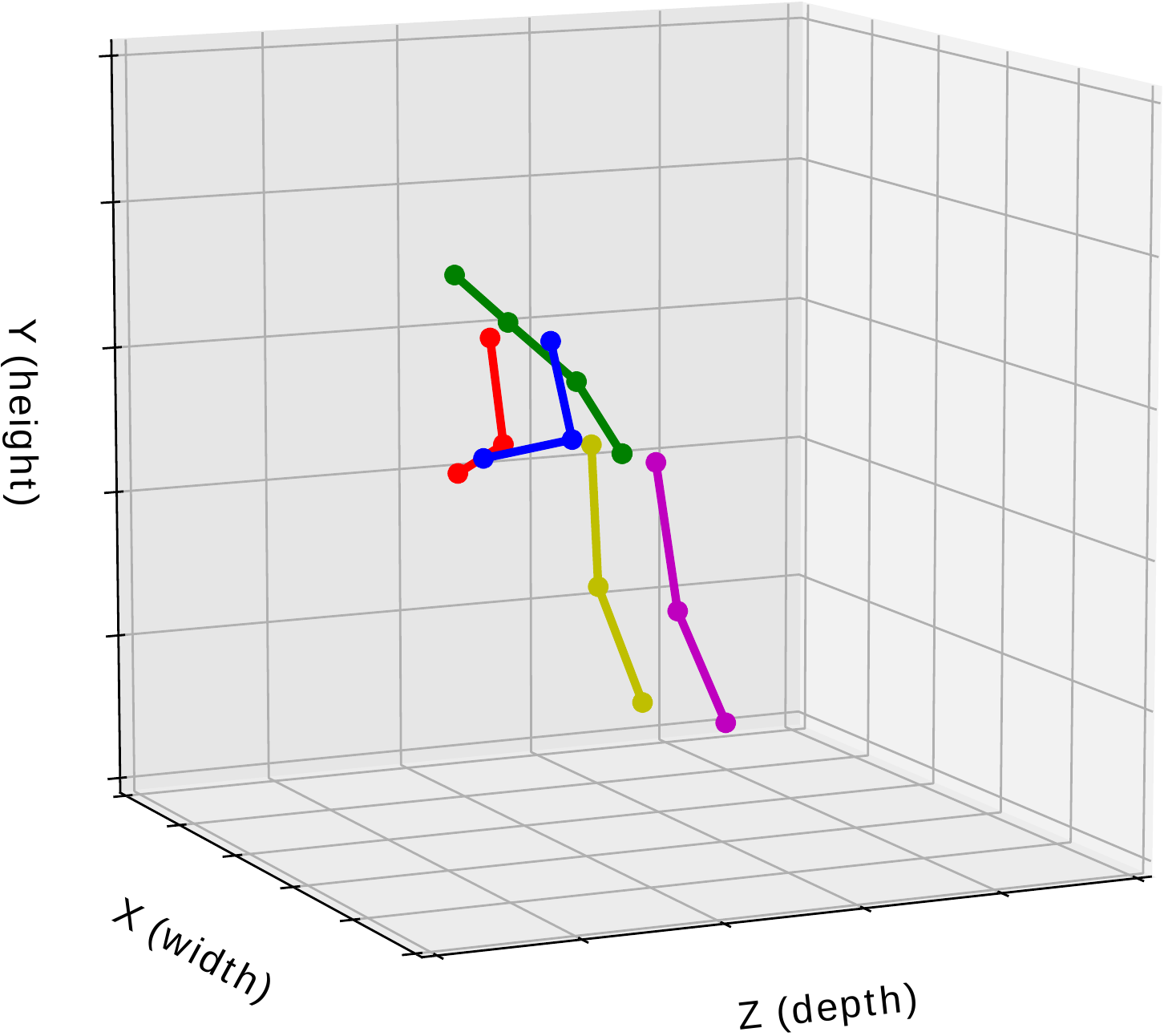}
    \includegraphics[width=0.15\textwidth]{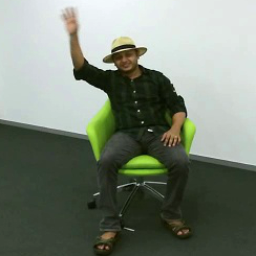}
    \includegraphics[width=0.17\textwidth]{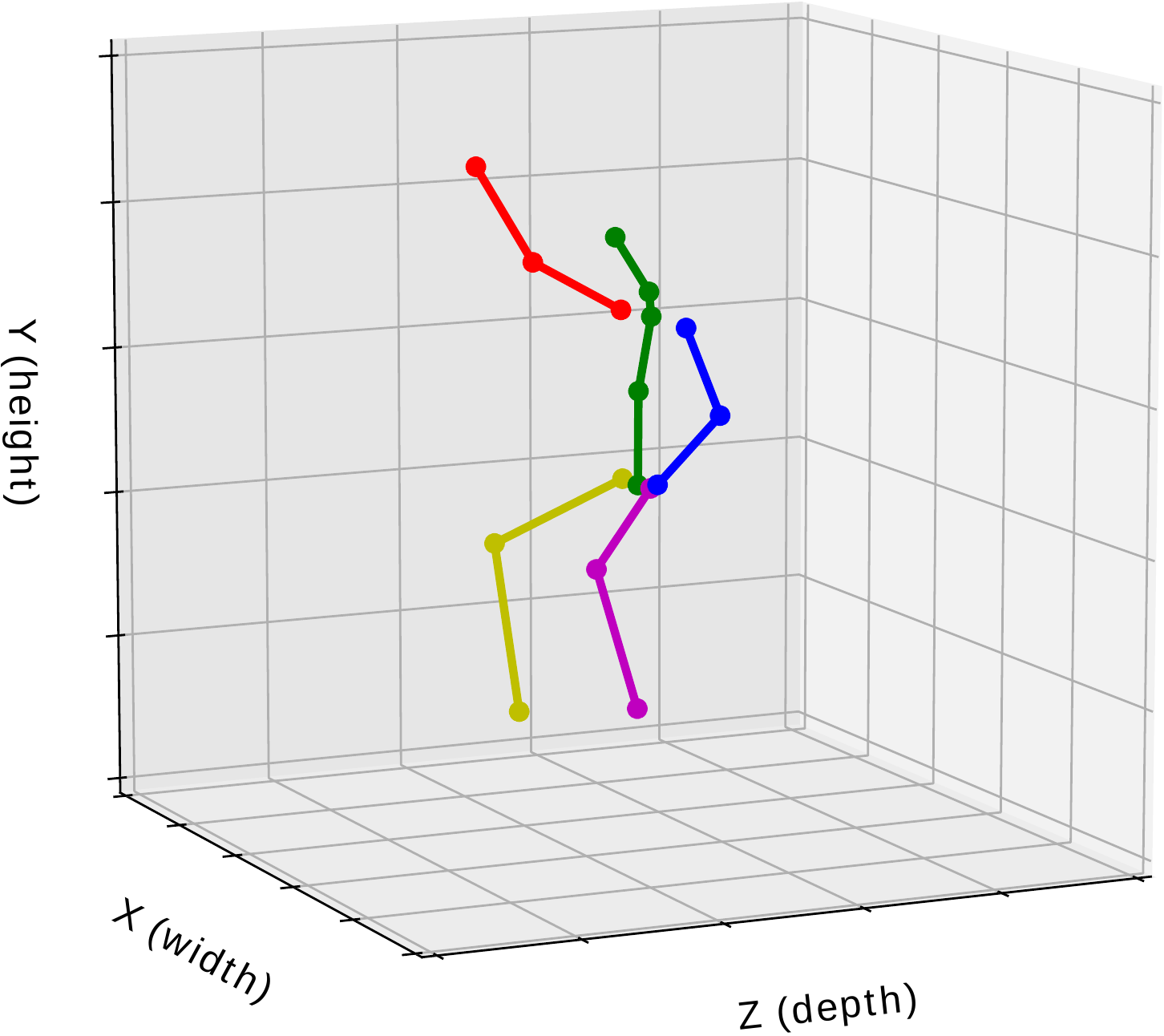}
    \includegraphics[width=0.15\textwidth]{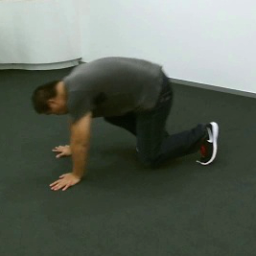}
    \includegraphics[width=0.17\textwidth]{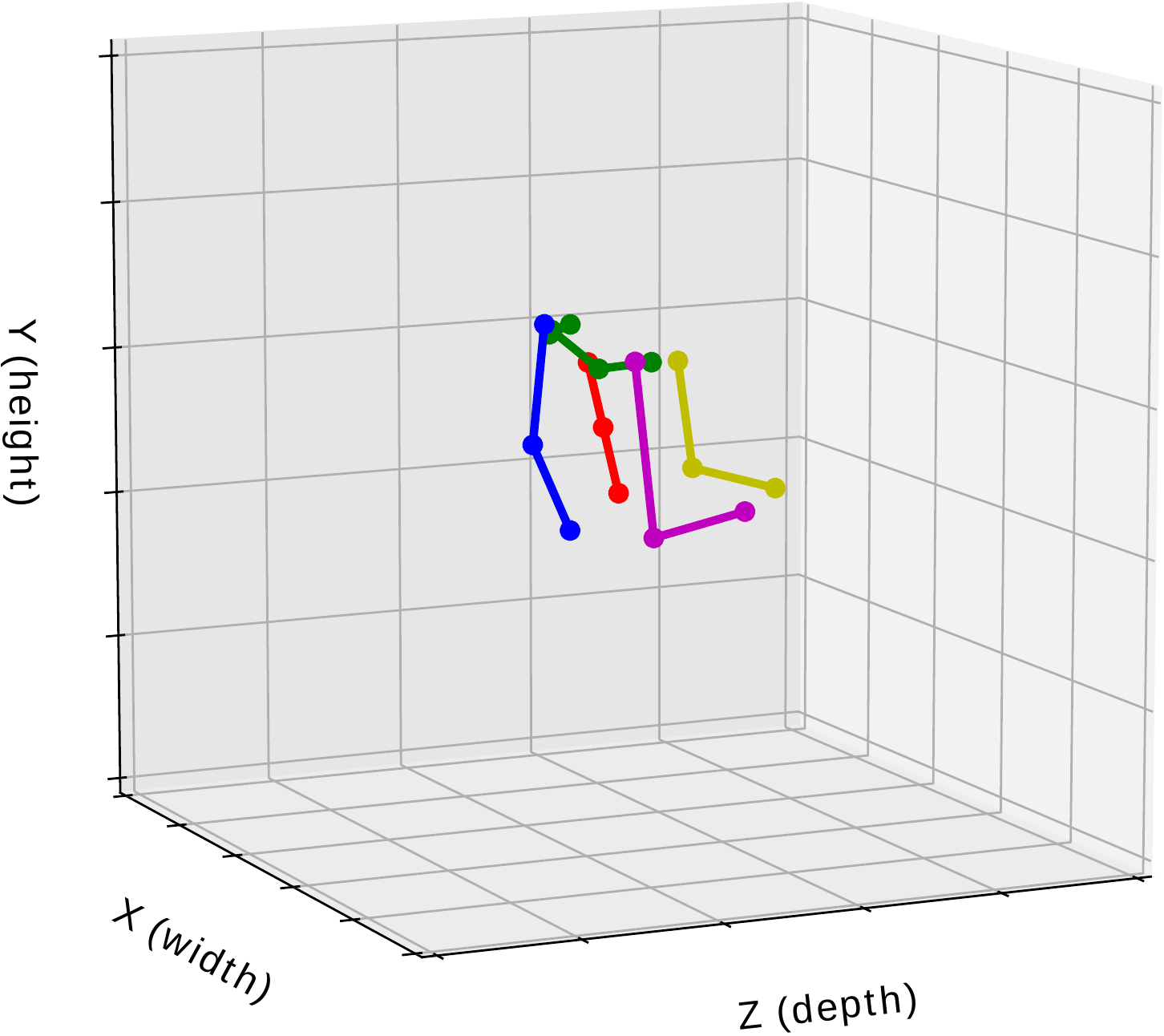}
  \caption{
    Predicted 3D poses from RGB images for both 2D and 3D datasets.
  }
  \label{fig:3dposes}
\end{figure*}

Our results compared to previous approaches are shown in Table~\ref{tab:result-h36m}.
Our multi-task method achieves the state-of-the-art average prediction error of
48.6 millimeters on Human3.6M for 3D pose estimation, improving our
\rev{previous work}~\cite{Luvizon_2018_CVPR} by 4.6 mm. Considering only the pose
estimation task, our average error is 49.5 mm, 0.9 mm higher than the
multi-tasking result, which shows the benefit of multi-task training for 3D pose
estimation.
For the activity ``Sit down'', which is the most challenging case, we
improve previous methods (\eg Yang \etal~\cite{Yang_2018_CVPR}) by 21 mm.
The generalization of our method is demonstrated by qualitative results of 3D
pose estimation for all datasets in Fig.~\ref{fig:3dposes}. 
Note that a single model and a single training procedure was used to produce
all the images and scores, including 3D pose estimation and 3D action
recognition, as discussed in the following.

\begin{table*}[ht]
  \centering
  \caption{Comparison with previous work on Human3.6M evaluated using the
  mean per joint position error (MPJPE, in millimeters) metric on reconstructed
  poses.}
  \label{tab:result-h36m}
  \small
  \begin{tabular}{l|cccccccc}
    \hline
    Methods & Direction & Discuss & Eat & Greet & Phone & Posing & Purchase & Sitting \\ \hline
        \hline
    Pavlakos \etal \cite{Pavlakos_2017_CVPR}    & 67.4 & 71.9 & 66.7 & 69.1 & 71.9 & 65.0 & 68.3 & 83.7 \\
    Mehta \etal \cite{MehtaRCSXT16}\textsuperscript{$\star$} & 52.5 & 63.8 & 55.4 & 62.3 & 71.8 & 52.6 & 72.2 & 86.2 \\
    Martinez \etal \cite{Martinez_2017}         & 51.8 & 56.2 & 58.1 & 59.0 & 69.5 & 55.2 & 58.1 & 74.0 \\
    Sun \etal \cite{Sun_2017_ICCV}\textsuperscript{$\dagger$} & 52.8 & 54.8 & 54.2 & 54.3 & 61.8 & 53.1 & 53.6 & 71.7 \\
    Yang \etal \cite{Yang_2018_CVPR}\textsuperscript{$\dagger$} & 51.5 & 58.9 & 50.4 & 57.0 & 62.1 & 49.8 & 52.7 & 69.2 \\
    Sun \etal \cite{Sun_2018_ECCV}\textsuperscript{$\dagger$} & -- & -- & -- & -- & -- & -- & -- & -- \\
    \hline
    \revb{3D heat maps (ours~\cite{Luvizon_2018_CVPR}, only H36M)} & 61.7 & 63.5 & 56.1 & 60.1 & 60.0 & 57.6 & 64.6 & 75.1 \\
    3D heat maps (ours~\cite{Luvizon_2018_CVPR})\textsuperscript{$\dagger$} & 49.2 & 51.6 & 47.6 & 50.5 & 51.8 & 48.5 & 51.7 & 61.5 \\
    \textbf{Ours} (single-task)\textsuperscript{$\dagger$}                   & \textbf{43.7} & \textbf{48.8} & \textbf{45.6} & \textbf{46.2} & \textbf{49.3} & \textbf{43.5} & \textbf{46.0} & \textbf{56.8} \\
    \textbf{Ours} (multi-task)\textsuperscript{$\dagger$}                   & \textbf{43.2} & \textbf{48.6} & \textbf{44.1} & \textbf{45.9} & \textbf{48.2} & \textbf{43.5} & \textbf{45.5} & \textbf{57.1} \\
    \hline
    Methods & Sit Down & Smoke & Photo & Wait & Walk & Walk Dog & Walk Pair & \multicolumn{1}{|c}{Average} \\ \hline
        \hline
        
    Pavlakos \etal \cite{Pavlakos_2017_CVPR}    & 96.5 & 71.4 & 76.9 & 65.8 & 59.1 & 74.9 & 63.2 & \multicolumn{1}{|c}{71.9} \\
    Mehta \etal \cite{MehtaRCSXT16}\textsuperscript{$\star$} & 120.0& 66.0 & 79.8 & 63.9 & 48.9 & 76.8 & 53.7 & \multicolumn{1}{|c}{68.6} \\
    Martinez \etal \cite{Martinez_2017}         & 94.6 & 62.3 & 78.4 & 59.1 & 49.5 & 65.1 & 52.4 & \multicolumn{1}{|c}{62.9} \\
    Sun \etal \cite{Sun_2017_ICCV} \textsuperscript{$\dagger$}             & 86.7 & 61.5 & 67.2 & 53.4 & 47.1 & 61.6 & 53.4 & \multicolumn{1}{|c}{59.1} \\
    Yang \etal \cite{Yang_2018_CVPR}\textsuperscript{$\dagger$}           & 85.2 & 57.4 & 65.4 & 58.4 & 60.1 & 43.6 & 47.7 & \multicolumn{1}{|c}{58.6} \\
    Sun \etal \cite{Sun_2018_ECCV}\textsuperscript{$\dagger$}              & -- & -- & -- & -- & -- & -- & -- & \multicolumn{1}{|c}{49.6} \\
    \hline
    \revb{3D heat maps (ours~\cite{Luvizon_2018_CVPR}, only H36M)} & 95.4 & 63.4 & 73.3 & 57.0 & 48.2 & 66.8 & 55.1 & \multicolumn{1}{|c}{\textbf{63.8}} \\
    3D heat maps (ours~\cite{Luvizon_2018_CVPR})\textsuperscript{$\dagger$} & 70.9 & 53.7 & 60.3 & 48.9 & 44.4 & 57.9 & 48.9 & \multicolumn{1}{|c}{53.2} \\
    \textbf{Ours} (\rev{single-task})\textsuperscript{$\dagger$}                   & \textbf{67.8} & \textbf{50.5} & \textbf{57.9} & \textbf{43.4} & \textbf{40.5} & \textbf{53.2} & \textbf{45.6} & \multicolumn{1}{|c}{\textbf{49.5}} \\
    \textbf{Ours} (multi-task)\textsuperscript{$\dagger$}                   & \textbf{64.2} & \textbf{50.6} & \textbf{53.8} & \textbf{44.2} & \textbf{40.0} & \textbf{51.1} & \textbf{44.0} & \multicolumn{1}{|c}{\textbf{48.6}} \\\hline
  \end{tabular}\\
  \small\textsuperscript{$\star$} Method not using ground-truth bounding boxes.\\
  \revb{\small\textsuperscript{$\dagger$} Methods using extra 2D data for training.}
\end{table*}

\subsection{Evaluation on Action Recognition}

For action recognition, we evaluate our method considering both 2D and 3D
scenarios. For the first, a single model was trained using MPII for single
frames (pose estimation) and Penn Action for video clips. In the second
scenario, we use Human3.6M for 3D pose supervision, MPII for data augmentation,
and NTU video clips for action. Similarly, a single model was trained for
all the reported 3D pose and action results.

For 2D, the pose estimation was trained using mixed data from MPII (80\%) and
Penn Action (20\%), using 16 body joints.
Results are shown in Table~\ref{tab:pennaction}.
We reached the state-of-the-art
action classification score of 98.7\% on Penn Action, improving our \rev{previous
work}~\cite{Luvizon_2018_CVPR} by 1.3\%. Our method outperformed all previous
methods, including the ones using ground truth (manually annotated) poses.

\begin{table}[h]
  \centering
  \caption[Penn Action results]{Results for action recognition on Penn Action.
  Results are given as the percentage of correctly classified actions.
  \revb{Our method uses extra 2D pose data from MPII for training}.
  }
  \label{tab:pennaction}
  \small
  \begin{tabular}{@{}lccccc@{}}
    \hline
    Methods & RGB & \begin{tabular}[c]{@{}c@{}}\footnotesize Optical\\Flow\end{tabular} & \begin{tabular}[c]{@{}c@{}}\footnotesize Annot.\\poses\end{tabular} & \begin{tabular}[c]{@{}c@{}}\footnotesize Est.\\poses\end{tabular} & Acc. \\ \hline
      Nie \etal \cite{Nie_2015_CVPR}                                           & \ok & -   & -   & \ok & 85.5 \\
    \multirow{2}{*}{Iqbal \etal \cite{Iqbal_2017}}                             & -   & -   & -   & \ok & 79.0 \\
                                                                               & \ok & \ok & -   & \ok & 92.9 \\
    \multirow{2}{*}{Cao \etal \cite{Cao_2017}}                                 & \ok & -   & \ok & -   & 98.1 \\
                                                                               & \ok & -   & -   & \ok & 95.3 \\
    Du \etal \cite{Du_2017_ICCV}\textsuperscript{$\star$}                      & \ok & \ok & -   & \ok & 97.4 \\
    \multirow{2}{*}{Liu \etal \cite{Liu_2018_CVPR}}\textsuperscript{$\dagger$} & \ok & -   & \ok & -   & 98.2 \\
                                                                               & \ok & -   & -   & \ok & 91.4 \\
    \hline
    \multirow{2}{*}{\footnotesize{Our previous work~\cite{Luvizon_2018_CVPR}}} & \ok & - & \ok & -   & 98.6 \\
                                                                & \ok & - & -   & \ok & 97.4 \\
    \textbf{Ours} (single-clip)                                 & \ok & - & -   & \ok & \textbf{98.2} \\
    \textbf{Ours} (multi-clip)                                  & \ok & - & -   & \ok & \textbf{98.7} \\
    \hline
  \end{tabular} \\
  \textsuperscript{$\star$} Including UCF101 data; \textsuperscript{$\dagger$} using add. deep features.
\end{table}

For 3D, we trained our multi-task network using mixed data from Human3.6M
(50\%), MPII (37.5\%) and NTU (12.5\%) for pose estimation and NTU video clips
for action recognition.
Our results compared to previous methods are presented in Table~\ref{tab:ntu}.
Our approach reached 89.9\% of correctly classified
actions on NTU, which is a strong result considering the hard task of
classifying among 60 different actions in the cross-subject split. Our method
improves previous results by at least 3.3\% and our \rev{previous work} by
4.4\%, which shows the effectiveness of the proposed approach.

\begin{table}[ht]
  \centering
  \caption[NTU results]{Comparison results on NTU cross-subject for 3D action recognition.
  Results are given as the percentage of correctly classified actions.
  \revb{Our method uses extra pose data from MPII and H36M for training}.
  }
  \label{tab:ntu}
  \small
  \begin{tabular}{@{}lcccc@{}}
    \hline
    Methods & RGB & \begin{tabular}[c]{@{}c@{}}\footnotesize Kinect\\poses \end{tabular} & \begin{tabular}[c]{@{}c@{}}\footnotesize Estimated\\poses \end{tabular} & \begin{tabular}[c]{@{}c@{}}Acc. cross\\subject\end{tabular} \\ \hline
      Shahroudy \etal \cite{Shahroudy_2016_CVPR}        & -   & \ok & - & 62.9 \\
    Liu \etal \cite{Liu2016}                            & -   & \ok & - & 69.2 \\
    Song \etal \cite{Song_2017_AAAI}                    & -   & \ok & - & 73.4 \\
    Liu \etal \cite{Liu_2017_CVPR}                      & -   & \ok & - & 74.4 \\
    Shahroudy \etal \cite{Shahroudy2017DeepMF}          & \ok & \ok & - & 74.9 \\
    Liu \etal \cite{Liu_2018_CVPR}                      & \ok & -   & \ok & 78.8 \\
    \multirow{3}{*}{Baradel \etal \cite{baradel2017a}}  & -   & \ok & - & 77.1 \\
                                                        & \ok & \textsuperscript{$\star$} & - & 75.6 \\
                                                        & \ok & \ok & - & 84.8 \\
    Baradel \etal \cite{Baradel_2018_CVPR}              & -   & - & - & 86.6 \\
    \hline
    Our previous work~\cite{Luvizon_2018_CVPR}          & \ok & - & \ok & 85.5 \\
    \textbf{Ours}                                       & \ok & - & \ok & \textbf{89.9} \\
    \hline
  \end{tabular} \\
  \small\textsuperscript{$\star$} Ground truth poses used on test to select visual features.
\end{table}

\subsection{Ablation Study}

\subsubsection{Network Design}
\label{sec:net-design}

We performed several experiments on the proposed network architecture
in order to identify its best arrangement for solving both tasks
with the best performance \textit{vs} computational cost trade-off.
In Table~\ref{tab:abla-network}, we show the results on 2D pose estimation
and on action recognition considering different network layouts.
For example, in the first line, a single PB is used at pyramid 1 and level 2.
In the second line, a pair of full downscaling and upscaling pyramids are used,
but with supervision only at the last PB. This results in 97.5\% of
accuracy on action recognition and 84.2\% on PCKh for pose estimation.
An equivalent network is used in the third line, but then with supervision
on all PB blocks, which brings an improvement of 0.9\% on pose and
0.6\% on action, with the same number of parameters.
\rev{Note that the networks from the second and third lines are exactly the
same, but in the first case, only the last PB is supervised, while in the
latter all PB receive supervision.}
Finally, the last line shows results with the full network, reaching 88.3\% on
MPII and 98.2\% on Penn Action (single-clip), with a single multi-task model.

\begin{table}[htbp]
  \centering
  \caption{The influence of the network architecture on pose estimation and on
  action recognition, evaluated respectively on MPII validation set (PCKh@0.5,
  single-crop) and on Penn Action (classification accuracy, single-clip).
  Single-PB are indexed by pyramid $p$ and level $l$, and $P$ and $L$ represent
  the \rev{total} number of pyramids and levels on Multi-PB scheme.
  }
  \label{tab:abla-network}
  \small
  \begin{tabular}{l|cccc}
    \hline
    Network               & Param. & PB & PCKh & \footnotesize{Action acc.} \\ \hline
    \footnotesize{Single-PB $(p=1,l=2)$} & 2M     & 1      & 74.3 & 97.2 \\ 
    \footnotesize{Single-PB $(p=2,l=1)$} & 10M     & 1      & 84.2 & 97.5 \\ 
    \footnotesize{Multi-PB $(P=2,L=4)$}  & 10M     & 6      & 85.1 & 98.1 \\ 
    \footnotesize{\textbf{Multi-PB $(P=8,L=4)$}}  & \textbf{26M} & \textbf{24} & \textbf{88.3} & \textbf{98.2} \\ 
    \hline
  \end{tabular}
\end{table}

\subsubsection{Pose and Appearance Features}
\label{sec:action-aggregation}

The proposed method benefits from both pose and appearance features, which
are complementary to the action recognition task. Additionally,
the confidence score $\hat{\confidence{}}$ is also complementary to
pose itself and leads to marginal action recognition gains if used to weight
pose predictions.
\rev{Similar results are achieved if confidence scores are concatenated to poses}.
In Table~\ref{tab:abla-features}, we present results on pose estimation and
on action recognition for different features extraction strategies. Considering pose
features or appearance features alone, the results on Penn Action
are respectively 97.4\% and 97.9\%, respectively 0.7\% and 0.2\% lower than
combined features.
We also show in the last row the influence of decoupled action poses, resulting
in a small gain of 0.1\% on action scores and 0.3\% on pose estimation, which
shows that decoupling action poses brings additional improvements,
\rev{specially for pose estimation}.
When not considering decoupled poses, note that the best score on pose
estimation happens when poses are not directly used for action, which also
supports the evidence of competing losses.

\begin{table}[htbp]
  \centering
  \caption{Results with pose and appearance features alone,
  combined pose and appearance features, and decoupled poses.
  Experiments with a Multi-PB network with $P=2$ and $L=4$.}
  \label{tab:abla-features}
  \small
  \begin{tabular}{l|cc}
    \hline
    \footnotesize{Action features} & \footnotesize{MPII val. PCKh} & \footnotesize{PennAction Acc.} \\ \hline
    \footnotesize{Pose features only}         & 84.9 & 97.7 \\ 
    \footnotesize{Appearance features only}   & 85.2 & 97.9 \\ 
    \footnotesize{Combined}                   & 85.1 & 98.1 \\ %
    \footnotesize{\textbf{Combined + decoupled poses}} & \textbf{85.4} & \textbf{98.2} \\ 
    \hline
  \end{tabular}
\end{table}

\begin{figure}[htbp]
  \centering
    \includegraphics[width=0.075\textwidth]{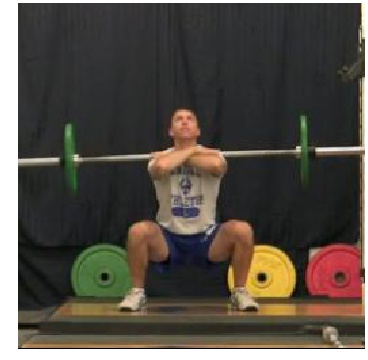}\hspace{-1mm}
    \includegraphics[width=0.075\textwidth]{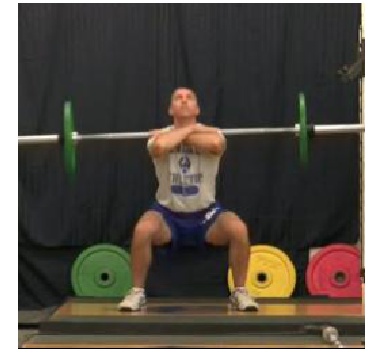}\hspace{-1mm}
    \includegraphics[width=0.075\textwidth]{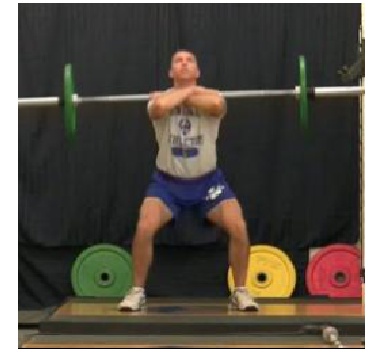}\hspace{1mm}
    \includegraphics[width=0.075\textwidth]{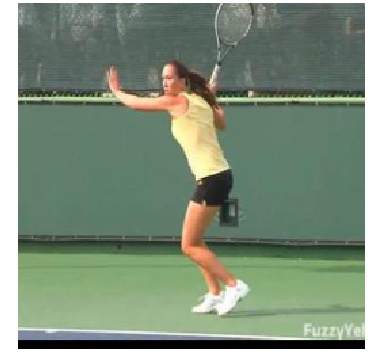}\hspace{-1mm}
    \includegraphics[width=0.075\textwidth]{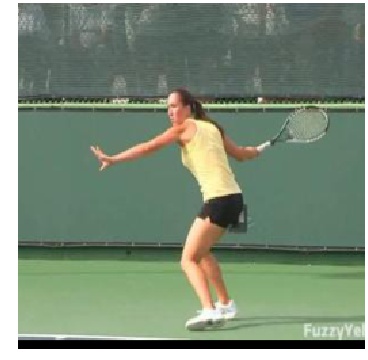}\hspace{-1mm}
    \includegraphics[width=0.075\textwidth]{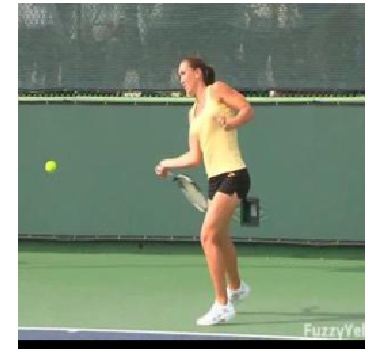}\hspace{-1mm}

    \includegraphics[width=0.075\textwidth]{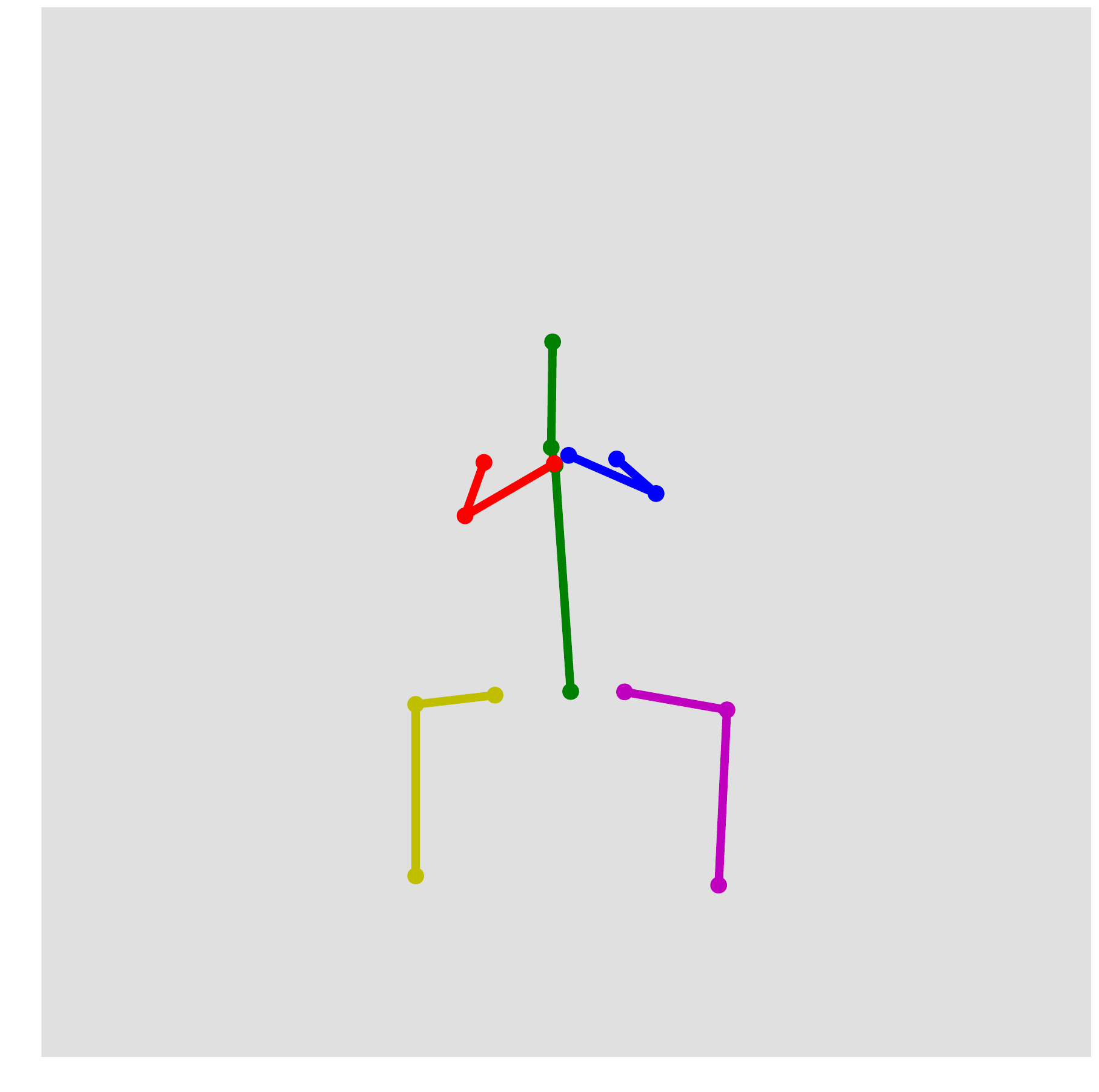}\hspace{-1mm}
    \includegraphics[width=0.075\textwidth]{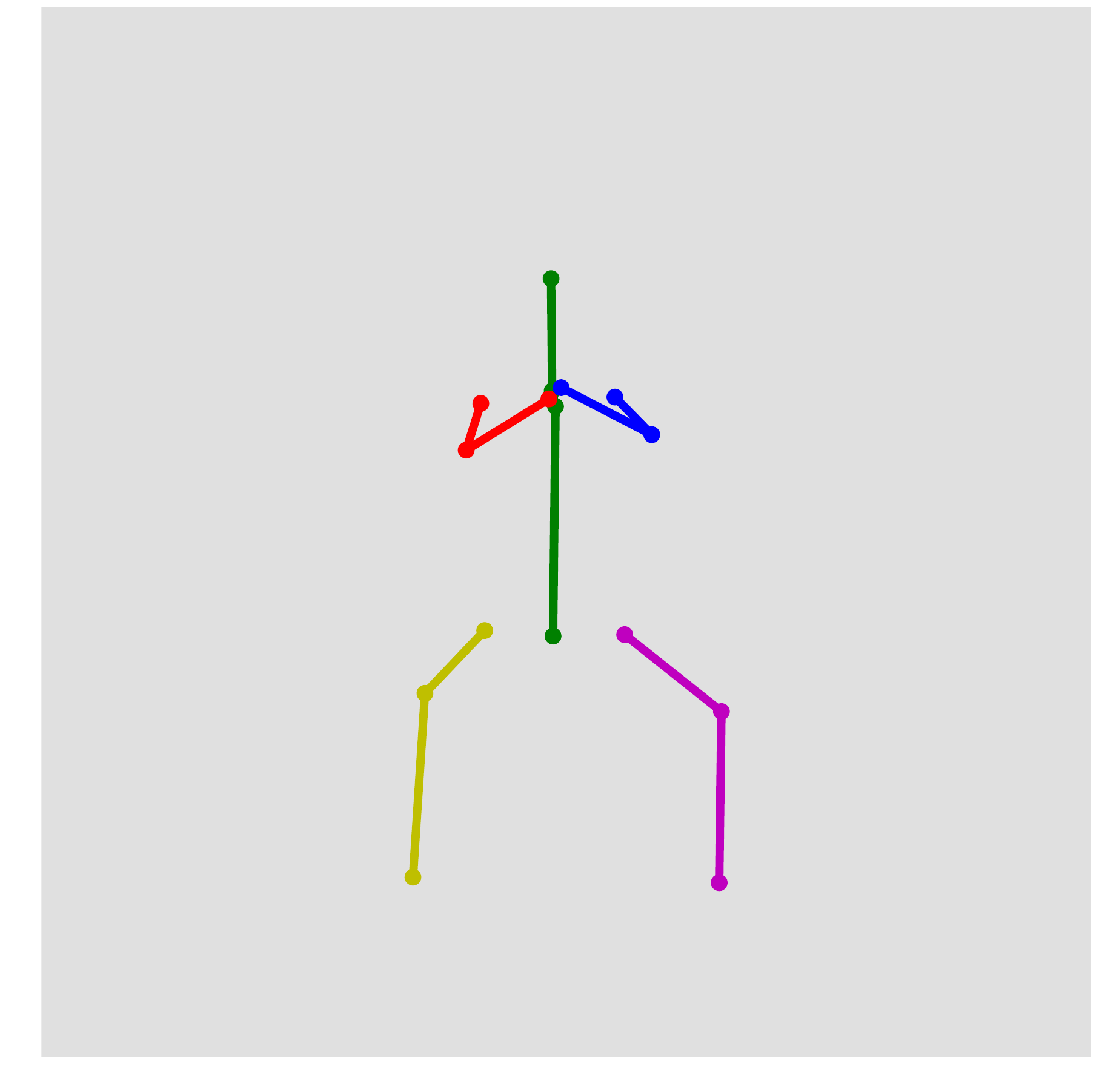}\hspace{-1mm}
    \includegraphics[width=0.075\textwidth]{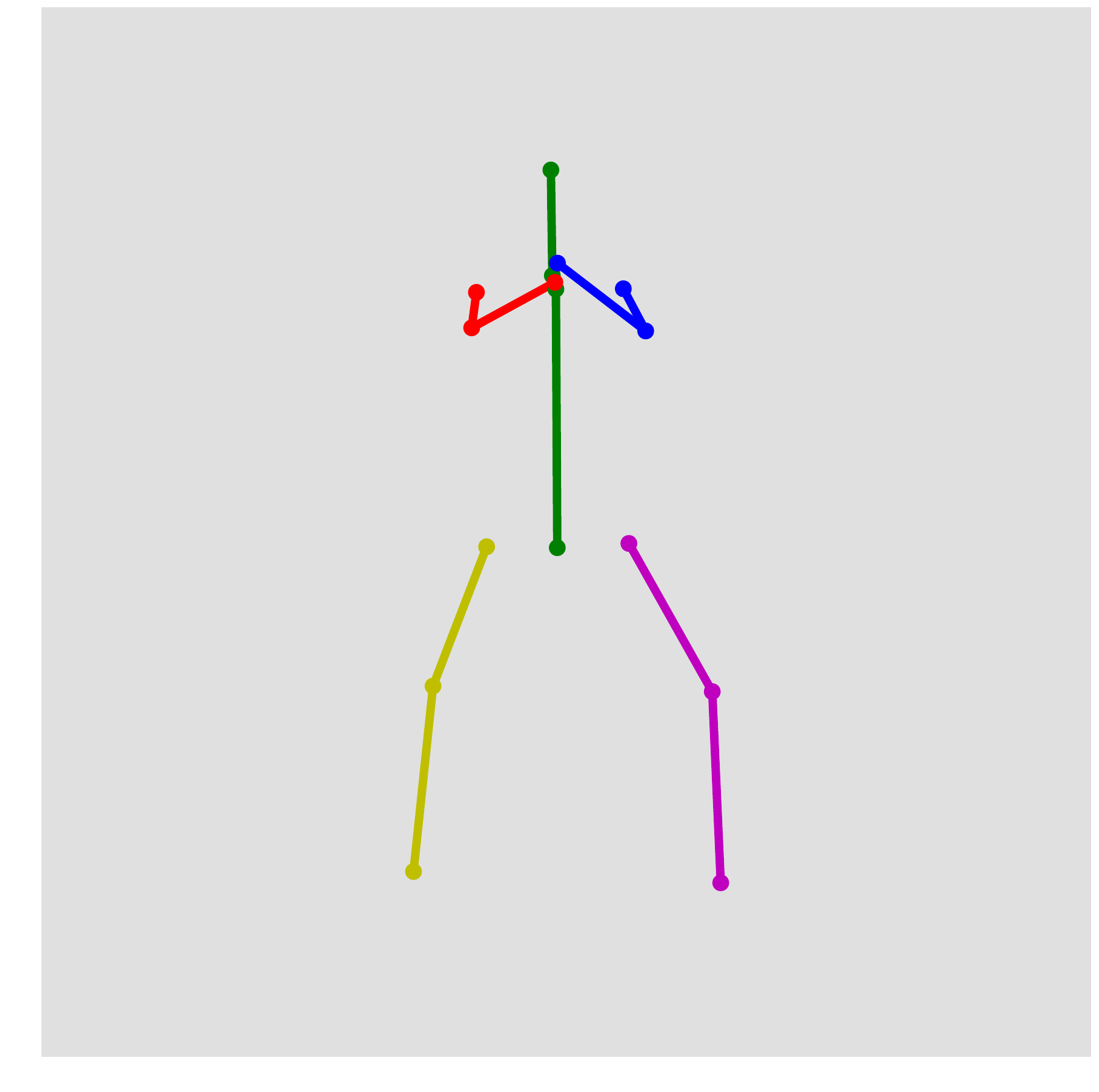}\hspace{1mm}
    \includegraphics[width=0.075\textwidth]{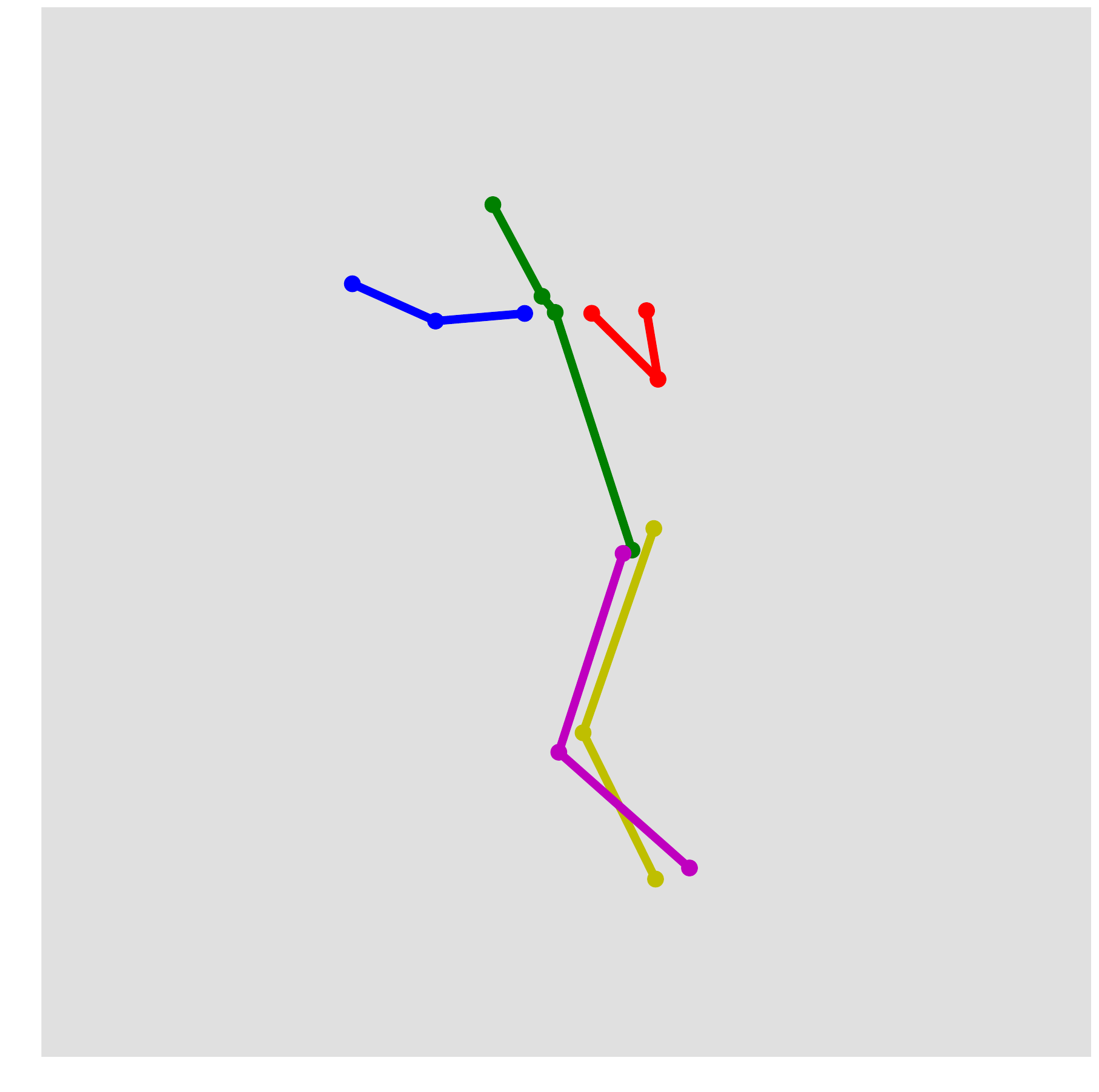}\hspace{-1mm}
    \includegraphics[width=0.075\textwidth]{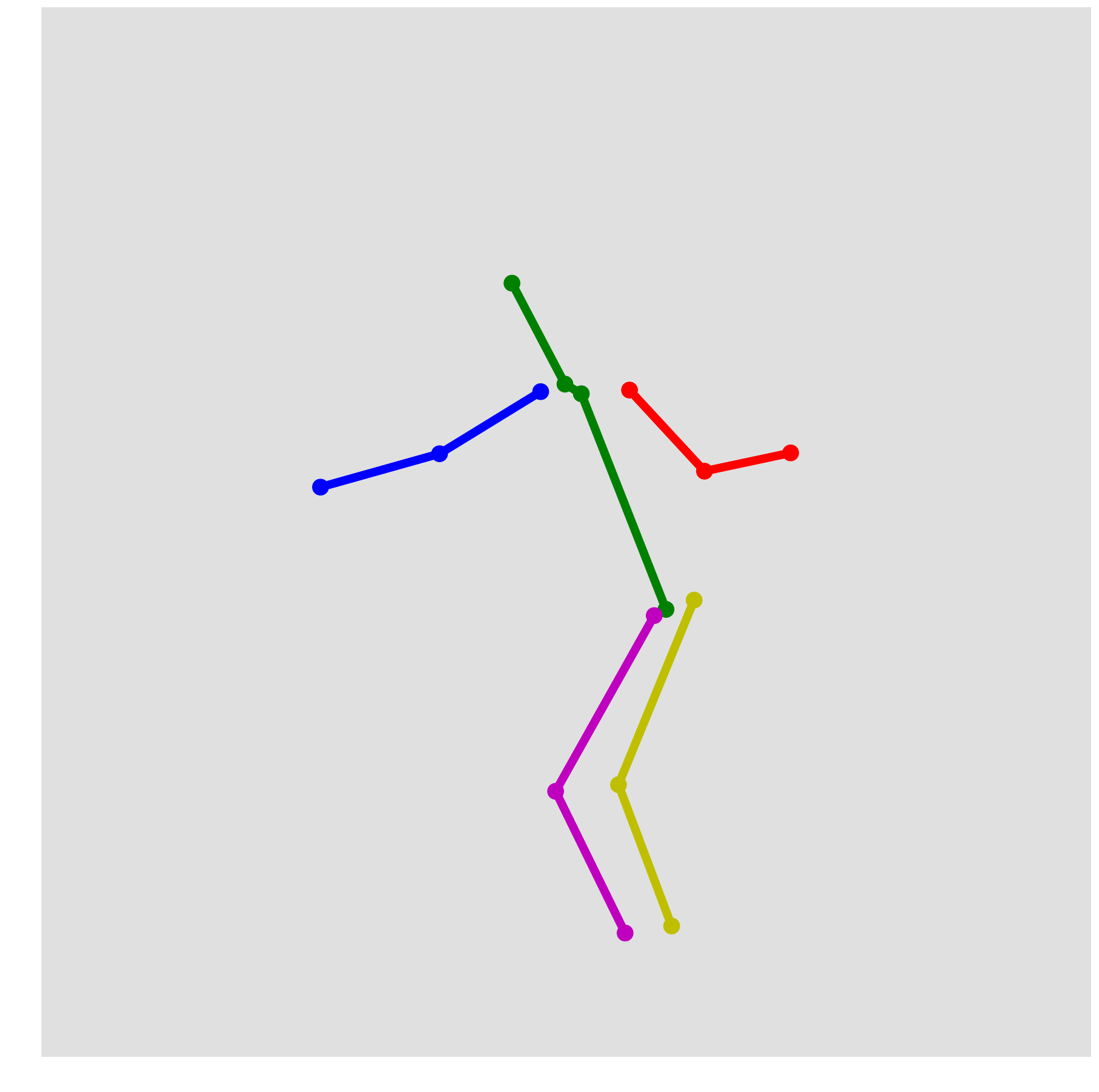}\hspace{-1mm}
    \includegraphics[width=0.075\textwidth]{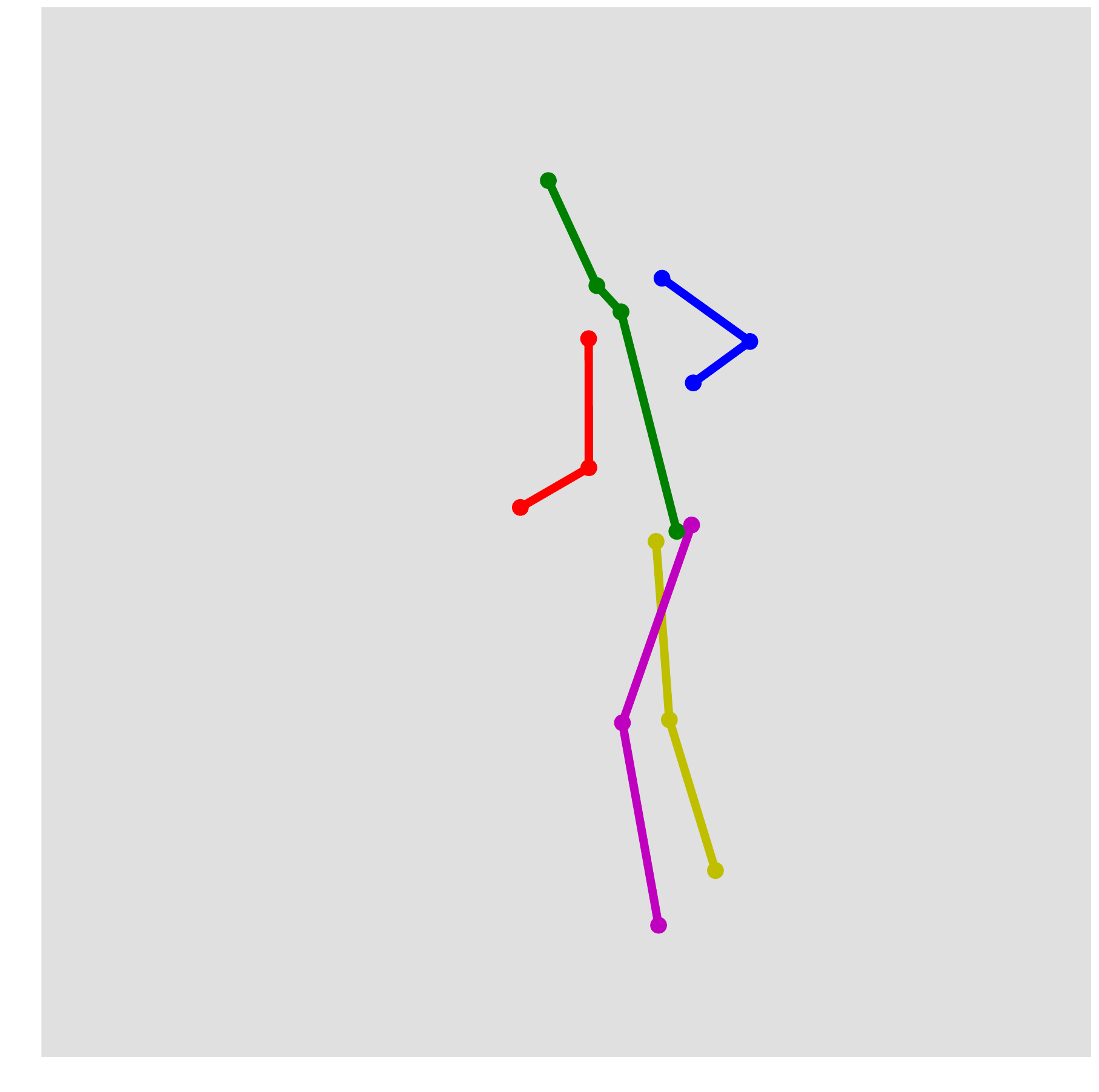}\hspace{-1mm}

    \includegraphics[width=0.075\textwidth]{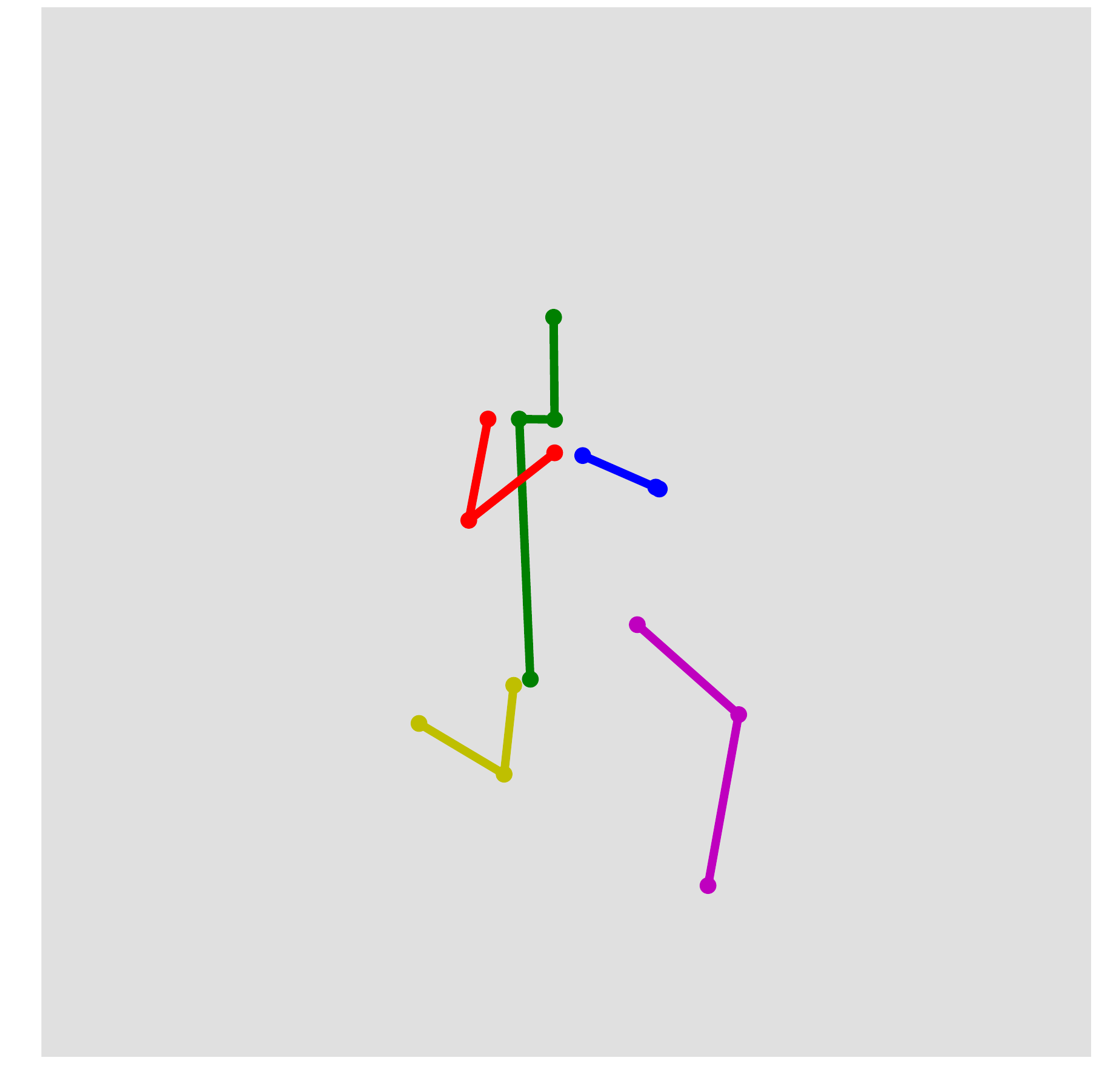}\hspace{-1mm}
    \includegraphics[width=0.075\textwidth]{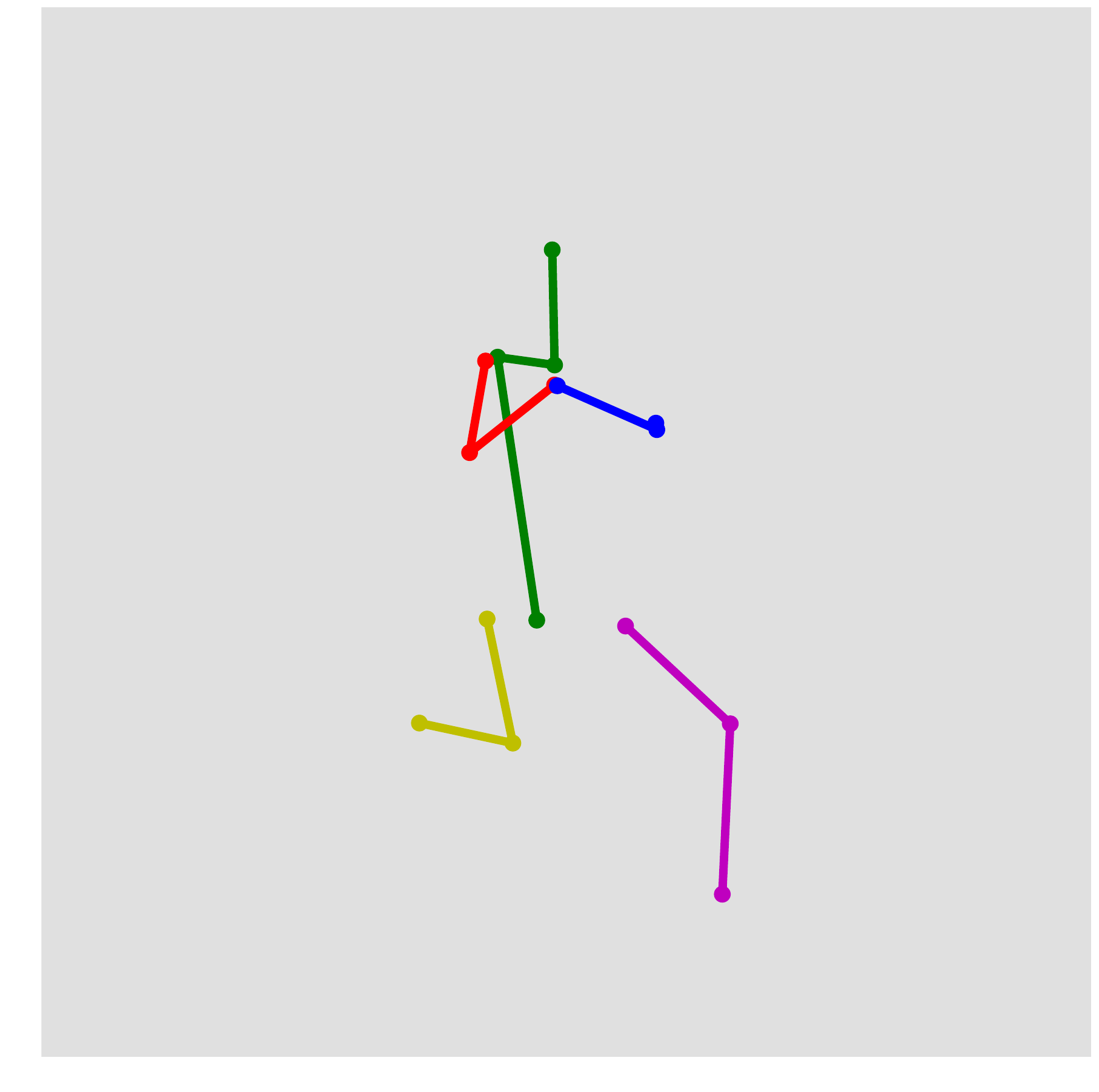}\hspace{-1mm}
    \includegraphics[width=0.075\textwidth]{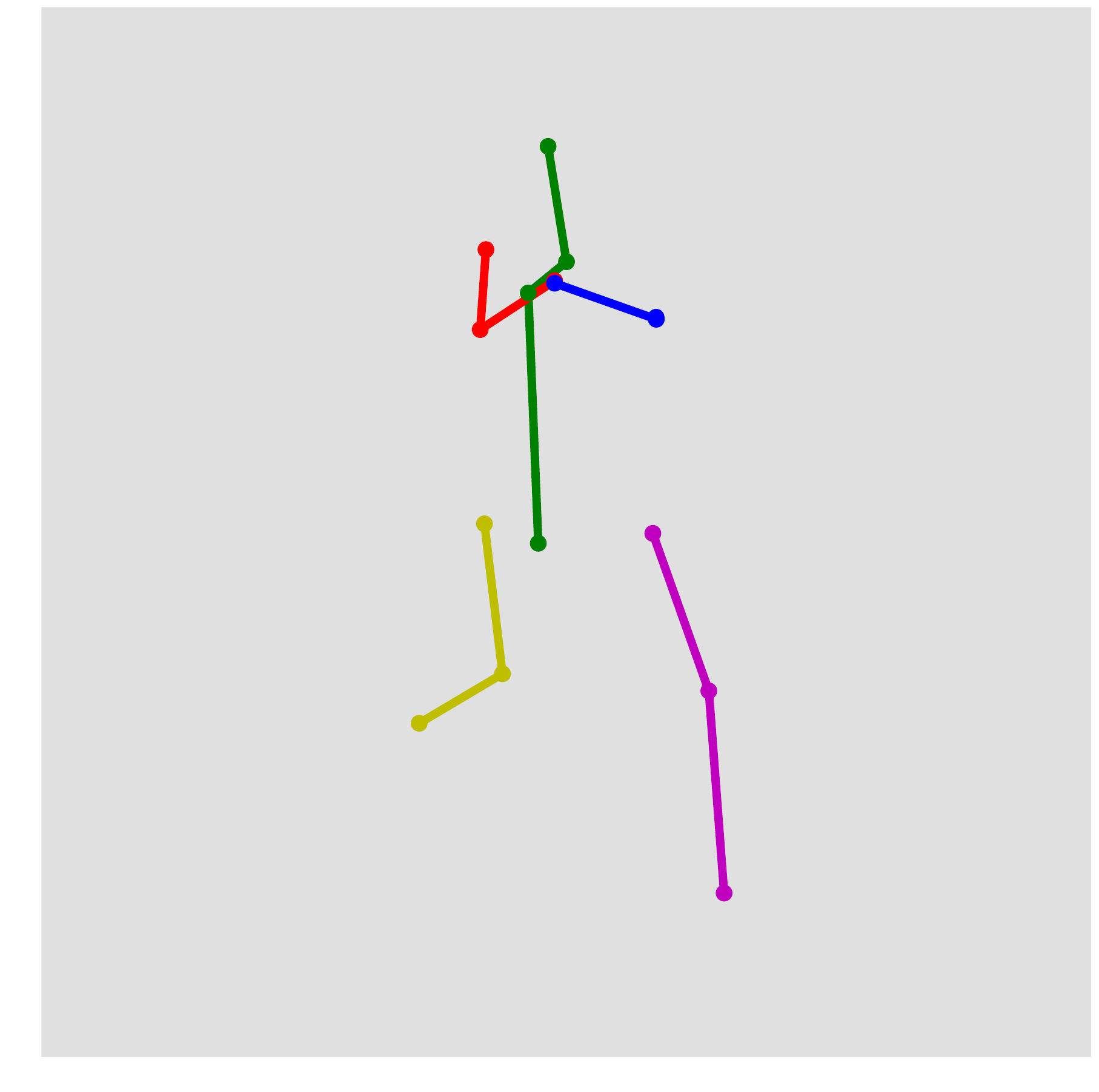}\hspace{1mm}
    \includegraphics[width=0.075\textwidth]{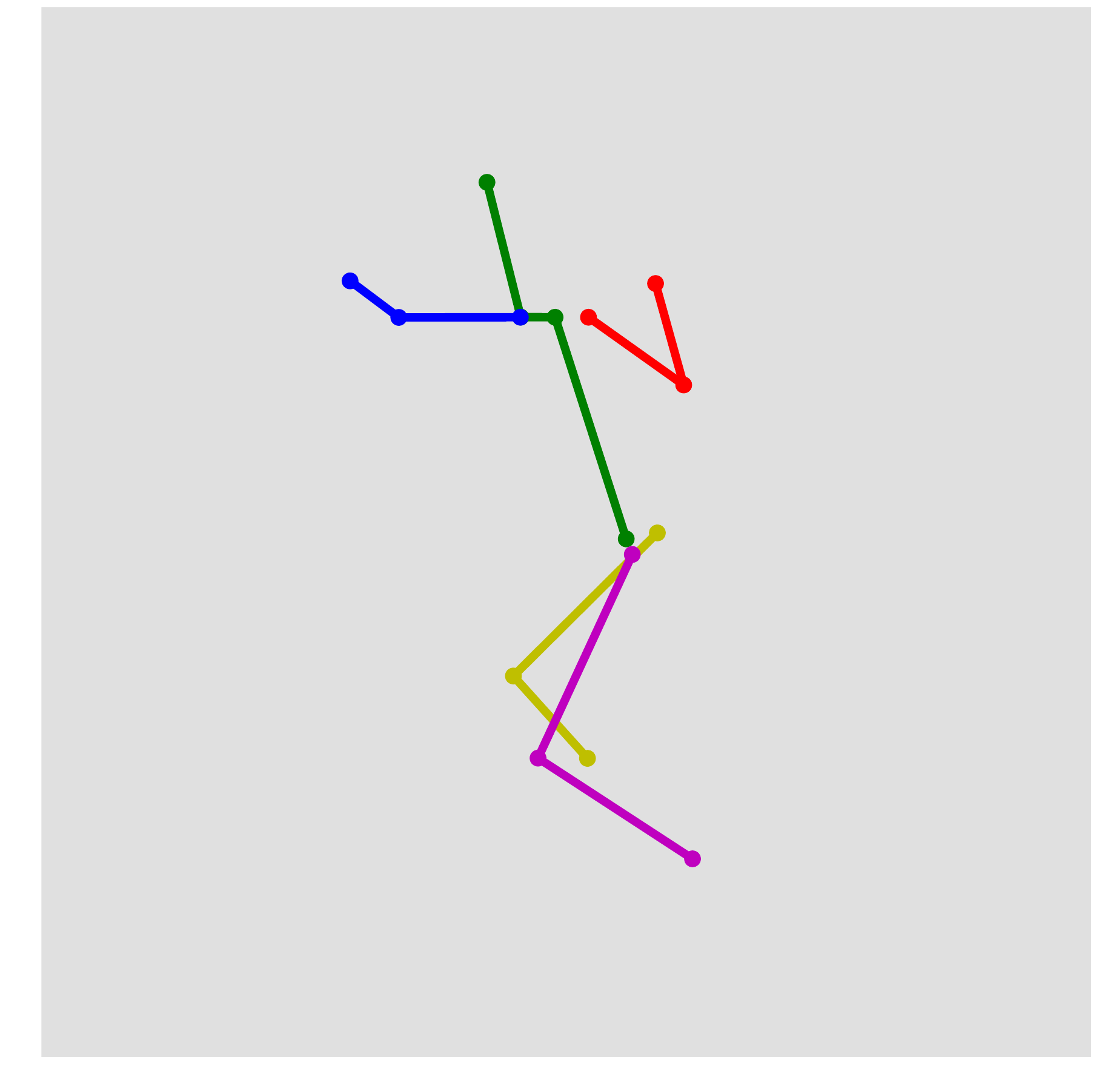}\hspace{-1mm}
    \includegraphics[width=0.075\textwidth]{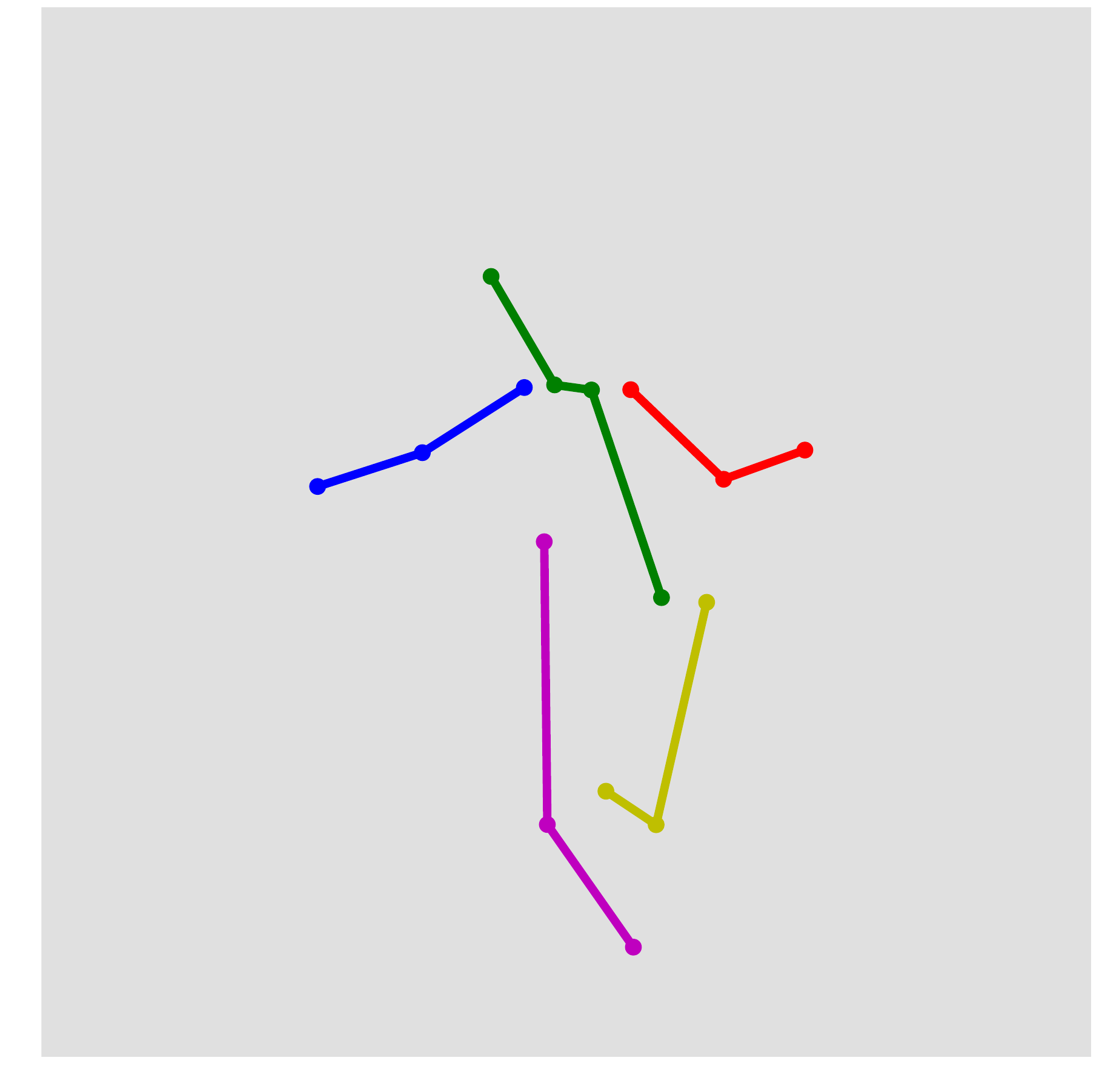}\hspace{-1mm}
    \includegraphics[width=0.075\textwidth]{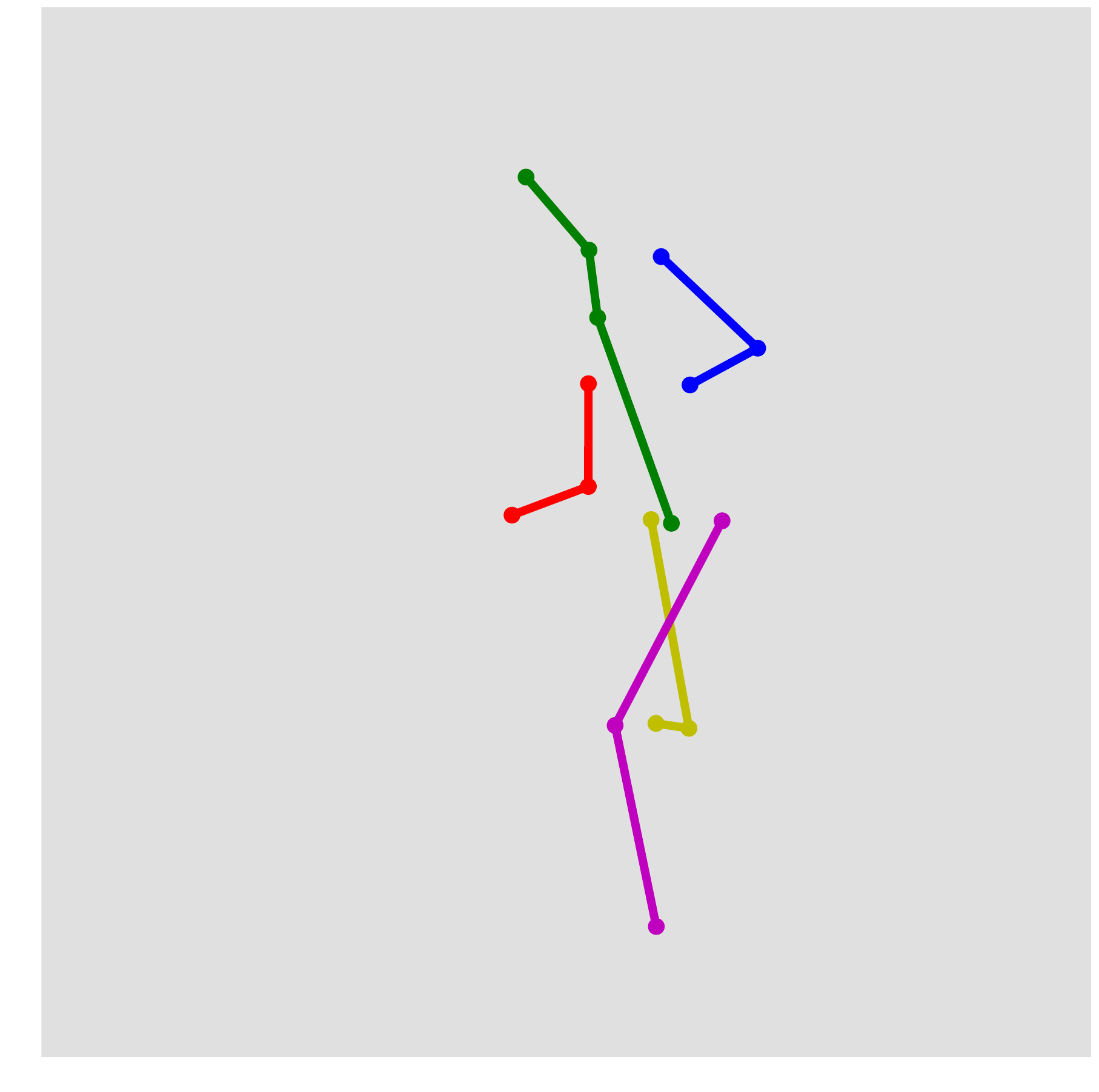}\hspace{-1mm}
  \caption{
    Two sequences of RGB images (top),
    predicted supervised poses (middle), and decoupled action poses (bottom).
  }
  \label{fig:ex-decoupled-poses}
\end{figure}

\begin{figure}[htbp]
  \centering
    \includegraphics[height=1.85cm]{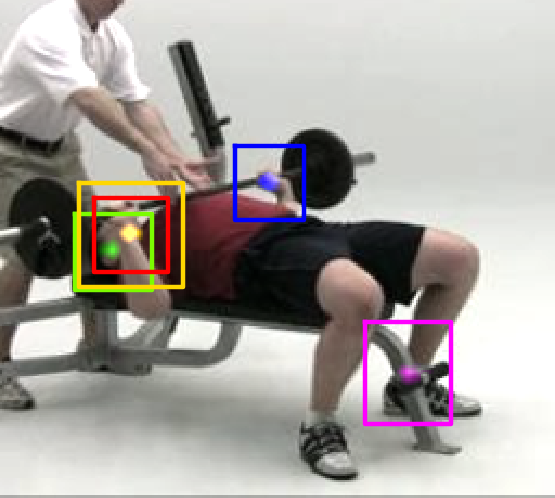}
    \includegraphics[height=1.85cm]{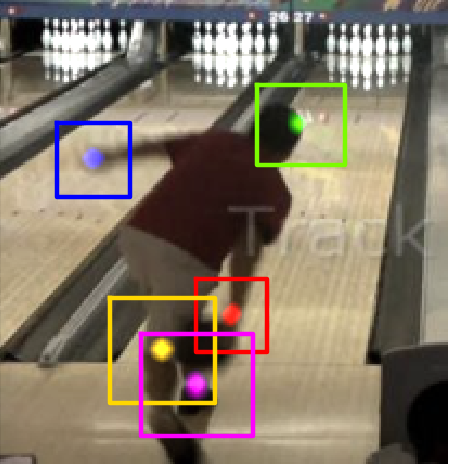}
    \includegraphics[height=1.85cm]{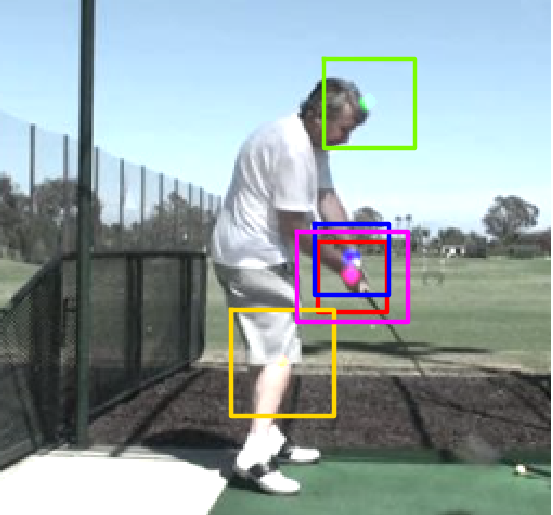}
    \includegraphics[height=1.85cm]{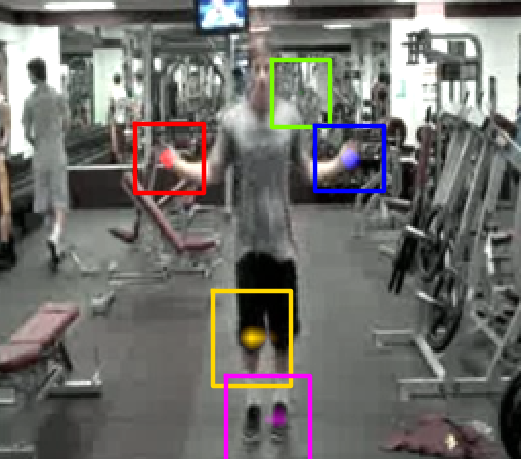}\vspace{1mm}
    \includegraphics[height=1.85cm]{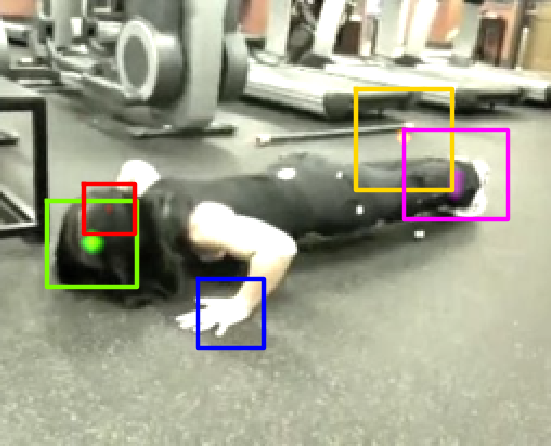}
    \includegraphics[height=1.85cm]{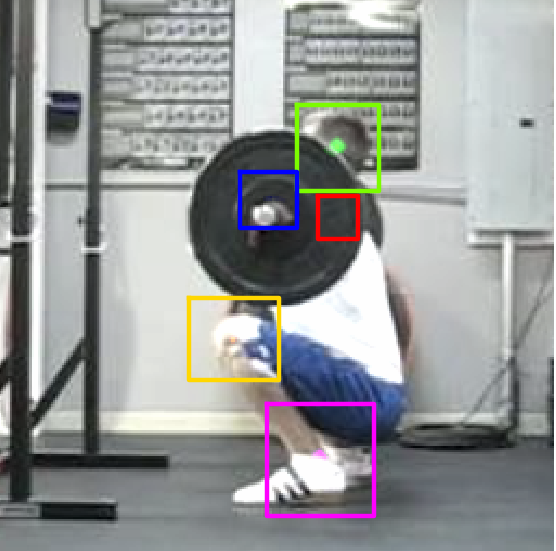}
    \includegraphics[height=1.85cm]{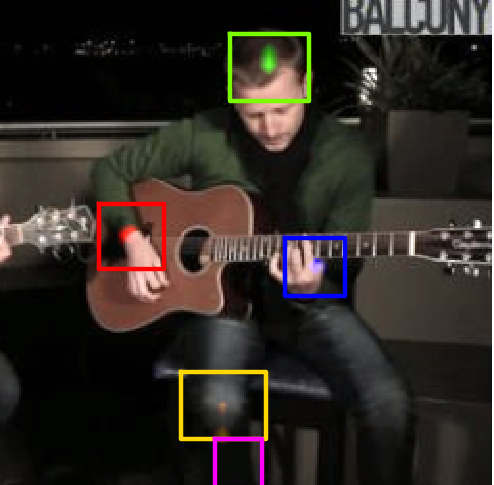}
    \includegraphics[height=1.85cm]{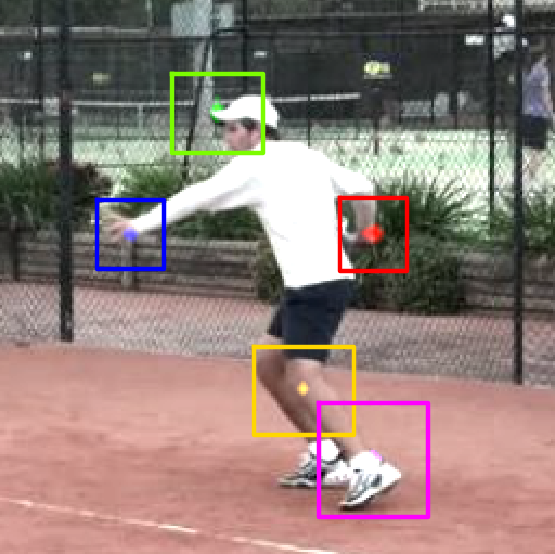}
  \caption{
    Drift of decoupled probability maps from their original positions (head,
    hands and feet) used as an attention mechanism for appearance features
    extraction. Bounding boxes are drawn here only to highlight the regions with
    high responses. \rev{Each color corresponds to a specific body part (see Fig.~\ref{fig:ex-decoupled-poses}).}
  }
  \label{fig:ex-apperance-attention}
\end{figure}

Additionally, we can observe that decoupled action poses remain coherent with
supervised poses, as shown in Fig.~\ref{fig:ex-decoupled-poses}, which
suggests that the initial pose supervision is a good initialization overall.
Nonetheless, in some cases, decoupled probability maps can drift to regions in
the image more relevant for action recognition, as illustrated
in Fig.~\ref{fig:ex-apperance-attention}. For example, feet heat maps can drift
to objects in the hands, since the last is more informative with respect to
the performed action.

\subsubsection{Single-task \textit{vs.} multi-task}

\rev{In this part we compare the results on human action recognition considering
single-task and multi-task training protocols.
In Table~\ref{tab:abla-singletask}, in the first row, are shown results on
PennAction and NTU datasets considering training with action supervision only,
\ie, with the full network architecture (including pose estimation layers)
but without pose supervision.
In the second row we show the results when using the manually annotated poses
from PennAction for pose supervision. We did not use NTU (Kinect) poses for
supervision since they are very noisy. From this, we can notice an improvement
of almost 10\% on PennAction, only by adding pose supervision. When mixing with
MPII data, it further increases 0.8\%. On NTU, multi-tasking improves a
significant 1.9\%. We believe that the improvement of multi-tasking on
PennAction is much more evident because this is a small dataset, therefore it
is difficult to learn good representations for complex actions without explicit
pose information. On the contrary, NTU is a large scale dataset, more suitable
for learning approaches. As a consequence, the gap between single and
multi-task on NTU is smaller, but still relevant.
}

\begin{table}[htbp]
  \centering
  \caption{\rev{Results comparing the effect of single and multi-task training
  for action recognition.}
  }
  \label{tab:abla-singletask}
  \small
  \begin{tabular}{l|cc}
    \hline
    \footnotesize{Training protocol} & \footnotesize{PennAction Acc.} & \footnotesize{NTU Acc.} \\ \hline
    \footnotesize{Single-task (action only)}  & 87.5 & 88.0 \\
    \footnotesize{Multi-task (same dataset)}  & 97.4 & -- \\
    \footnotesize{Multi-task (+MPII +H36M for 3D)} & 98.2 & 89.9 \\
    \hline
  \end{tabular}
\end{table}

\subsubsection{Inference Speed}

\begin{figure*}[htbp]
  \centering
  \begin{subfigure}[b]{0.32\textwidth}
    \centering
    \includegraphics[width=0.9\textwidth]{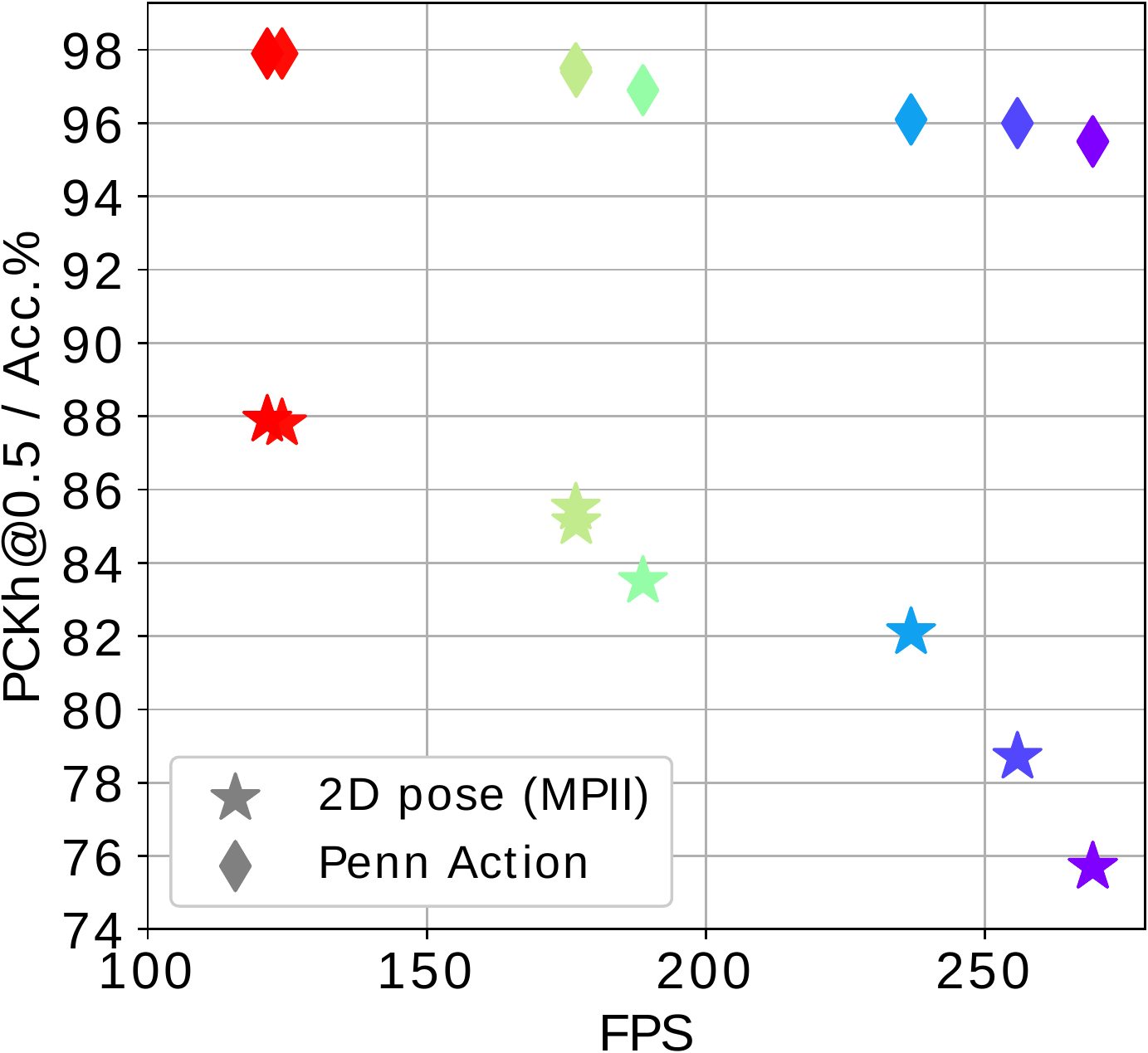}
    \caption{}
  \end{subfigure}
  \begin{subfigure}[b]{0.32\textwidth}
    \centering
    \includegraphics[width=0.9\textwidth]{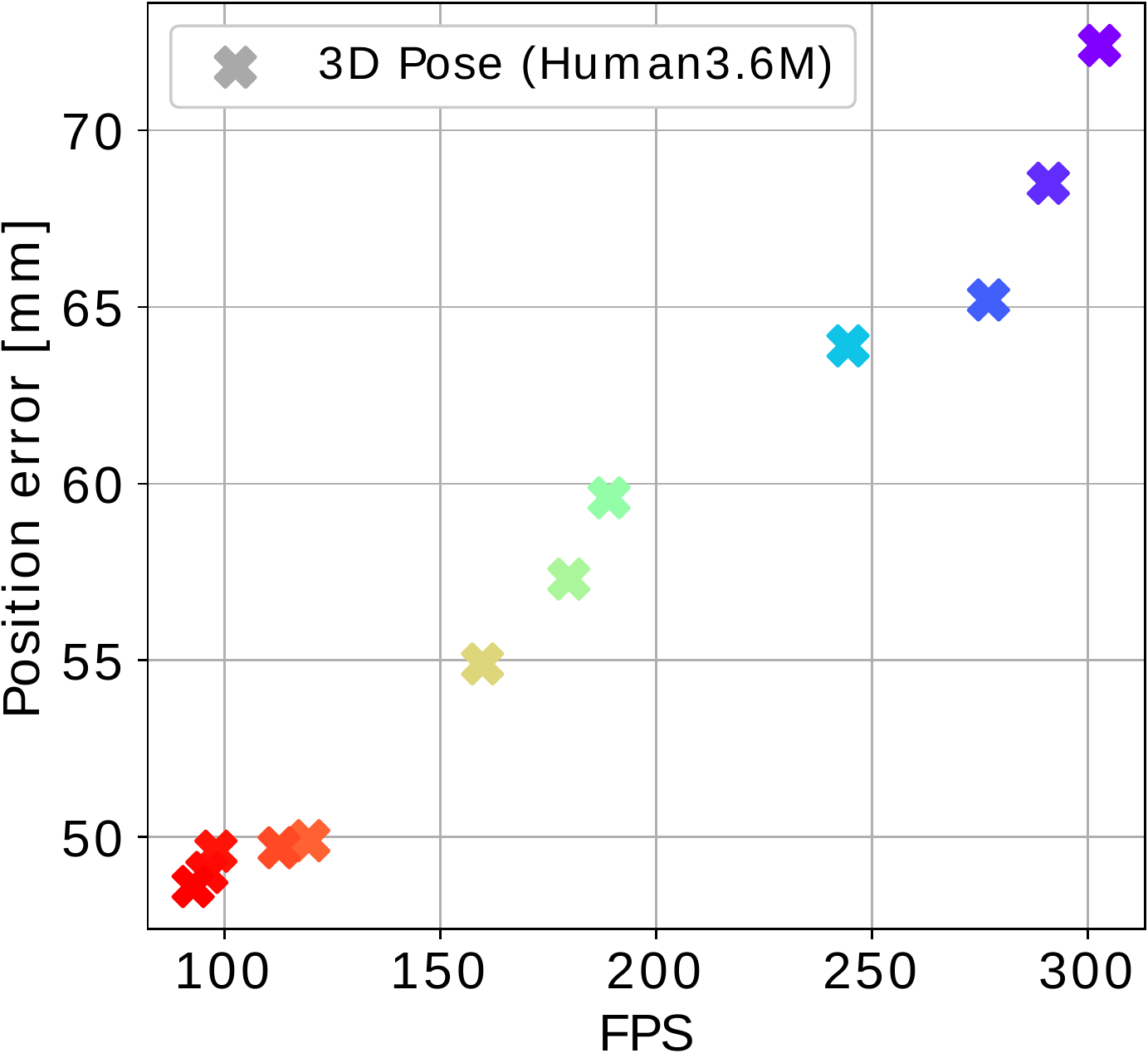}
    \caption{}
  \end{subfigure}
  \begin{subfigure}[b]{0.32\textwidth}
    \centering
    \includegraphics[width=0.9\textwidth]{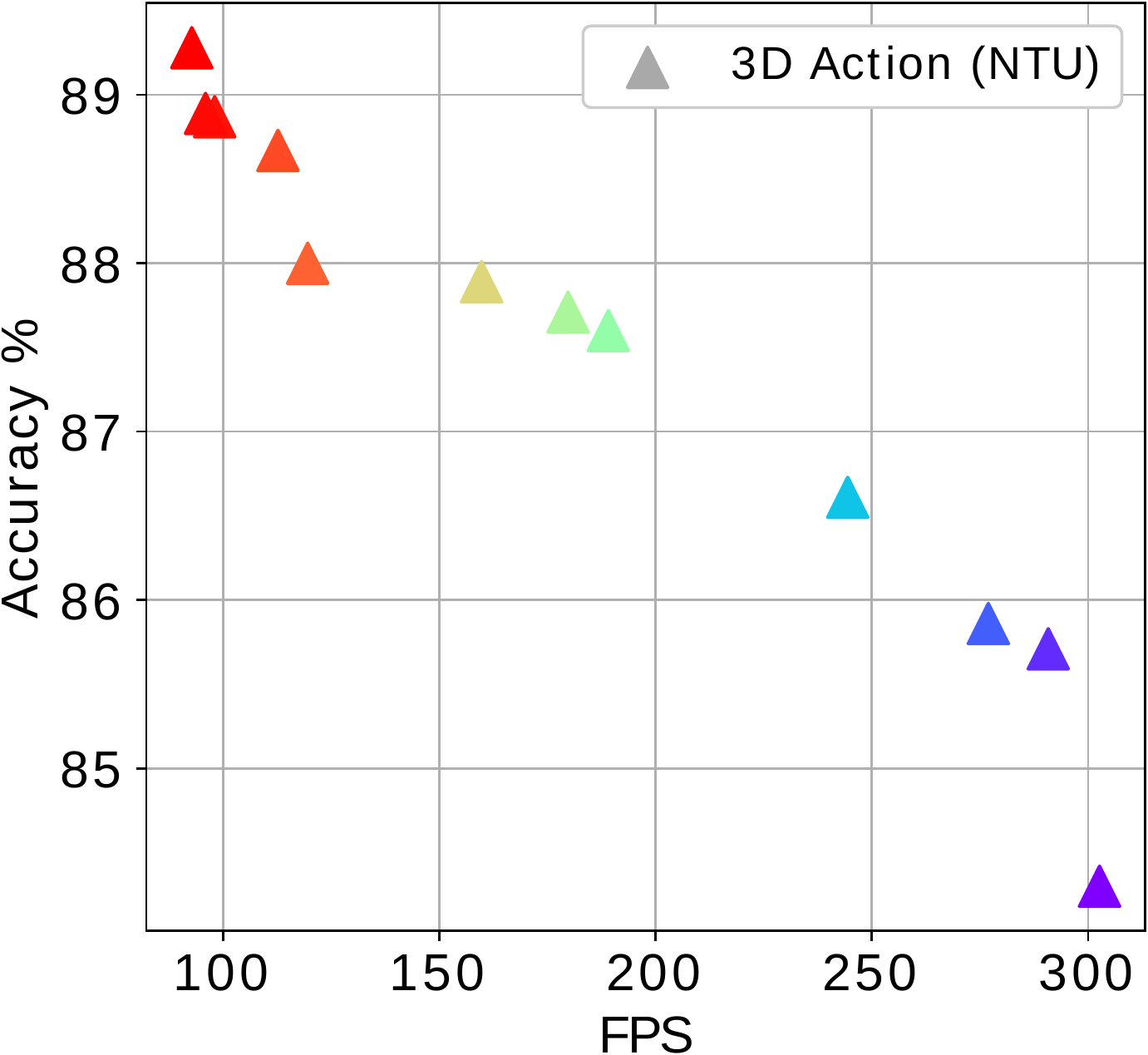}
    \caption{}
  \end{subfigure}
  \caption{
    Inference speed of the proposed method considering 2D (a) and 3D (b,c)
    scenarios. A single multi-task model was trained for each scenario.
    The trained models were cut \textit{a posteriori} for inference analysis.
    \rev{
      Markers with gradient colors from purple to red represent respectively network inferences from faster to slower.}
  }
  \label{fig:prediction-speed}
\end{figure*}

Once the network is trained, it can be easily cut to perform faster
inferences. For instance, the full model with 8 pyramids can be cut at the
4th or 2nd pyramids, which generally degrades the performance, but allows
faster predictions.
To show the trade-off between precision and speed, we cut the trained multi-task
model at different prediction blocks and estimate the \revb{throughput} in frames
per second (FPS), evaluating pose estimation precision and action recognition
classification accuracy. We consider mini batches with 16 images \rev{for pose estimation and single video clips of 8 frames for action}. The results
are shown in Fig.~\ref{fig:prediction-speed}.
For both 2D and 3D scenarios, the best predictions are at more than 90 FPS.
For the 3D scenario, pose estimation on Human3.6M can be performed at more than
180 FPS and still reach a competitive result of 57.3 millimeters error, while
for action recognition on NTU, at the same speed, we still obtain state of the
art results with 87.7\% of correctly classified actions, or even comparable
results with recent approaches at more than 240 FPS.
Finally, we show our results for both 2D and 3D scenarios compared to
previous methods in Table~\ref{tab:results-multitask}, considering
different inference speed. Note that our method is the only to perform both
pose and action estimation in a single prediction, while achieving
state-of-the-art results at a very high speed.

\begin{table}[h]
  \centering
  \caption{Results on all tasks with the proposed multi-task model
  compared to recent approaches using RGB images and/or estimated poses
  on MPII PCKh validation set (higher is better),
  Human3.6M MPJPE (lower is better), Penn Action and
  NTU RGB+D action classification accuracy (higher is better).
  }
  \label{tab:results-multitask}
  \footnotesize
  \begin{tabular}{@{}lcccc@{}}
    \hline
    Methods & \begin{tabular}[c]{@{}c@{}}MPII\\\footnotesize PCKh\end{tabular} & \begin{tabular}[c]{@{}c@{}}H36M\\\footnotesize MPJPE\end{tabular} & \begin{tabular}[c]{@{}c@{}}\footnotesize PennAction\\\textit{half/half}\end{tabular} & \begin{tabular}[c]{@{}c@{}}\footnotesize NTU RGB+D\\Cross-sub.\end{tabular} \\ \hline
    Pavlakos \etal \cite{Pavlakos_2017_CVPR}   & -    & 71.9 & -    & - \\
    Mehta \etal \cite{MehtaRCSXT16}            & -    & 68.6 & -    & - \\
    Martinez \etal \cite{Martinez_2017}        & -    & 62.9 & -    & - \\
    Sun \etal \cite{Sun_2017_ICCV}             & -    & 59.1 & -    & - \\
    Yang \etal~\cite{Yang_2018_CVPR}           & \textbf{88.6} & 58.6 & -    & - \\
    Sun \etal \cite{Sun_2018_ECCV}             & 87.3 & 49.6 & -    & - \\
    Nie \etal \cite{Nie_2015_CVPR}             & -    & -    & 85.5 & - \\
    Iqbal \etal \cite{Iqbal_2017}              & -    & -    & 92.9 & - \\
    Cao \etal \cite{Cao_2017}                  & -    & -    & 95.3 & - \\
    Du \etal \cite{Du_2017_ICCV}               & -    & -    & 97.4 & - \\
    \footnotesize{Shahroudy \etal \cite{Shahroudy2017DeepMF}} & -    & -    & -    & 74.9 \\
    Baradel \etal \cite{Baradel_2018_CVPR}     & -    & -    & -    & 86.6 \\
    \hline
    Ours~\cite{Luvizon_2018_CVPR} \footnotesize{@ 85 fps} & - & 53.2  & 97.4 & 85.5 \\ \hline
    \textbf{Ours} \footnotesize{2D @ 240 fps} & 85.5 & -    & \textbf{97.5} & - \\
    \textbf{Ours} \footnotesize{2D @ 120 fps} & 88.3 & -    & \textbf{98.7} & - \\ \hline
    \textbf{Ours} \footnotesize{3D @ 240 fps}   & 80.7 & 63.9    & -    & 86.6 \\
    \textbf{Ours} \footnotesize{3D @ 180 fps}   & 83.8 & 57.3    & -    & \textbf{87.7} \\
    \textbf{Ours} \footnotesize{3D @ 90 fps}    & 87.0 & \textbf{48.6} & -    & \textbf{89.9} \\
    \hline
  \end{tabular}
\end{table}

\section{Conclusion}
\label{sec:conclusions}

In this work, we presented a new approach for human pose estimation and
action recognition using multi-task deep learning.
The proposed method for 3D pose provides highly precise estimations
with low resolution feature maps and departs from requiring the expensive
volumetric heat maps by predicting specialized depth maps per body joints.
The proposed CNN architecture, along with the pose regression method, allows
multi-scale pose and action supervision and re-injection, resulting in a highly
efficient densely supervised approach.
Our method can be trained with mixed 2D and 3D data, benefiting from precise
indoor 3D data, as well as ``in-the-wild'' images manually annotated with
2D poses. This has demonstrated significant improvements for 3D pose
estimation. The proposed method can also be trained with single frames and
video clips simultaneously and in a seamless way.

More importantly, \rev{we show that the hard problem of multi-tasking human poses and action recognition can be handled
by a carefully designed architecture, resulting in a better solution for
each task than learning them separately. In addition, we show that joint
learning human poses results in consistent improvement of action recognition.}
Finally, with a single training procedure, our multi-task model can be cut at
different levels for pose and action predictions, resulting in
a highly scalable approach.

\section*{Acknowledgment}

This work was partially supported by the Brazilian National Council for
Scientific and Technological Development (CNPq) -- Grant 233342/2014-1.


{
  \small
  \bibliographystyle{ieee}
  \bibliography{references}

\begin{thebibliography}{10}\itemsep=-1pt

\bibitem{Andriluka_CVPR_2014}
M.~Andriluka, L.~Pishchulin, P.~Gehler, and B.~Schiele.
\newblock {2D Human Pose Estimation: New Benchmark and State of the Art
  Analysis}.
\newblock In {\em {IEEE Conference on Computer Vision and Pattern Recognition
  (CVPR)}}, 2014.

\bibitem{Andriluka_CVPR_2009}
M.~Andriluka, S.~Roth, and B.~Schiele.
\newblock {Pictorial structures revisited: People detection and articulated
  pose estimation}.
\newblock In {\em Computer Vision and Pattern Recognition (CVPR)}, pages
  1014--1021, 2009.

\bibitem{baradel2017a}
F.~Baradel, C.~Wolf, and J.~Mille.
\newblock Pose-conditioned spatio-temporal attention for human action
  recognition.
\newblock {\em CoRR}, abs/1703.10106, 2017.

\bibitem{Baradel_CVPR_2018}
F.~Baradel, C.~Wolf, J.~Mille, and G.~W. Taylor.
\newblock Glimpse clouds: Human activity recognition from unstructured feature
  points.
\newblock In {\em Computer Vision and Pattern Recognition (CVPR) (To appear)},
  June 2018.

\bibitem{Baradel_2018_CVPR}
F.~Baradel, C.~Wolf, J.~Mille, and G.~W. Taylor.
\newblock {Glimpse Clouds: Human Activity Recognition from Unstructured Feature
  Points}.
\newblock In {\em {Computer Vision and Pattern Recognition (CVPR)}}, June 2018.

\bibitem{Belagiannis_ICCV_2015}
V.~Belagiannis, C.~Rupprecht, G.~Carneiro, and N.~Navab.
\newblock Robust optimization for deep regression.
\newblock In {\em International Conference on Computer Vision (ICCV)}, pages
  2830--2838, Dec 2015.

\bibitem{Bulat_ECCV_2016}
A.~Bulat and G.~Tzimiropoulos.
\newblock {Human pose estimation via Convolutional Part Heatmap Regression}.
\newblock In {\em European Conference on Computer Vision (ECCV)}, pages
  717--732, 2016.

\bibitem{Cao_2017}
C.~{Cao}, Y.~{Zhang}, C.~{Zhang}, and H.~{Lu}.
\newblock Body joint guided 3-d deep convolutional descriptors for action
  recognition.
\newblock {\em IEEE Transactions on Cybernetics}, 48(3):1095--1108, March 2018.

\bibitem{Carreira_CVPR_2016}
J.~Carreira, P.~Agrawal, K.~Fragkiadaki, and J.~Malik.
\newblock Human pose estimation with iterative error feedback.
\newblock In {\em 2016 IEEE Conference on Computer Vision and Pattern
  Recognition (CVPR)}, pages 4733--4742, 2016.

\bibitem{Carreira_2017_CVPR}
J.~Carreira and A.~Zisserman.
\newblock Quo vadis, action recognition? a new model and the kinetics dataset.
\newblock In {\em CVPR}, July 2017.

\bibitem{Chen_2017_CVPR}
C.-H. Chen and D.~Ramanan.
\newblock 3d human pose estimation = 2d pose estimation + matching.
\newblock In {\em The IEEE Conference on Computer Vision and Pattern
  Recognition (CVPR)}, July 2017.

\bibitem{Chen_2017_ICCV}
Y.~Chen, C.~Shen, X.-S. Wei, L.~Liu, and J.~Yang.
\newblock Adversarial posenet: A structure-aware convolutional network for
  human pose estimation.
\newblock In {\em The IEEE International Conference on Computer Vision (ICCV)},
  Oct 2017.

\bibitem{cheronICCV15}
G.~Ch\'{e}ron, I.~Laptev, and C.~Schmid.
\newblock {P-CNN: Pose-based CNN Features for Action Recognition}.
\newblock In {\em ICCV}, 2015.

\bibitem{Chollet_2017_CVPR}
F.~Chollet.
\newblock Xception: Deep learning with depthwise separable convolutions.
\newblock In {\em The IEEE Conference on Computer Vision and Pattern
  Recognition (CVPR)}, July 2017.

\bibitem{ChouCC17}
C.~Chou, J.~Chien, and H.~Chen.
\newblock Self adversarial training for human pose estimation.
\newblock {\em CoRR}, abs/1707.02439, 2017.

\bibitem{Choutas_2018_CVPR}
V.~Choutas, P.~Weinzaepfel, J.~Revaud, and C.~Schmid.
\newblock Potion: Pose motion representation for action recognition.
\newblock In {\em The IEEE Conference on Computer Vision and Pattern
  Recognition (CVPR)}, June 2018.

\bibitem{Chu_CVPR_17}
X.~Chu, W.~Yang, W.~Ouyang, C.~Ma, A.~L. Yuille, and X.~Wang.
\newblock Multi-context attention for human pose estimation.
\newblock In {\em The IEEE Conference on Computer Vision and Pattern
  Recognition (CVPR)}, July 2017.

\bibitem{Dantone_CVPR_2013}
M.~Dantone, J.~Gall, C.~Leistner, and L.~V. Gool.
\newblock {Human Pose Estimation Using Body Parts Dependent Joint Regressors}.
\newblock In {\em {Computer Vision and Pattern Recognition (CVPR)}}, pages
  3041--3048, June 2013.

\bibitem{Du_2017_ICCV}
W.~Du, Y.~Wang, and Y.~Qiao.
\newblock Rpan: An end-to-end recurrent pose-attention network for action
  recognition in videos.
\newblock In {\em The IEEE International Conference on Computer Vision (ICCV)},
  Oct 2017.

\bibitem{Gkioxari_ECCV_2016}
G.~Gkioxari, A.~Toshev, and N.~Jaitly.
\newblock {Chained Predictions Using Convolutional Neural Networks}.
\newblock {\em European Conference on Computer Vision (ECCV)}, 2016.

\bibitem{Goodfellow_GANS}
I.~Goodfellow, J.~Pouget-Abadie, M.~Mirza, B.~Xu, D.~Warde-Farley, S.~Ozair,
  A.~Courville, and Y.~Bengio.
\newblock Generative adversarial nets.
\newblock In Z.~Ghahramani, M.~Welling, C.~Cortes, N.~D. Lawrence, and K.~Q.
  Weinberger, editors, {\em Advances in Neural Information Processing Systems
  27}, pages 2672--2680. Curran Associates, Inc., 2014.

\bibitem{Herath_2017}
S.~Herath, M.~Harandi, and F.~Porikli.
\newblock Going deeper into action recognition: A survey.
\newblock {\em Image and Vision Computing}, 60(Supplement C):4 -- 21, 2017.
\newblock Regularization Techniques for High-Dimensional Data Analysis.

\bibitem{Insafutdinov_ECCV_2016}
E.~Insafutdinov, L.~Pishchulin, B.~Andres, M.~Andriluka, and B.~Schiele.
\newblock {DeeperCut: A Deeper, Stronger, and Faster Multi-Person Pose
  Estimation Model}.
\newblock In {\em {European Conference on Computer Vision (ECCV)}}, May 2016.

\bibitem{h36m_pami}
C.~Ionescu, D.~Papava, V.~Olaru, and C.~Sminchisescu.
\newblock Human3.6m: Large scale datasets and predictive methods for 3d human
  sensing in natural environments.
\newblock {\em TPAMI}, 36(7):1325--1339, jul 2014.

\bibitem{Iqbal_2017}
U.~Iqbal, M.~Garbade, and J.~Gall.
\newblock Pose for action - action for pose.
\newblock {\em FG-2017}, 2017.

\bibitem{Iqbal_2018_ECCV}
U.~Iqbal, P.~Molchanov, T.~Breuel Juergen~Gall, and J.~Kautz.
\newblock Hand pose estimation via latent 2.5d heatmap regression.
\newblock In {\em The European Conference on Computer Vision (ECCV)}, September
  2018.

\bibitem{Jhuang_2013_ICCV}
H.~Jhuang, J.~Gall, S.~Zuffi, C.~Schmid, and M.~J. Black.
\newblock Towards understanding action recognition.
\newblock In {\em The IEEE International Conference on Computer Vision (ICCV)},
  December 2013.

\bibitem{Ke_2017_CVPR}
Q.~Ke, M.~Bennamoun, S.~An, F.~Sohel, and F.~Boussaid.
\newblock A new representation of skeleton sequences for 3d action recognition.
\newblock In {\em The IEEE Conference on Computer Vision and Pattern
  Recognition (CVPR)}, July 2017.

\bibitem{Kokkinos17cvpr}
I.~Kokkinos.
\newblock Ubernet: Training a 'universal' convolutional neural network for
  low-, mid-, and high-level vision using diverse datasets and limited memory.
\newblock {\em {Computer Vision and Pattern Recognition (CVPR)}}, 2017.

\bibitem{Lifshitz_ECCV_2016}
I.~Lifshitz, E.~Fetaya, and S.~Ullman.
\newblock {\em Human Pose Estimation Using Deep Consensus Voting}, pages
  246--260.
\newblock Springer International Publishing, Cham, 2016.

\bibitem{Liu2016}
J.~Liu, A.~Shahroudy, D.~Xu, and G.~Wang.
\newblock Spatio-temporal lstm with trust gates for 3d human action
  recognition.
\newblock In B.~Leibe, J.~Matas, N.~Sebe, and M.~Welling, editors, {\em ECCV},
  pages 816--833, Cham, 2016.

\bibitem{Liu_2017_CVPR}
J.~Liu, G.~Wang, P.~Hu, L.-Y. Duan, and A.~C. Kot.
\newblock Global context-aware attention lstm networks for 3d action
  recognition.
\newblock In {\em The IEEE Conference on Computer Vision and Pattern
  Recognition (CVPR)}, 2017.

\bibitem{Liu_2018_CVPR}
M.~Liu and J.~Yuan.
\newblock Recognizing human actions as the evolution of pose estimation maps.
\newblock In {\em The IEEE Conference on Computer Vision and Pattern
  Recognition (CVPR)}, June 2018.

\bibitem{Luvizon_2018_CVPR}
D.~C. Luvizon, D.~Picard, and H.~Tabia.
\newblock 2d/3d pose estimation and action recognition using multitask deep
  learning.
\newblock In {\em The IEEE Conference on Computer Vision and Pattern
  Recognition (CVPR)}, June 2018.

\bibitem{Luvizon_PRL_2017}
D.~C. Luvizon, H.~Tabia, and D.~Picard.
\newblock {Learning features combination for human action recognition from
  skeleton sequences}.
\newblock {\em Pattern Recognition Letters}, 2017.

\bibitem{Luvizon_2017_CoRR}
D.~C. Luvizon, H.~Tabia, and D.~Picard.
\newblock Human pose regression by combining indirect part detection and
  contextual information.
\newblock {\em Computers and Graphics}, 85:15 -- 22, 2019.

\bibitem{Martinez_2017}
J.~Martinez, R.~Hossain, J.~Romero, and J.~J. Little.
\newblock A simple yet effective baseline for 3d human pose estimation.
\newblock In {\em ICCV}, 2017.

\bibitem{MehtaRCSXT16}
D.~{Mehta}, H.~{Rhodin}, D.~{Casas}, P.~{Fua}, O.~{Sotnychenko}, W.~{Xu}, and
  C.~{Theobalt}.
\newblock Monocular 3d human pose estimation in the wild using improved cnn
  supervision.
\newblock In {\em 2017 International Conference on 3D Vision (3DV)}, pages
  506--516, Oct 2017.

\bibitem{VNect_SIGGRAPH2017}
D.~Mehta, S.~Sridhar, O.~Sotnychenko, H.~Rhodin, M.~Shafiei, H.-P. Seidel,
  W.~Xu, D.~Casas, and C.~Theobalt.
\newblock Vnect: Real-time 3d human pose estimation with a single rgb camera.
\newblock In {\em ACM Transactions on Graphics}, volume~36, 2017.

\bibitem{Newell_ECCV_2016}
A.~Newell, K.~Yang, and J.~Deng.
\newblock {Stacked Hourglass Networks for Human Pose Estimation}.
\newblock {\em {European Conference on Computer Vision (ECCV)}}, pages
  483--499, 2016.

\bibitem{Ning2017}
G.~Ning, Z.~Zhang, and Z.~He.
\newblock Knowledge-guided deep fractal neural networks for human pose
  estimation.
\newblock {\em IEEE Transactions on Multimedia}, PP(99):1--1, 2017.

\bibitem{Pavlakos_2017_CVPR}
G.~Pavlakos, X.~Zhou, K.~G. Derpanis, and K.~Daniilidis.
\newblock Coarse-to-fine volumetric prediction for single-image 3{D} human
  pose.
\newblock In {\em the IEEE Conference on Computer Vision and Pattern
  Recognition}, 2017.

\bibitem{pfister2014deep}
T.~Pfister, K.~Simonyan, J.~Charles, and A.~Zisserman.
\newblock Deep convolutional neural networks for efficient pose estimation in
  gesture videos.
\newblock In {\em Asian Conference on Computer Vision (ACCV)}, 2014.

\bibitem{Pishchulin_CVPR_2013}
L.~Pishchulin, M.~Andriluka, P.~Gehler, and B.~Schiele.
\newblock {Poselet Conditioned Pictorial Structures}.
\newblock In {\em Computer Vision and Pattern Recognition (CVPR)}, pages
  588--595, 2013.

\bibitem{Pishchulin_CVPR_2016}
L.~Pishchulin, E.~Insafutdinov, S.~Tang, B.~Andres, M.~Andriluka, P.~Gehler,
  and B.~Schiele.
\newblock {DeepCut: Joint Subset Partition and Labeling for Multi Person Pose
  Estimation}.
\newblock In {\em IEEE Conference on Computer Vision and Pattern Recognition
  (CVPR)}, June 2016.

\bibitem{Popa_2017_CVPR}
A.-I. Popa, M.~Zanfir, and C.~Sminchisescu.
\newblock Deep multitask architecture for integrated 2d and 3d human sensing.
\newblock In {\em The IEEE Conference on Computer Vision and Pattern
  Recognition (CVPR)}, July 2017.

\bibitem{Presti20163DSH}
L.~L. Presti and M.~L. Cascia.
\newblock 3d skeleton-based human action classification: A survey.
\newblock {\em Pattern Recognition}, 53:130--147, 2016.

\bibitem{Rafi_BMVC_2016}
U.~Rafi, I.~Kostrikov, J.~Gall, and B.~Leibe.
\newblock An efficient convolutional network for human pose estimation.
\newblock In {\em BMVC}, volume~1, page~2, 2016.

\bibitem{SARAFIANOS20161}
N.~Sarafianos, B.~Boteanu, B.~Ionescu, and I.~A. Kakadiaris.
\newblock 3d human pose estimation: A review of the literature and analysis of
  covariates.
\newblock {\em Computer Vision and Image Understanding}, 152(Supplement C):1 --
  20, 2016.

\bibitem{Shahroudy_2016_CVPR}
A.~Shahroudy, J.~Liu, T.-T. Ng, and G.~Wang.
\newblock Ntu rgb+d: A large scale dataset for 3d human activity analysis.
\newblock In {\em CVPR}, June 2016.

\bibitem{Shahroudy2017DeepMF}
A.~Shahroudy, T.-T. Ng, Y.~Gong, and G.~Wang.
\newblock Deep multimodal feature analysis for action recognition in rgb+d
  videos.
\newblock {\em TPAMI}, 2017.

\bibitem{Song_2017_AAAI}
S.~Song, C.~Lan, J.~Xing, W.~Z. (wezeng), and J.~Liu.
\newblock An end-to-end spatio-temporal attention model for human action
  recognition from skeleton data.
\newblock In {\em AAAI Conference on Artificial Intelligence}, 2017.

\bibitem{Sun_2019_CVPR}
K.~Sun, B.~Xiao, D.~Liu, and J.~Wang.
\newblock Deep high-resolution representation learning for human pose
  estimation.
\newblock In {\em The IEEE Conference on Computer Vision and Pattern
  Recognition (CVPR)}, June 2019.

\bibitem{Sun_2017_ICCV}
X.~Sun, J.~Shang, S.~Liang, and Y.~Wei.
\newblock Compositional human pose regression.
\newblock In {\em The IEEE International Conference on Computer Vision (ICCV)},
  Oct 2017.

\bibitem{Sun_2018_ECCV}
X.~Sun, B.~Xiao, F.~Wei, S.~Liang, and Y.~Wei.
\newblock Integral human pose regression.
\newblock In {\em The European Conference on Computer Vision (ECCV)}, September
  2018.

\bibitem{Tekin_2016}
B.~Tekin, P.~M{\'{a}}rquez{-}Neila, M.~Salzmann, and P.~Fua.
\newblock Fusing 2d uncertainty and 3d cues for monocular body pose estimation.
\newblock {\em CoRR}, abs/1611.05708, 2016.

\bibitem{Tome_2017_CVPR}
D.~Tome, C.~Russell, and L.~Agapito.
\newblock Lifting from the deep: Convolutional 3d pose estimation from a single
  image.
\newblock In {\em CVPR}, July 2017.

\bibitem{Tompson_CVPR_2015}
J.~Tompson, R.~Goroshin, A.~Jain, Y.~LeCun, and C.~Bregler.
\newblock {Efficient object localization using Convolutional Networks}.
\newblock In {\em {IEEE Conference on Computer Vision and Pattern Recognition
  (CVPR)}}, pages 648--656, June 2015.

\bibitem{Toshev_CVPR_2014}
A.~Toshev and C.~Szegedy.
\newblock {DeepPose: Human Pose Estimation via Deep Neural Networks}.
\newblock In {\em {Computer Vision and Pattern Recognition (CVPR)}}, pages
  1653--1660, 2014.

\bibitem{varol17a}
G.~Varol, I.~Laptev, and C.~Schmid.
\newblock {Long-term Temporal Convolutions for Action Recognition}.
\newblock {\em TPAMI}, 2017.

\bibitem{Wang_2018_ECCV}
D.~Wang, W.~Ouyang, W.~Li, and D.~Xu.
\newblock Dividing and aggregating network for multi-view action recognition.
\newblock In {\em The European Conference on Computer Vision (ECCV)}, September
  2018.

\bibitem{Wei_CVPR_2016}
S.-E. Wei, V.~Ramakrishna, T.~Kanade, and Y.~Sheikh.
\newblock {Convolutional pose machines}.
\newblock In {\em {IEEE Conference on Computer Vision and Pattern Recognition
  (CVPR)}}, 2016.

\bibitem{Nie_2015_CVPR}
B.~Xiaohan~Nie, C.~Xiong, and S.-C. Zhu.
\newblock Joint action recognition and pose estimation from video.
\newblock In {\em The IEEE Conference on Computer Vision and Pattern
  Recognition (CVPR)}, June 2015.

\bibitem{Yang_2017_ICCV}
W.~Yang, S.~Li, W.~Ouyang, H.~Li, and X.~Wang.
\newblock Learning feature pyramids for human pose estimation.
\newblock In {\em The IEEE International Conference on Computer Vision (ICCV)},
  2017.

\bibitem{Yang_2018_CVPR}
W.~Yang, W.~Ouyang, X.~Wang, J.~S.~J. Ren, H.~Li, and X.~Wang.
\newblock 3d human pose estimation in the wild by adversarial learning.
\newblock In {\em The IEEE Conference on Computer Vision and Pattern
  Recognition (CVPR)}, 2018.

\bibitem{Yao2012}
A.~Yao, J.~Gall, and L.~Van~Gool.
\newblock Coupled action recognition and pose estimation from multiple views.
\newblock {\em International Journal of Computer Vision}, 100(1):16--37, Oct
  2012.

\bibitem{Yi_2016}
K.~M. Yi, E.~Trulls, V.~Lepetit, and P.~Fua.
\newblock {LIFT: Learned Invariant Feature Transform}.
\newblock {\em European Conference on Computer Vision (ECCV)}, 2016.

\bibitem{Zhang_ICCV_2013}
W.~Zhang, M.~Zhu, and K.~G. Derpanis.
\newblock From actemes to action: A strongly-supervised representation for
  detailed action understanding.
\newblock In {\em ICCV}, pages 2248--2255, Dec 2013.

\bibitem{Zhou_2017}
X.~Zhou, M.~Zhu, G.~Pavlakos, S.~Leonardos, K.~G. Derpanis, and K.~Daniilidis.
\newblock Monocap: Monocular human motion capture using a {CNN} coupled with a
  geometric prior.
\newblock {\em CoRR}, abs/1701.02354, 2017.

\bibitem{Zolfaghari_2017_ICCV}
M.~Zolfaghari, G.~L. Oliveira, N.~Sedaghat, and T.~Brox.
\newblock Chained multi-stream networks exploiting pose, motion, and appearance
  for action classification and detection.
\newblock In {\em The IEEE International Conference on Computer Vision (ICCV)},
  Oct 2017.

\bibitem{Zou05regularizationand}
H.~Zou and T.~Hastie.
\newblock Regularization and variable selection via the elastic net.
\newblock {\em Journal of the Royal Statistical Society, Series B},
  67:301--320, 2005.

\end{thebibliography}
}

\end{document}